\documentclass[british,10pt,twocolumn,letterpaper]{article}
\usepackage[T1]{fontenc}
\usepackage[latin9]{inputenc}
\usepackage{color}
\usepackage{array}
\usepackage{verbatim}
\usepackage{units}
\usepackage{multirow}
\usepackage{amsmath}
\usepackage{amssymb}
\usepackage{graphicx}

\makeatletter

\providecommand{\tabularnewline}{\\}
\newcommand{\lyxdot}{.}

\usepackage{cvpr}
\usepackage{times}
\usepackage{epsfig}
\usepackage{graphicx}
\usepackage{amsmath}
\usepackage{amssymb}

\usepackage{colortbl}
\usepackage{arydshln}
\usepackage{makecell, pict2e}

\definecolor{lightgray}{gray}{0.8}

\usepackage{caption} 
\usepackage[pagebackref=true,breaklinks=true,letterpaper=true,colorlinks,bookmarks=false]{hyperref}

 \cvprfinalcopy 

\def\cvprPaperID{238} 

\ifcvprfinal\fi

\@ifundefined{showcaptionsetup}{}{%
 \PassOptionsToPackage{caption=false}{subfig}}
\usepackage{subfig}
\makeatother

\usepackage{babel}
\begin{document}

\title{How Far are We from Solving Pedestrian Detection?}

\author{Shanshan Zhang, Rodrigo Benenson, Mohamed Omran, Jan Hosang and Bernt
Schiele\\
\begin{tabular}{c}
Max Planck Institute for Informatics\tabularnewline
Saarbrücken, Germany\tabularnewline
\texttt{\small{}firstname.lastname@mpi-inf.mpg.de}\tabularnewline
\end{tabular}\vspace{-1em}
 \and}
\maketitle
\begin{abstract}
Encouraged by the recent progress in pedestrian detection, we investigate
the gap between current state-of-the-art methods and the ``perfect
single frame detector''. We enable our analysis by creating a human
baseline for pedestrian detection (over the Caltech dataset), and
by manually clustering the recurrent errors of a top detector. Our
results characterise both localisation and background-versus-foreground
errors.

To address localisation errors we study the impact of training annotation
noise on the detector performance, and show that we can improve even
with a small portion of sanitised training data. To address background/foreground
discrimination, we study convnets for pedestrian detection, and discuss
which factors affect their performance.

Other than our in-depth analysis, we report top performance on the
Caltech dataset, and provide a new sanitised set of training and test
annotations.
\end{abstract}
\makeatletter 
\renewcommand{\paragraph}{%
\@startsection{paragraph}{4}%
{\z@}{1.0ex \@plus 1ex \@minus .2ex}{-0.5em}%
{\normalfont \normalsize \bfseries}%
}
\makeatother\setlength{\textfloatsep}{1.5em}

\vspace{-0.5em}

\section{\label{sec:Introduction}Introduction}

Object detection has received great attention during recent years.
Pedestrian detection is a canonical sub-problem that remains a popular
topic of research due to its diverse applications.

Despite the extensive research on pedestrian detection, recent papers
still show significant improvements, suggesting that a saturation
point has not yet been reached. In this paper we analyse the gap between
the state of the art and a newly created human baseline (section \ref{sec:Reaching-saturation}).
The results indicate that there is still a ten fold improvement to
be made before reaching human performance. We aim to investigate which
factors will help close this gap.

We analyse failure cases of top performing pedestrian detectors and
diagnose what should be changed to further push performance. We show
several different analysis, including human inspection, automated
analysis of problem cases (e.g.~blur, contrast), and oracle experiments
(section \ref{sec:Errors-analysis}). Our results indicate that localisation
is an important source of high confidence false positives. We address
this aspect by improving the training set alignment quality, both
by manually sanitising the Caltech training annotations and via algorithmic
means for the remaining training samples (sections \ref{sec:New-annotations}
and \ref{sec:Annotations-impact}).

To address background versus foreground discrimination, we study convnets
for pedestrian detection, and discuss which factors affect their performance
(section \ref{sec:ConvNets}).

\begin{figure}
\noindent \begin{centering}
\includegraphics[width=0.98\columnwidth]{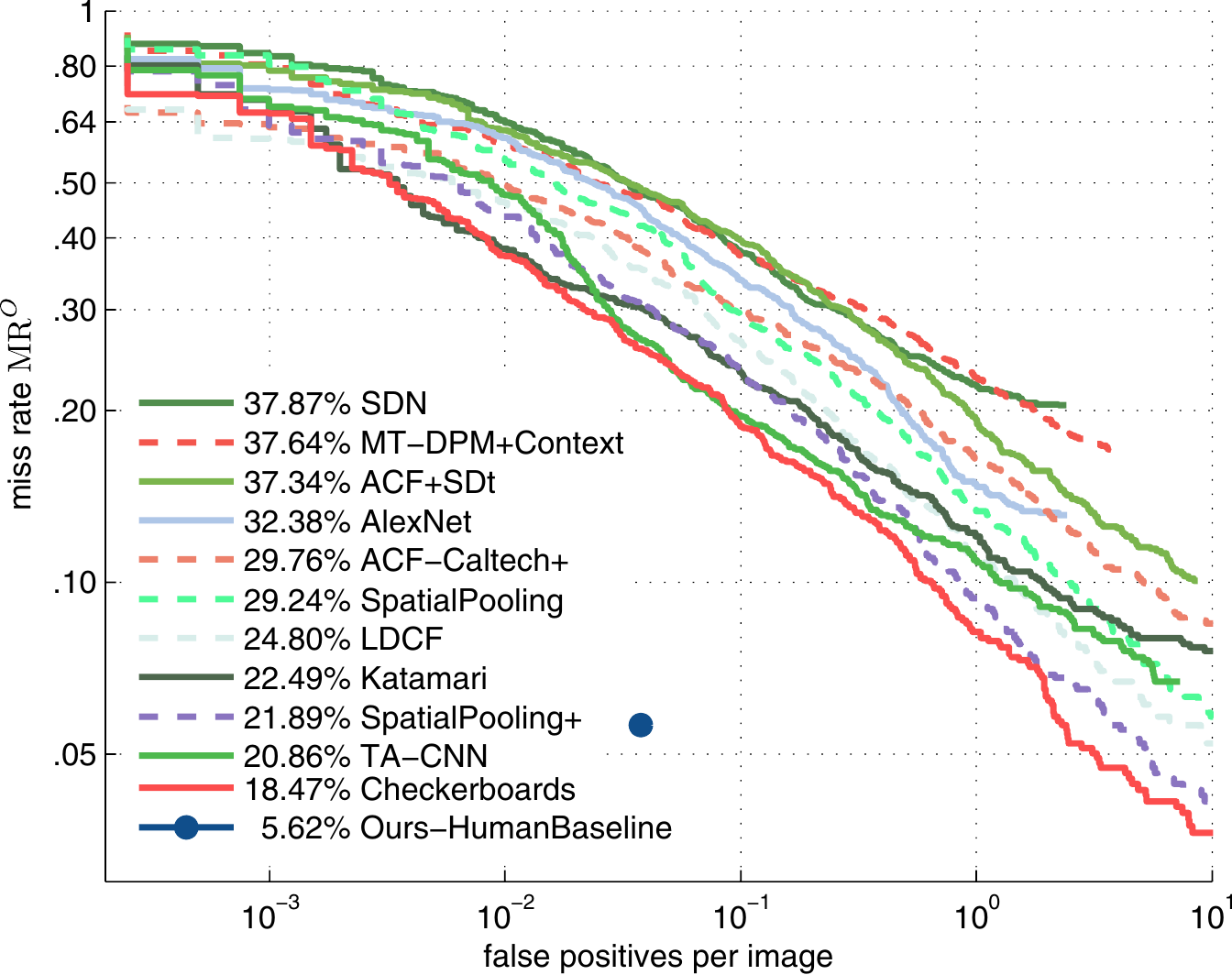}
\par\end{centering}

\caption{\label{fig:cpvr15-top-results}Overview of the top results on the
Caltech-USA pedestrian benchmark (CVPR2015 snapshot). At $\sim\!\!95\%$
recall, state-of-the-art detectors make ten times more errors than
the human baseline.}
\vspace{-0.5em}
\end{figure}

\subsection{\label{sub:Related-work}Related work}

In the last years, diverse efforts have been made to improve the performance
of pedestrian detection. Following the success of integral channel
feature detector (ICF) \cite{Dollar2009Bmvc,Dollar2014Pami}, many
variants \cite{Zhang2014CvprInformedHaar,Zhang2015Cvpr,Nam2014arXiv,Paisitkriangkrai2014Eccv,shanshan_csvt15}
were proposed and showed significant improvement. A recent review
of pedestrian detection \cite{Benenson2014Eccvw} concludes that improved
features have been driving performance and are likely to continue
doing so. It also shows that optical flow \cite{Park2013Cvpr} and
context information \cite{Ouyang2013Cvpr} are complementary to image
features and can further boost detection accuracy. 

By fine-tuning a model pre-trained on external data convolution neural
networks (convnets) have also reached state-of-the-art performance
\cite{Hosang2015Cvpr,Yang2015Cvpr}.

Most of the recent papers focus on introducing novelty and better
results, but neglect the analysis of the resulting system. Some analysis
work can be found for general object detection \cite{Agrawal2014Eccv,Hoiem2012Eccv};
in contrast, in the field of pedestrian detection, this kind of analysis
is rarely done. In 2008, \cite{Wojek08Dagma} provided a failure analysis
on the INRIA dataset, which is relatively small. The best method considered
in the 2012 Caltech dataset survey \cite{Dollar2012Pami} had $10\times$
more false positives at $20\,\%$ recall than the methods considered
here, and no method had reached the $95\,\%$ mark.

Since pedestrian detection has improved significantly in recent years,
a deeper and more comprehensive analysis based on state-of-the-art
detectors is valuable to provide better understanding as to where
future efforts would best be invested.

\subsection{\label{sub:Contributions}Contributions}

Our key contributions are as follows:\\
(a) We provide a detailed analysis of a state-of-the-art pedestrian
detector, providing insights into failure cases. \\
(b) We provide a human baseline for the Caltech Pedestrian Benchmark;
as well as a sanitised version of the annotations to serve as new,
high quality ground truth for the training and test sets of the benchmark.
This data is public\footnote{http://www.mpi-inf.mpg.de/pedestrian\_detection\_cvpr16}.
\\
(c) We analyse the effects of training data quality. More specifically
we quantify how much better alignment and fewer annotation mistakes
can improve performance. \\
(d) Using the insights of the analysis, we explore variants of top
performing methods: filtered channel feature detector \cite{Zhang2015Cvpr}
and R-CNN detector \cite{Girshick2014Cvpr,Hosang2015Cvpr}, and show
improvements over the baselines.

\section{\label{sec:Preliminaries}Preliminaries}

Before delving into our analysis, let us describe the datasets in
use, their metrics, and our baseline detector.

\subsection{\label{sec:Pedestrian-datasets}Caltech-USA pedestrian detection
benchmark}

Amongst existing pedestrian datasets \cite{Dalal2005Cvpr,Ess2008Cvpr,Enzweiler2009PAMI},
KITTI \cite{Geiger2012CVPR} and Caltech-USA are currently the most
popular ones.%
In this work we focus on the Caltech-USA benchmark \cite{Dollar2012Pami}
which consists of 2.5 hours of 30Hz video recorded from a vehicle
traversing the streets of Los Angeles, USA. The video annotations
amount to a total of 350\,000 bounding boxes covering $\sim\negmedspace2\,300$
unique pedestrians. Detection methods are evaluated on a test set
consisting of 4\,024 frames. The provided evaluation toolbox generates
plots for different subsets of the test set based on annotation size,
occlusion level and aspect ratio. The established procedure for training
is to use every 30th video frame which results in a total of 4\,250
frames with $\sim\negmedspace1\,600$ pedestrian cut-outs. More recently,
methods which can leverage more data for training have resorted to
a finer sampling of the videos \cite{Nam2014arXiv,Zhang2015Cvpr},
yielding up to $10\times$ as much data for training than the standard
``$1\times$'' setting.

\paragraph{$\mbox{MR}^{O}$, $\mbox{MR}^{N}$}

In the standard Caltech evaluation \cite{Dollar2012Pami} the miss
rate (MR) is averaged over the low precision range of $[10^{-2},10^{0}]$
FPPI (false positives per image). This metric does not reflect well
improvements in localisation errors (lowest FPPI range). Aiming for
a more complete evaluation, we extend the evaluation FPPI range from
traditional $[10^{-2},10^{0}]$ to $[10^{-4},10^{0}]$, we denote
these \emph{$\mbox{MR}_{-2}^{O}$} and \emph{$\mbox{MR}_{-4}^{O}$.
}$O$ stands for ``original annotations''. In section \ref{sec:New-annotations} we introduce new annotations,
and mark evaluations done there as \emph{$\mbox{MR}_{-2}^{N}$} and
\emph{$\mbox{MR}_{-4}^{N}$}. We expect the \emph{$\mbox{MR}_{-4}$
}metric to become more important as detectors get stronger.

\subsection{\label{sec:FltrChnFtrs}Filtered channel feature detectors}

\begin{table}
\hspace*{-0.5em}%
\begin{minipage}[t]{0.35\columnwidth}%
\begin{center}
\setlength{\tabcolsep}{1pt} 
\begin{tabular}{lc}
{\small{}Filter type} & {\small{}$\mbox{MR}_{-2}^{O}$}\tabularnewline
\hline 
\hline 
{\small{}ACF \cite{Dollar2014Pami}} & 44.2\tabularnewline
{\small{}SCF \cite{Benenson2014Eccvw}} & 34.8\tabularnewline
{\small{}LDCF \cite{Nam2014arXiv}} & 24.8\tabularnewline
{\small{}RotatedFilters}\hspace*{-0.5em} & 19.2\tabularnewline
{\small{}Checkerboards}\hspace*{-0.5em}{\small{} } & 18.5\tabularnewline
\end{tabular}
\par\end{center}

\vspace{-1.5em}

\caption{\label{tab:filtered-channels-detectors}The filter type determines
the ICF methods quality.}
\end{minipage}\hspace{1em}%
\begin{minipage}[t]{0.65\columnwidth}%
\begin{center}
\vspace{-4.1em}
\setlength{\tabcolsep}{2pt} 
\begin{tabular}{lc|cc}
\multirow{1}{*}{{\small{}Base detector}} & {\small{}$\mbox{MR}_{-2}^{O}$} & {\small{}+Context} & {\small{}+Flow}\tabularnewline
\hline 
\hline 
Orig. 2Ped \cite{Ouyang2013Cvpr}\hspace*{-4em} & 48 & \textasciitilde{}5pp & /\tabularnewline
Orig. SDt \cite{Park2013Cvpr} & 45 & / & 8pp\tabularnewline
\hline 
SCF \cite{Benenson2014Eccvw} & 35 & 5pp & 4pp\tabularnewline
Checkerboards\hspace*{-4em} & 19 & \textasciitilde{}0 & 1pp\tabularnewline
\end{tabular}\caption{\label{tab:add-ons-returns}Detection quality gain of adding context
\cite{Ouyang2013Cvpr} and optical flow \cite{Park2013Cvpr}, as function
of the base detector. }

\par\end{center}%
\end{minipage}\vspace{-1em}
\end{table}
For the analysis in this paper we consider all methods published on
the Caltech Pedestrian benchmark, up to the last major conference
(CVPR2015). As shown in figure \ref{fig:cpvr15-top-results}, the
best method at the time is \texttt{Checkerboards}, and most of the
top performing methods are of its same family.

The \texttt{Checkerboards} detector \cite{Zhang2015Cvpr} is a generalisation
of the Integral Channels Feature detector (ICF) \cite{Dollar2009Bmvc},
which filters the HOG+LUV feature channels before feeding them into
a boosted decision forest.

We compare the performance of several detectors from the ICF family
in table \ref{tab:filtered-channels-detectors}, where we can see
a big improvement from 44.2\% to 18.5\% \emph{$\mbox{MR}_{-2}^{O}$}
by introducing filters over the feature channels and optimising the
filter bank.

Current top performing convnets methods \cite{Hosang2015Cvpr,Yang2015Cvpr}
are sensitive to the underlying detection proposals, thus we first
focus on the proposals by optimising the filtered channel feature
detectors (more on convnets in section \ref{sec:ConvNets}).

\paragraph{\label{sec:Rotated-filters}Rotated filters}

For the experiments involving training new models (in section \ref{sec:Annotations-impact})
we use our own re-implementation of \texttt{Checker\-boards} \cite{Zhang2015Cvpr},
based on the \texttt{LDCF} \cite{Nam2014arXiv} codebase. To improve
the training time we decrease the number of filters from 61 in the
original \texttt{Checker\-boards} down to 9 filters. Our so-called
\texttt{Ro\-ta\-ted\-Fil\-ters} are a simplified version of \texttt{LDCF},
applied at three different scales (in the same spirit as \texttt{Squa\-res\-Chn\-Ftrs}
(SCF) {\small{}\cite{Benenson2014Eccvw}}). More details on the filters
are given in the supplementary material. As shown in table \ref{tab:filtered-channels-detectors},
\texttt{Ro\-ta\-ted\-Fil\-ters} are significantly better than
the original \texttt{LDCF}, and only $1\ \mbox{pp}$ (percent point)
worse than \texttt{Che\-cker\-boards}, yet run $6\times$ faster
at training and test time.

\paragraph{Additional cues}

The review \cite{Benenson2014Eccvw} showed that context and optical
flow information can help improve detections. However, as the detector
quality improves (table \ref{tab:filtered-channels-detectors}) the
returns obtained from these additional cues erodes (table \ref{tab:add-ons-returns}).
Without re-engineering such cues, gains in detection must come from
the core detector.

\section{\label{sec:Analysing}Analysing the state of the art}

In this section we estimate a lower bound on the remaining progress
available, analyse the mistakes of current pedestrian detectors, and
propose new annotations to better measure future progress.

\subsection{\label{sec:Reaching-saturation}Are we reaching saturation?}

Progress on pedestrian detection has been showing no sign of slowing
in recent years \cite{Zhang2015Cvpr,Yang2015Cvpr,Benenson2014Eccvw},
despite recent impressive gains in performance. How much progress
can still be expected on current benchmarks? To answer this question,
we propose to use a human baseline as lower bound. We asked domain
experts to manually ``detect'' pedestrians in the Caltech-USA test
set; machine detection algorithms should be able to at least reach
human performance and, eventually, superhuman performance.

\paragraph{Human baseline protocol}

To ensure a fair comparison with existing detectors, most of which
operate at test time over a single image, we focus on the single frame
monocular detection setting. Frames are presented to annotators in
random order, and without access to surrounding frames from the source
videos. Annotators have to rely on pedestrian appearance and single-frame
context rather than (long-term) motion cues.

\begin{figure}
\begin{centering}
\includegraphics[width=1\columnwidth]{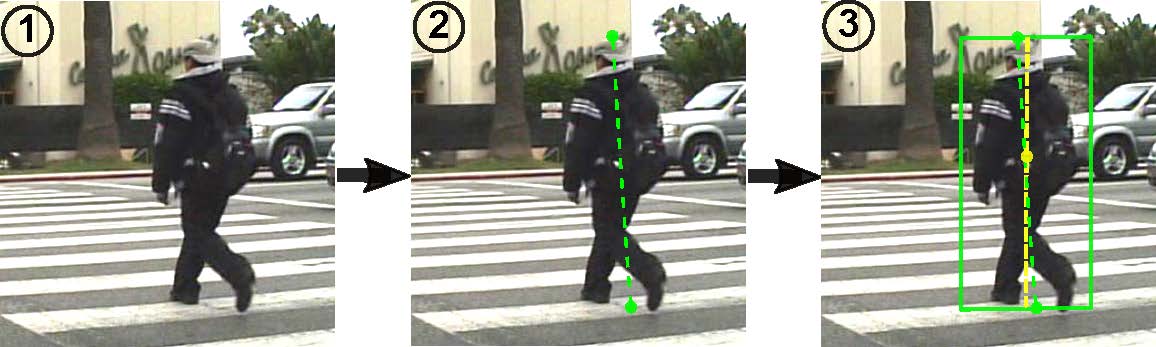}
\par\end{centering}

\caption{\label{fig:Illustration-of-bb-drawing}Illustration of bounding box
generation for human baseline. The annotator only needs to draw a
line from the top of the head to the central point between both feet,
a tight bounding box is then automatically generated.}
\end{figure}

The Caltech benchmark normalises the aspect ratio of all detection
boxes \cite{Dollar2012Pami}. Thus our human annotations are done
by drawing a line from the top of the head to the point between both
feet. A bounding box is then automatically generated such that its
centre coincides with the centre point of the manually-drawn axis,
see illustration in figure \ref{fig:Illustration-of-bb-drawing}.
This procedure ensures the box is well centred on the subject (which
is hard to achieve when marking a bounding box).

To check for consistency among the two annotators, we produced duplicate
annotations for a subset of the test images ($\sim10\%$), and evaluated
these separately. With a $\mbox{Intersection over Union (IoU)}\geq0.5$
matching criterion, the results were identical up to a single bounding
box.

\paragraph{Conclusion}

\begin{figure*}
\begin{centering}
\hspace*{-0.5em}\includegraphics[width=0.9\textwidth]{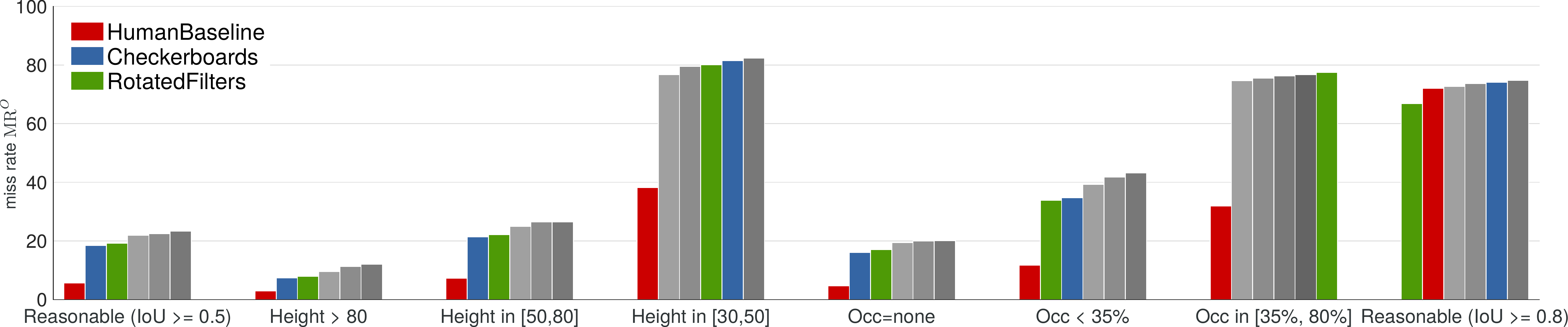}
\par\end{centering}

\caption{\label{fig:subsets-bar-plot}Detection quality (log-average miss rate)
for different test set subsets. Each group shows the human baseline,
the \texttt{Checkerboards} \cite{Zhang2015Cvpr} and \texttt{RotatedFilters}
detectors, as well as the next top three (unspecified) methods (different
for each setting). The corresponding curves are provided in the supplementary
material.}

\vspace{-1.5em}
\end{figure*}

In figure \ref{fig:subsets-bar-plot}, we compare our human baseline
with other top performing methods on different subsets of the test
data. We find that the human baseline widely outperforms state-of-the-art
detectors in all settings\footnote{Except for $\mbox{IoU}\geq0.8$. This is due to issues with the ground
truth, discussed in section \ref{sec:New-annotations}.}, indicating that there is still room for improvement for automatic
methods.

\subsection{\label{sec:Errors-analysis}Failure analysis}

Since there is room to grow for existing detectors, one might want
to know: when do they fail? In this section we analyse detection mistakes
of \texttt{Checkerboards}, which obtains top performance on most subsets
of the test set (see figure \ref{fig:subsets-bar-plot}). Since most
top methods of figure \ref{fig:cpvr15-top-results} are of the ICF
family, we expect a similar behaviour for them too. Methods using
convnets with proposals based on ICF detectors will also be affected.

\subsubsection{\label{sec:Error-sources}Error sources}

There are two types of errors a detector can do: false positives (detections
on background or poorly localised detections) and false negatives
(low-scoring or missing pedestrian detections). In this analysis,
we look into false positive and false negative detections at 0.1 false
positives per image (FPPI, 1 false positive every 10 images), and
manually cluster them (one to one mapping) into visually distinctive
groups. A total of 402 false positive and 148 false negative detections
(missing recall) are categorised by error type.

\paragraph{False positives}

\begin{figure}
\begin{centering}
\subfloat[\label{fig:False-positive-sources}False positive sources]{\begin{centering}
\includegraphics[width=1\columnwidth]{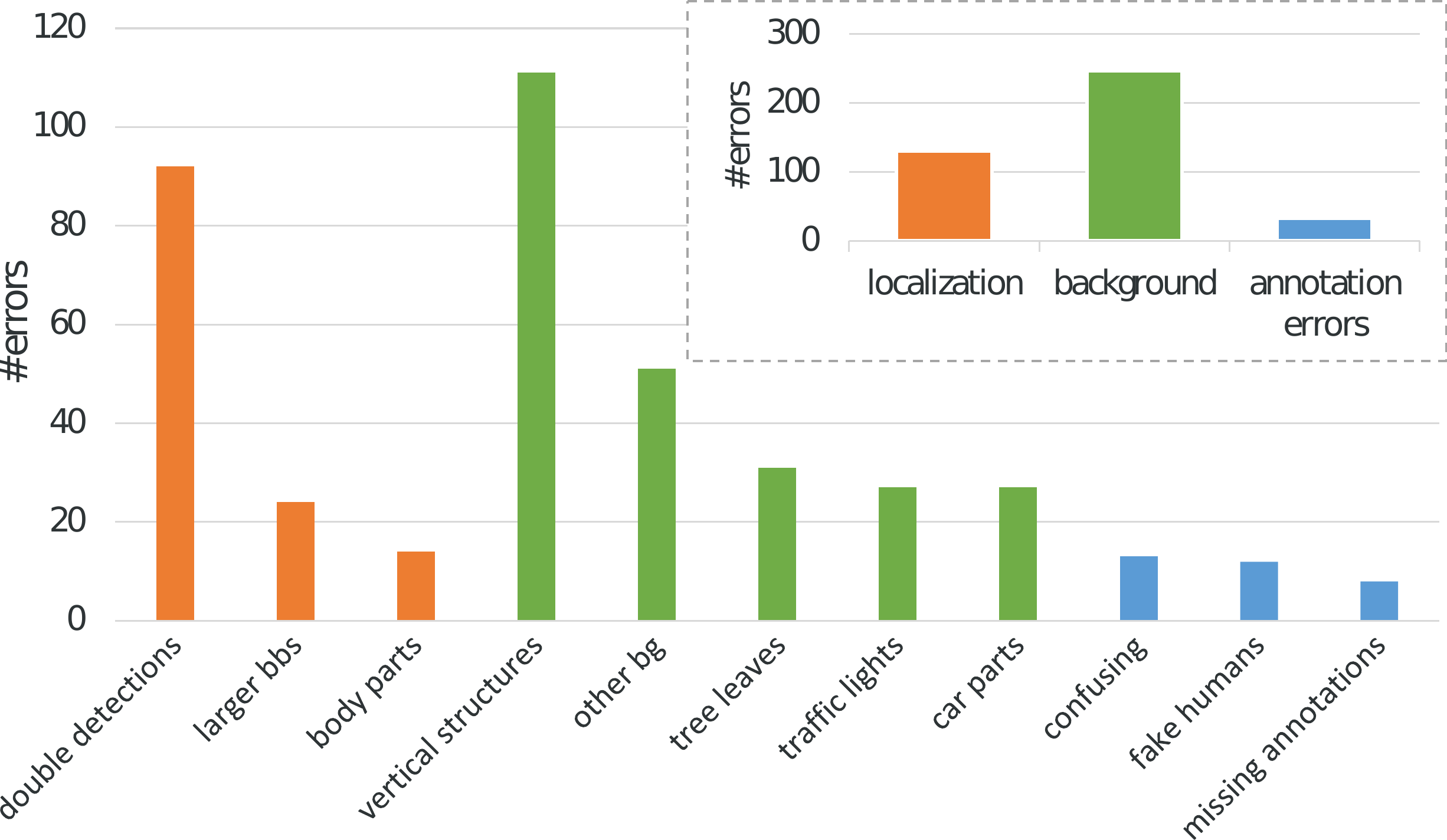}
\par\end{centering}

}\vspace{-1em}

\par\end{centering}

\begin{centering}
\subfloat[\label{fig:False-negative-sources}False negative sources]{\begin{centering}
\includegraphics[width=0.85\columnwidth]{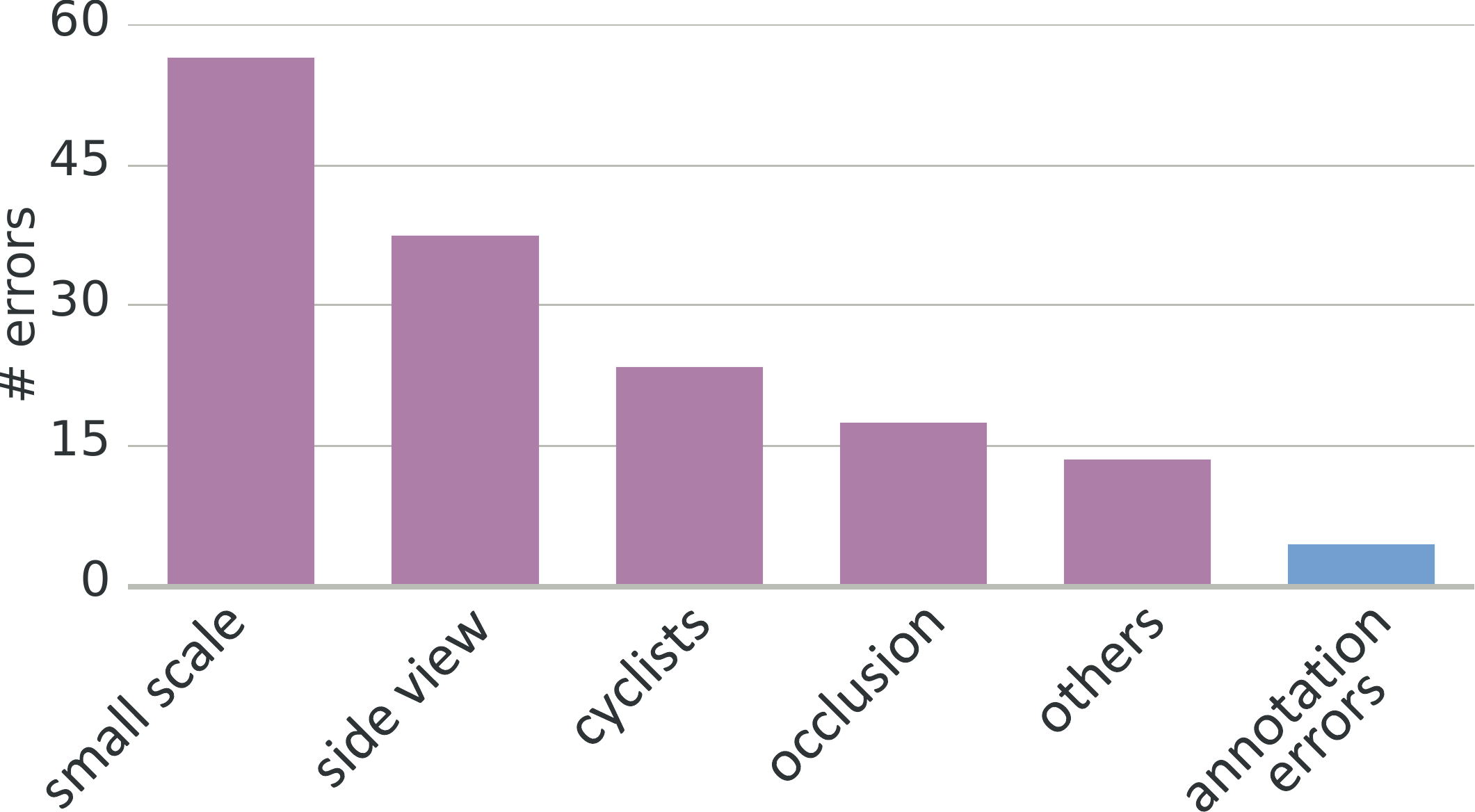}
\par\end{centering}

}
\par\end{centering}

\begin{centering}
\vspace{-1em}

\par\end{centering}

\begin{centering}
\subfloat[\label{fig:Contrast-vs-score}Contrast versus detection score]{\begin{centering}
\includegraphics[width=0.8\columnwidth,height=0.55\columnwidth]{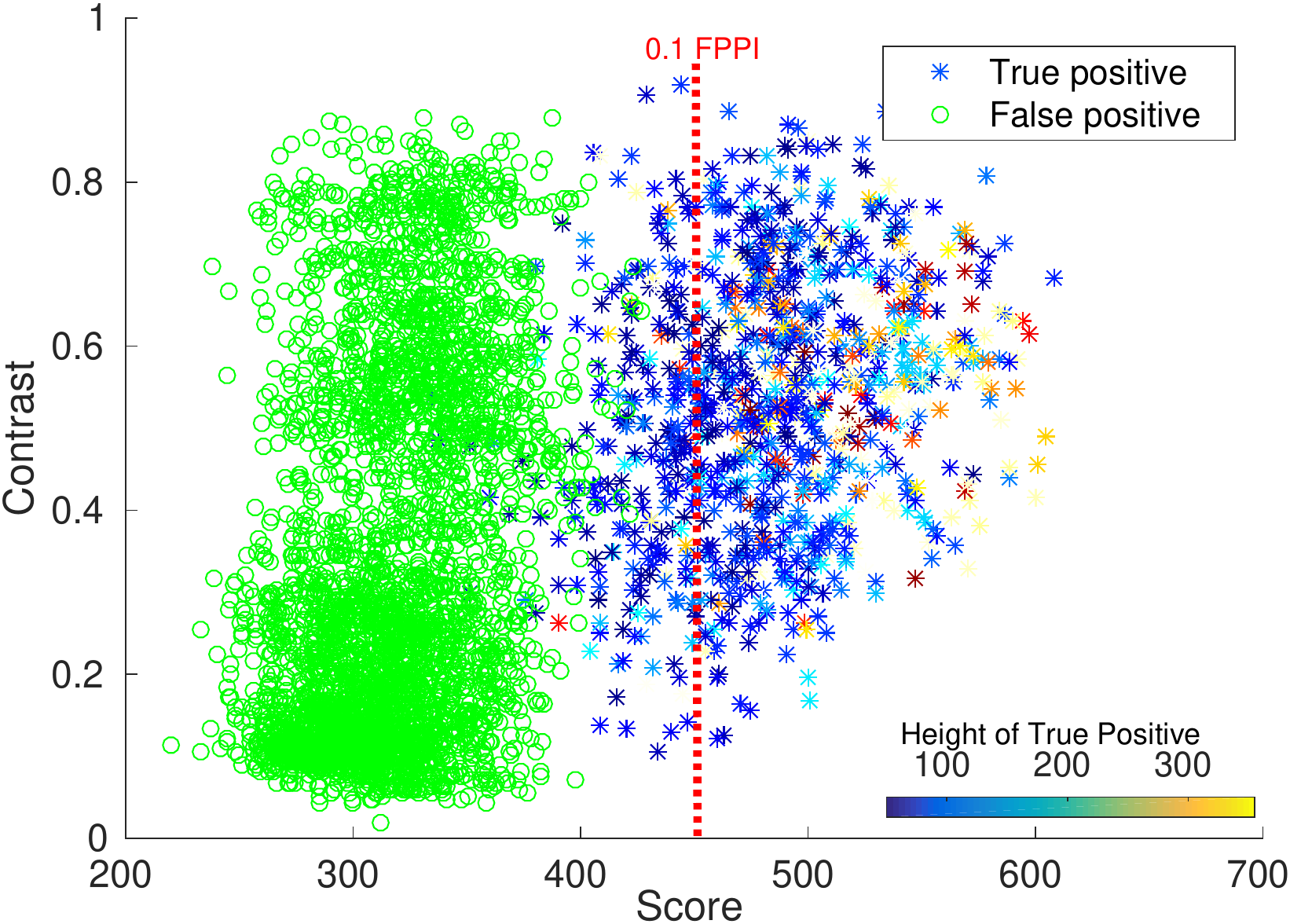}
\par\end{centering}

}
\par\end{centering}

\caption{\label{fig:errors-analysis-plots}Errors analysis of \texttt{Checkerboards}
\cite{Zhang2015Cvpr} on the test set. }
\end{figure}

\begin{figure}
\begin{centering}
\vspace{-1em}
\subfloat[\label{fig:example-error-double-detection}double detection]{\begin{centering}
\includegraphics[width=0.4\columnwidth]{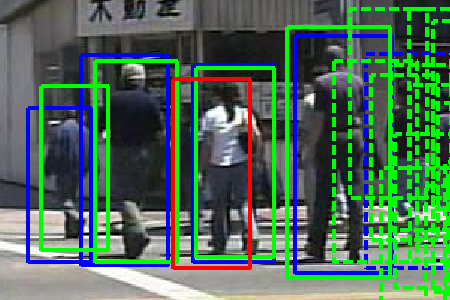}\vspace{-0.5em}

\par\end{centering}

} \subfloat[\label{fig:example-error-vertical-structure}vertical structure]{\begin{centering}
\includegraphics[bb=0bp 0bp 305bp 205bp,width=0.4\columnwidth]{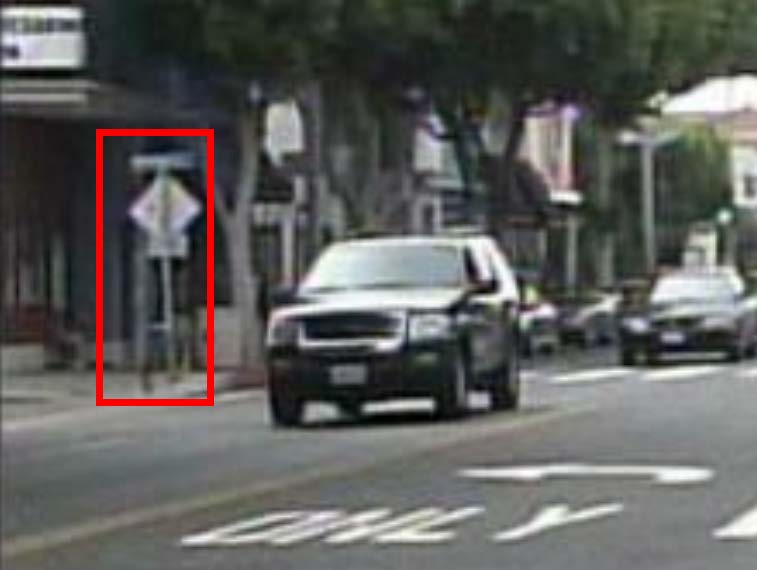}\vspace{-0.5em}

\par\end{centering}

} \vspace{-0.5em}

\par\end{centering}

\caption{\label{fig:errors-examples}Example of analysed false positive cases
(red box). Additional ones in supplementary material.}
\end{figure}

After inspection, we end up having all false positives clustered in
eleven categories, shown in figure \ref{fig:False-positive-sources}.
These categories fall into three groups: localisation, background,
and annotation errors. Localisation errors are defined as false detections
overlapping with ground truth bounding boxes, while background errors
have zero overlap with any ground truth annotation.\\
Background errors are the most common ones, mainly vertical structures
(e.g. figure \ref{fig:example-error-vertical-structure}), tree leaves,
and traffic lights. This indicates that the detectors need to be extended
with a better \emph{vertical context}, providing visibility over larger
structures and a rough height estimate.\\
Localisation errors are dominated by double detections (high scoring
detections covering the same person, e.g. figure \ref{fig:example-error-double-detection}).
This indicates that improved detectors need to have more localised
responses (peakier score maps) and/or a different non-maxima suppression
strategy. In sections \ref{sec:New-annotations} and \ref{sec:Annotations-impact}
we explore how to improve the detector localisation.\\
The annotation errors are mainly missing ignore regions, and a few
missing person annotations. In section \ref{sec:New-annotations}
we revisit the Caltech annotations.

\paragraph{False negatives}

Our clustering results in figure \ref{fig:False-negative-sources}
show the well known difficulty of detecting small and occluded objects.
We hypothesise that low scoring side-view persons and cyclists may
be due to a dataset bias, i.e. these cases are under-represented in
the training set (most persons are non-cyclist walking on the side-walk,
parallel to the car). Augmenting the training set with external images
for these cases might be an effective strategy.\\
To understand better the issue with small pedestrians, we measure
size, blur, and contrast for each (true or false) detection. We observed
that small persons are commonly saturated (over or under exposed)
and blurry, and thus hypothesised that this might be an underlying
factor for weak detection (other than simply having fewer pixels to
make the decision). Our results indicate however that this is not
the case. As figure \ref{fig:Contrast-vs-score} illustrates, there
seems to be no correlation between low detection score and low contrast.
This also holds for the blur case, detailed plots are in the supplementary
material. We conclude that the small number of pixels is the true
source of difficulty. Improving small objects detection thus need
to rely on making proper use of all pixels available, both inside
the window and in the surrounding context, as well as across time.

\paragraph{Conclusion}

Our analysis shows that false positive errors have well defined sources
that can be specifically targeted with the strategies suggested above.
A fraction of the false negatives are also addressable, albeit the
small and occluded pedestrians remain a (hard and) significant problem.

\subsubsection{\label{sec:Oracle-tests}Oracle test cases}

The analysis of section \ref{sec:Error-sources} focused on errors
counts. For area-under-the-curve metrics, such as the ones used in
Caltech, high-scoring errors matter more than low-scoring ones. In
this section we directly measure the impact of localisation and background-vs-foreground
errors on the detection quality metric (log-average miss-rate) by
using oracle test cases.

In the oracle case for localisation, all false positives that overlap
with ground truth are ignored for evaluation. In the oracle tests
for background-vs-foreground, all false positives that do not overlap
with ground truth are ignored.

\begin{figure}
\begin{centering}
\vspace{-1em}
\subfloat[\label{fig:oracle-gain-checkerboards}Original and two oracle curves
for \texttt{Checkerboards} detector. Legend indicates \emph{$\mbox{MR}_{-2}^{O}\left(\mbox{MR}_{-4}^{O}\right)$.} ]{\begin{centering}
\includegraphics[width=0.8\columnwidth]{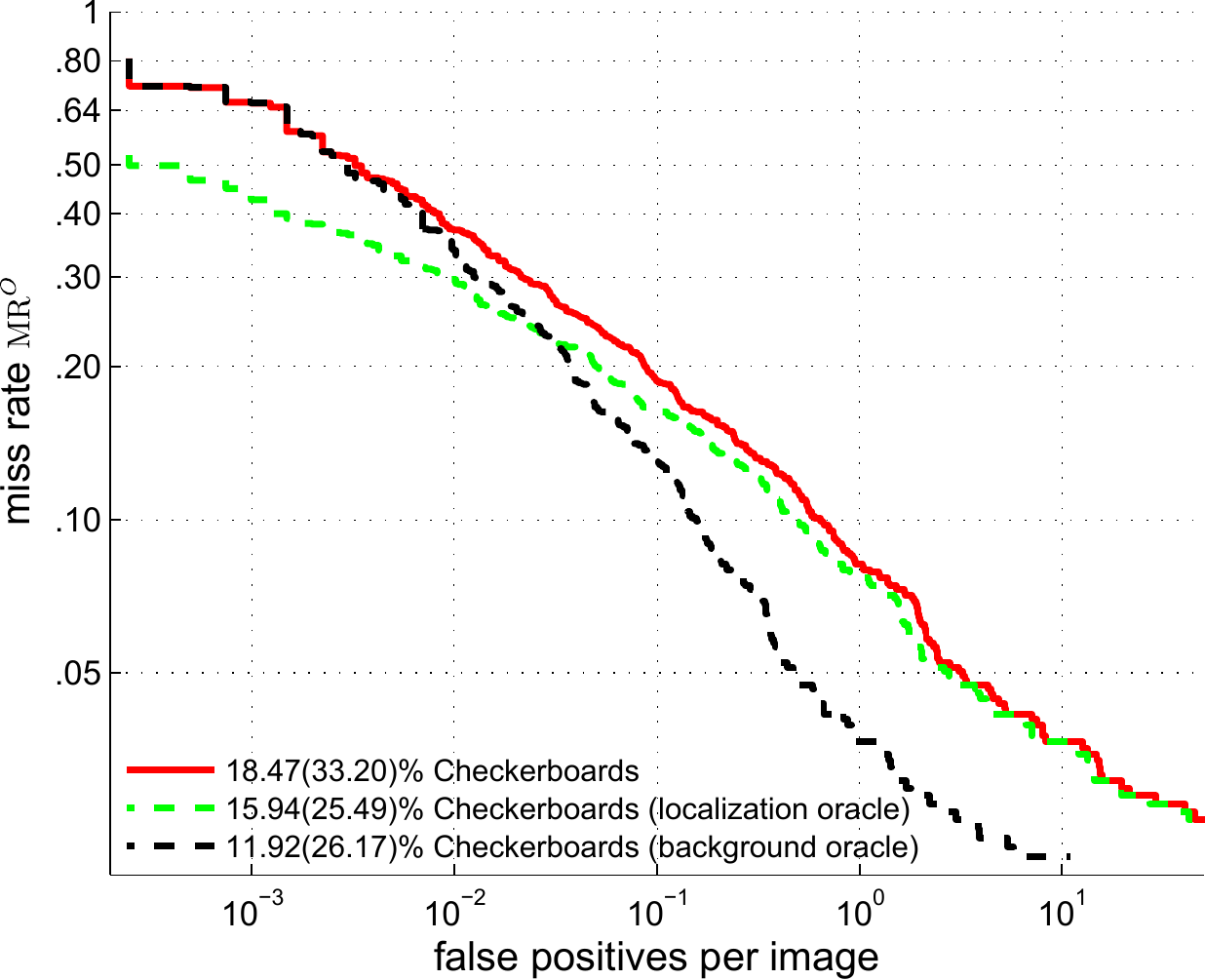}
\par\end{centering}

\centering{}}
\par\end{centering}

\begin{centering}
\subfloat[\label{fig:oracle-gain-all-methods}Comparison of miss-rate gain ($\Delta\mbox{MR}_{-4}^{O}$)
for top performing methods.]{\begin{centering}
\includegraphics[width=1\columnwidth]{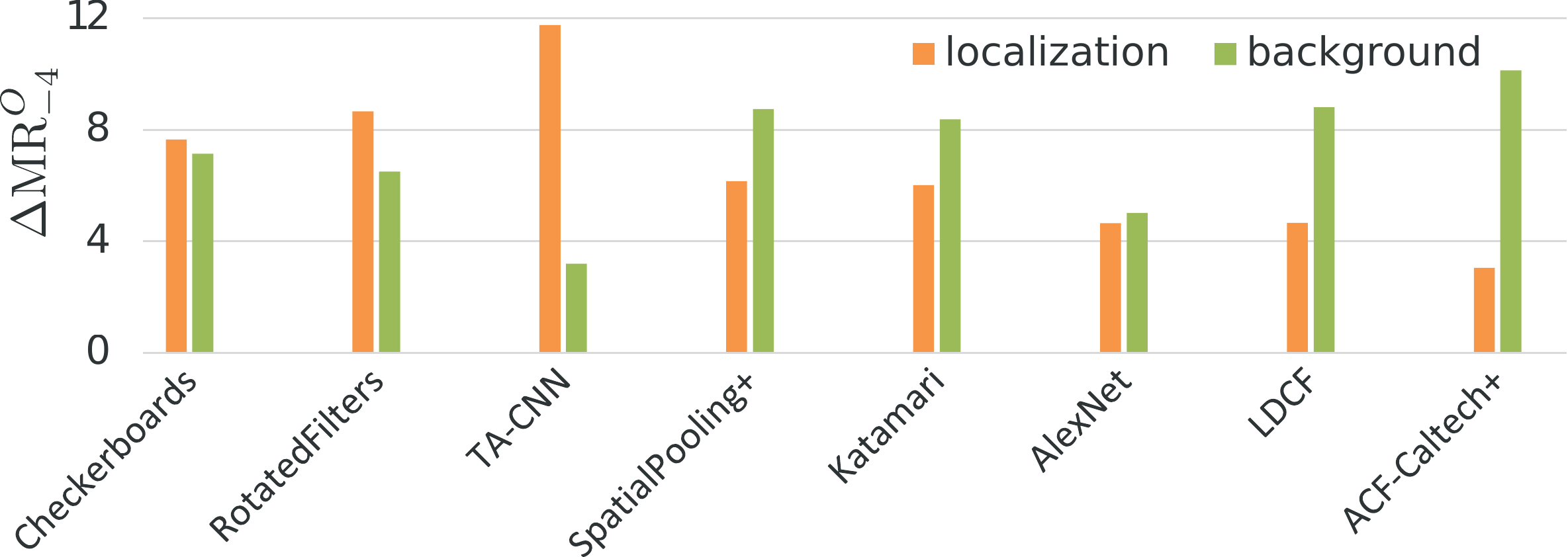}
\par\end{centering}

}
\par\end{centering}

\caption{\label{fig:oracle-cases}Oracle cases evaluation over Caltech test
set. Both localisation and background-versus-foreground show important
room for improvement.}
\end{figure}

Figure \ref{fig:oracle-gain-checkerboards} shows that fixing localisation
mistakes improves performance in the low FPPI region; while fixing
background mistakes improves results in the high FPPI region. Fixing
both types of mistakes results zero errors, even though this is not
immediately visible in the double log plot. 

In figure \ref{fig:oracle-gain-all-methods} we show the gains to
be obtained in \emph{$\mbox{MR}_{-4}^{O}$} terms by fixing localisation
or background issues. When comparing the eight top performing methods
we find that most methods would boost performance significantly by
fixing either problem. Note that due to the log-log nature of the
numbers, the sum of localisation and background deltas do not add
up to the total miss-rate.

\paragraph{Conclusion}

For most top performing methods localisation and background-vs-foreground
errors have equal impact on the detection quality. They are equally
important.

\subsection{\label{sec:New-annotations}Improved Caltech-USA annotations}

\begin{figure}
\begin{centering}
\vspace{-2em}
\hspace*{\fill}\subfloat[\label{fig:False-annotations}False annotations]{\begin{centering}
\includegraphics[height=0.25\columnwidth]{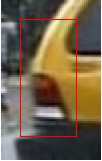}\includegraphics[height=0.25\columnwidth]{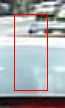}\vspace{-0.5em}

\par\end{centering}

\centering{}}\hspace*{\fill}\subfloat[\label{fig:Poor-alignment}Poor alignment]{\begin{centering}
\includegraphics[bb=120bp 0bp 270bp 180bp,clip,height=0.3\columnwidth]{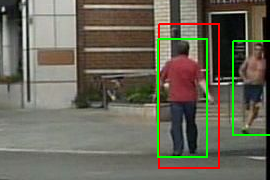}\hspace*{-2em}\includegraphics[bb=150bp 0bp 360bp 240bp,clip,height=0.3\columnwidth]{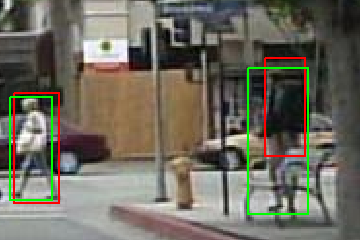}\hspace*{-2em}\includegraphics[bb=120bp 0bp 360bp 240bp,clip,height=0.3\columnwidth]{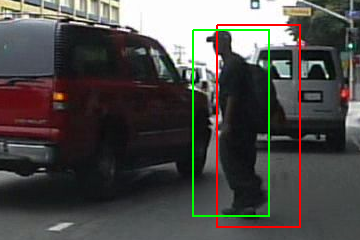}\hspace*{-3em}
\par\end{centering}

}\hspace*{\fill}
\par\end{centering}

\caption{\label{fig:improved-annotations}\label{fig:annotation-example-alignement}Examples
of errors in original annotations. New annotations in green, original
ones in red. }
\vspace{-1em}
\end{figure}
When evaluating our human baseline (and other methods) with a strict
$\mbox{IoU}\geq0.8$ we notice in figure \ref{fig:subsets-bar-plot}
that the performance drops. The original annotation protocol is based
on interpolating sparse annotations across multiple frames \cite{Dollar2012Pami},
and these sparse annotations are not necessarily located on the evaluated
frames. After close inspection we notice that this interpolation generates
a systematic offset in the annotations. Humans walk with a natural
up and down oscillation that is not modelled by the linear interpolation
used, thus in most frames have shifted bounding box annotations. This
effect is not noticeable when using the forgiving $\mbox{IoU}\geq0.5$,
however such noise in the annotations is a hurdle when aiming to improve
object localisation.

This localisation issues together with the annotation errors detected
in section \ref{sec:Error-sources} motivated us to create a new set
of improved annotations for the Caltech pedestrians dataset. Our aim
is two fold; on one side we want to provide a more accurate evaluation
of the state of the art, in particular an evaluation suitable to close
the ``last 20\%'' of the problem. On the other side, we want to
have training annotations and evaluate how much improved annotations
lead to better detections. We evaluate this second aspect in section
\ref{sec:Annotations-impact}.

\paragraph{New annotation protocol}

Our new annotations are done both on the test and training $1\times$
set, and focus on high quality. The annotators are allowed to look
at the full video to decide if a person is present or not, they are
requested to mark ignore regions in areas covering crowds, human shapes
that are not persons (posters, statues, etc.), and in areas that could
not be decided as certainly not containing a person. Each person annotation
is done by drawing a line from the top of the head to the point between
both feet, the same as human baseline. The annotators must hallucinate
head and feet if these are not visible. When the person is not fully
visible, they must also annotate a rectangle around the largest visible
region. This allows to estimate the occlusion level in a similar
fashion as the original annotations. The new annotations do share
some bounding boxes with the human baseline (when no correction was
needed), thus the human baseline cannot be used to do analysis across
different $\mbox{IoU}$ thresholds over the new test set.\\
In summary, our new annotations differ from the human baseline in
the following aspects: both training and test sets are annotated,
ignore regions and occlusions are also annotated, full video data
is used for decision, and multiple revisions of the same image are
allowed.\\
After creating a full independent set of annotations, we consolidated
the new annotations by cross-validating with the old annotations.
Any correct old annotation not accounted for in the new set, was added
too.

Our new annotations correct several types of errors in the existing
annotations, such as misalignments (figure \ref{fig:Poor-alignment}),
missing annotations (false negatives), false annotations (false positives,
figure \ref{fig:False-annotations}), and the inconsistent use of
\textquotedblleft ignore\textquotedblright{} regions. More examples
of ``original versus new annotations'' provided in the supplementary
material, as well as a visualisation software to inspect them frame
by frame.

\paragraph{Better alignment}

In table \ref{tab:median-iou} we show quantitative evidence that
our new annotations are at least more precisely localised than the
original ones. We summarise the alignment quality of a detector via
the median $\mbox{IoU}$ between true positive detections and a given
set of annotations. When evaluating with the original annotations
(``median $\mbox{IoU}^{O}$'' column in table \ref{tab:median-iou}),
only the model trained with original annotations has good localisation.
However, when evaluating with the new annotations (``median $\mbox{IoU}^{N}$''
column) \emph{both} the model trained on INRIA data, and on the new
annotations reach high localisation accuracy. This indicates that
our new annotations are indeed better aligned, just as INRIA annotations
are better aligned than Caltech.%

Detailed $\mbox{IoU}$ curves for multiple detectors are provided
in the supplementary material. Section \ref{sec:Annotations-impact}
describes the \texttt{Ro\-ta\-ted\-Fil\-ters\--New\-10$\mathtt{\times}$}
entry.

\begin{table}
\vspace{-0em}

\begin{centering}
\setlength{\tabcolsep}{4pt} 
\begin{tabular}{ll|cc}
Detector & %
\begin{tabular}{c}
Training\tabularnewline
data\tabularnewline
\end{tabular} & %
\begin{tabular}{c}
Median\tabularnewline
$\mbox{IoU}^{O}$\tabularnewline
\end{tabular} & %
\begin{tabular}{c}
Median\tabularnewline
$\mbox{IoU}^{N}$\tabularnewline
\end{tabular}\tabularnewline
\hline 
\hline 
Roerei \cite{Benenson2013Cvpr} & INRIA & 0.76 & \emph{0.84}\tabularnewline
RotatedFilters & Orig. $10\times$ & \emph{0.80} & 0.77\tabularnewline
RotatedFilters & New $10\times$ & 0.76 & \emph{0.85}\tabularnewline
\end{tabular}
\par\end{centering}

\caption{\label{tab:median-iou}Median IoU of true positives for detectors
trained on different data, evaluated on original and new Caltech test.
Models trained on INRIA align well with our new annotations, confirming
that they are more precise than previous ones. Curves for other detectors
in the supplement. }
\vspace{-0em}
\end{table}

\section{\label{sec:Improving}Improving the state of the art}

In this section we leverage the insights of the analysis, to improve
localisation and background-versus-foreground discrimination of our
baseline detector.

\subsection{\label{sec:Annotations-impact}Impact of training annotations}

With new annotations at hand we want to understand what is the impact
of annotation quality on detection quality. We will train \texttt{ACF
}\cite{Dollar2014Pami}\texttt{ }and \texttt{Rotated\-Filters} models
(introduced in section \ref{sec:FltrChnFtrs}) using different training
sets and evaluate on both original and new annotations (i.e. \emph{$\mbox{MR}_{-2}^{O}$,
$\mbox{MR}_{-4}^{O}$} and \emph{$\mbox{MR}_{-2}^{N}$, $\mbox{MR}_{-4}^{N}$}).
Note that both detectors are trained via boosting and thus inherently
sensitive to annotation noise.

\paragraph{Pruning benefits}

\begin{table}
\begin{centering}
\begin{tabular}{lc|c|c}
\multirow{1}{*}{Detector} & Anno. variant & $\mbox{MR}_{-2}^{O}$ & $\mbox{MR}_{-2}^{N}$\tabularnewline
\hline 
\hline 
\multirow{3}{*}{ACF} & Original & \emph{36.90} & 40.97\tabularnewline
 & Pruned & 36.41 & 35.62\tabularnewline
 & New & 41.29 & \emph{34.33}\tabularnewline
\hline 
\multirow{3}{*}{RotatedFilters} & Original & \emph{28.63} & 33.03\tabularnewline
 & Pruned & 23.87 & 25.91\tabularnewline
 & New & 31.65 & \emph{25.74}\tabularnewline
\end{tabular}
\par\end{centering}

\begin{centering}

\par\end{centering}

\caption{\label{tab:original-new-prunned-validation}Effects of different training
annotations on detection quality on validation set ($1\times$ training
set). Italic numbers have matching training and test sets. Both detectors
improve on the original annotations, when using the ``pruned'' variant
(see \S\ref{sec:Annotations-impact}).}
\end{table}
Table \ref{tab:original-new-prunned-validation} shows results when
training with original, new and pruned annotations (using a $\nicefrac{5}{6}+\nicefrac{1}{6}$
training and validation split of the full training set). As expected,
models trained on original/new and tested on original/new perform
better than training and testing on different annotations. To understand
better what the new annotations bring to the table, we build a hybrid
set of annotations. Pruned annotations is a mid-point that allows
to decouple the effects of removing errors and improving alignment.
\\
Pruned annotations are generated by matching new and original annotations
($\mbox{IoU}\geq0.5$), marking as ignore region any original annotation
absent in the new ones, and adding any new annotation absent in the
original ones.\\
From original to pruned annotations the main change is removing annotation
errors, from pruned to new, the main change is better alignment. From
table \ref{tab:original-new-prunned-validation} both \texttt{ACF
}and \texttt{Rotated\-Filters} benefit from removing annotation errors,
even in \emph{$\mbox{MR}_{-2}^{O}$}. This indicates that our new
training set is better sanitised than the original one.\\
We see in \emph{$\mbox{MR}_{-2}^{N}$} that the stronger detector
benefits more from better data, and that the largest gain in detection
quality comes from removing annotation errors.%

\paragraph{Alignment benefits}

The detectors from the ICF family benefit from training with increased
training data \cite{Nam2014arXiv,Zhang2015Cvpr}, using $10\times$
data is better than $1\times$ (see section \ref{sec:Pedestrian-datasets}).
To leverage the $9\times$ remaining data using the new $1\times$
annotations we train a model over the new annotations and use this
model to re-align the original annotations over the $9\times$ portion.
Because the new annotations are better aligned, we expect this model
to be able to recover slight position and scale errors in the original
annotations. Figure \ref{fig:automatic-alignment} shows example results
of this process. See supplementary material for details.\\
Table \ref{tab:alignment-experiments} reports results using the automatic
alignment process, and a few degraded cases: using the original $10\times$,
self-aligning the original $10\times$ using a model trained over
original $10\times$, and aligning the original $10\times$ using
only a fraction of the new annotations (without replacing the $1\times$
portion). The results indicate that using a detector model to improve
overall data alignment is indeed effective, and that better aligned
training data leads to better detection quality (both in $\mbox{MR}^{O}$
and $\mbox{MR}^{N}$). This is in line with the analysis of section
\ref{sec:Errors-analysis}. Already using a model trained on $\nicefrac{1}{2}$
of the new annotations for alignment, leads to a stronger model than
obtained when using original annotations.\\
We name the \texttt{Rotated\-Filters }model trained using the new
annotations and the aligned $9\times$ data, \texttt{Rotated\-Filters\--New10$\mathtt{\times}$}.
This model also reaches high median true positives IoU in table \ref{tab:median-iou},
indicating that indeed it obtains more precise detections at test
time.
\begin{figure}
\begin{centering}
\includegraphics[width=1\columnwidth]{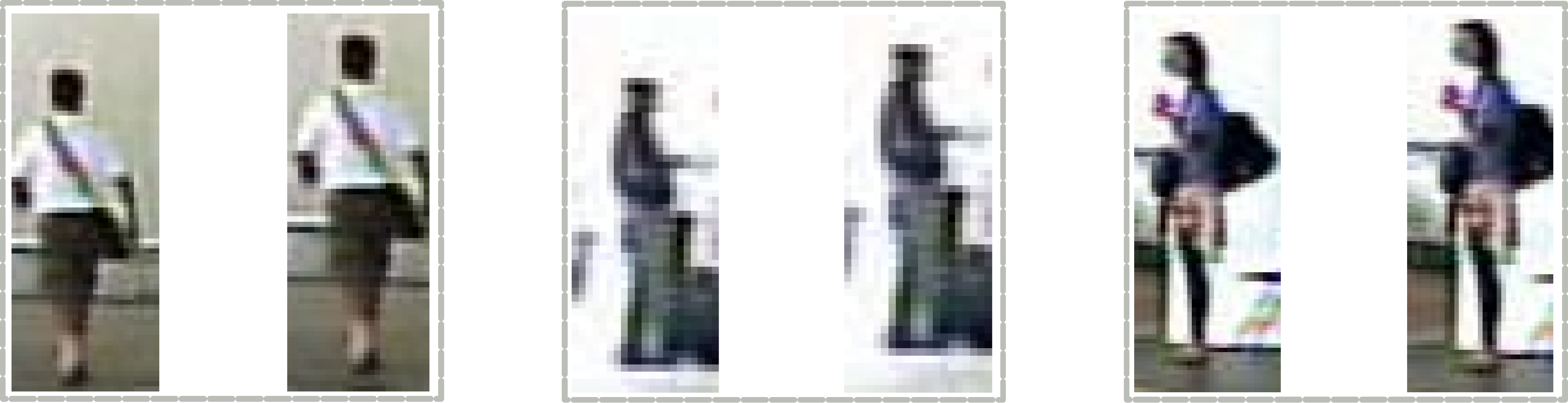}
\par\end{centering}

\caption{\label{fig:automatic-alignment}Examples of automatically aligned
ground truth annotations. Left/right$\rightarrow$ before/after alignment.}
\end{figure}
\begin{table}
\begin{centering}
\setlength{\tabcolsep}{4pt} 
\begin{tabular}{cc|c|c}
\begin{tabular}{c}
$1\times$\tabularnewline
data\tabularnewline
\end{tabular} & %
\begin{tabular}{c}
$10\times$ data\tabularnewline
aligned with\tabularnewline
\end{tabular} & $\mbox{MR}_{-2}^{O}$ ($\mbox{MR}_{-4}^{O}$) & $\mbox{MR}_{-2}^{N}$ ($\mbox{MR}_{-4}^{N}$)\tabularnewline
\hline 
\hline 
Orig. & $\textrm{Ø}$ & 19.20 (34.28) & 17.22 (31.65)\tabularnewline
\hline 
Orig. & Orig. $10\times$ & 19.16 (32.28) & 15.71 (28.13)\tabularnewline
Orig. & New $\nicefrac{1}{2}\times$ & 16.97 (28.01) & 14.54 (25.06)\tabularnewline
\hline 
New & New $1\times$ & 16.77 (29.76) & 12.96 (22.20)\tabularnewline
\end{tabular}
\par\end{centering}

\caption{\label{tab:alignment-experiments}Detection quality of \texttt{Rotated\-Filters}
on test set when using different aligned training sets. All models
trained with Caltech $10\times$, composed with different $1\times\,+\,9\times$
combinations.}
\end{table}

\paragraph{Conclusion}

Using high quality annotations for training improves the overall detection
quality, thanks both to improved alignment and to reduced annotation
errors.

\subsection{\label{sec:ConvNets}Convnets for pedestrian detection}

The results of section \ref{sec:Errors-analysis} indicate that there
is room for improvement by focusing on the core background versus
foreground discrimination task (the ``classification part of object
detection''). Recent work \cite{Hosang2015Cvpr,Yang2015Cvpr} showed
competitive performance with convolutional neural networks (convnets)
for pedestrian detection. We include convnets into our analysis, and
explore to what extent performance is driven by the quality of the
detection proposals.

\paragraph{AlexNet and VGG}

We consider two convnets. 1) The AlexNet from \cite{Hosang2015Cvpr},
and 2) The VGG16 model from \cite{Girshick2015IccvFastRCNN}. Both
are pre-trained on ImageNet and fine-tuned over Caltech $10\times$
(original annotations) using \texttt{SquaresChnFtrs} proposals. Both
networks are based on open source, and both are instances of the R-CNN
framework \cite{Girshick2014Cvpr}. Albeit their training/test time
architectures are slightly different (R-CNN versus Fast R-CNN), we
expect the result differences to be dominated by their respective
discriminative power (VGG16 improves $8\ \mbox{pp}$ in mAP over AlexNet
in the Pascal detection task \cite{Girshick2014Cvpr}).

\begin{table}
\begin{centering}
{\footnotesize{}\hspace{-1em}}\definecolor{red}{RGB}{150, 150, 150}\setlength{\tabcolsep}{4pt} 
\begin{tabular}{lc|c|cc}
\multirow{2}{*}{Test proposals} & \multirow{2}{*}{{\footnotesize{}Proposal}} & \multirow{2}{*}{{\footnotesize{}+AlexNet}} & \multirow{2}{*}{{\footnotesize{}+VGG}} & {\footnotesize{}+bbox reg }\tabularnewline
 &  &  &  & {\footnotesize{}\& NMS}\tabularnewline
\hline 
\hline 
ACF \cite{Dollar2014Pami} & 48.0\% & 28.5\%  & 22.8\% & 20.8\%\tabularnewline
SquaresChnFtrs \cite{Benenson2014Eccvw}{\footnotesize{}\hspace{-1em}} & 31.5\% & 21.2\% & 15.9\% & 14.7\%\tabularnewline
LDCF \cite{Nam2014arXiv} & 23.7\% & 21.6\% & 16.0\% & 13.7\%\tabularnewline
Rot.Filters & 17.2\% & \textcolor{red}{21.5\%} & \textcolor{red}{17.8\%} & 13.8\%\tabularnewline
Checkerboards \cite{Zhang2015Cvpr} & 16.1\% & \textcolor{red}{21.0\%} & 15.3\% & 11.1\%\tabularnewline
Rot.Filters-New10$\times$ & 12.9\% & \textcolor{red}{17.2\%} & 11.7\% & 10.0\%\tabularnewline
\end{tabular}
\par\end{centering}

\begin{centering}
\definecolor{red}{RGB}{180, 0, 0}
\par\end{centering}

\caption{\label{tab:rcnn-proposals}Detection quality of convnets with different
proposals. Grey numbers indicate worse results than the input proposals.
All numbers are MR$_{-2}^{N}$ on the Caltech test set.}
\end{table}
\begin{figure}
\begin{centering}
\vspace{-1em}
\includegraphics[width=0.9\columnwidth]{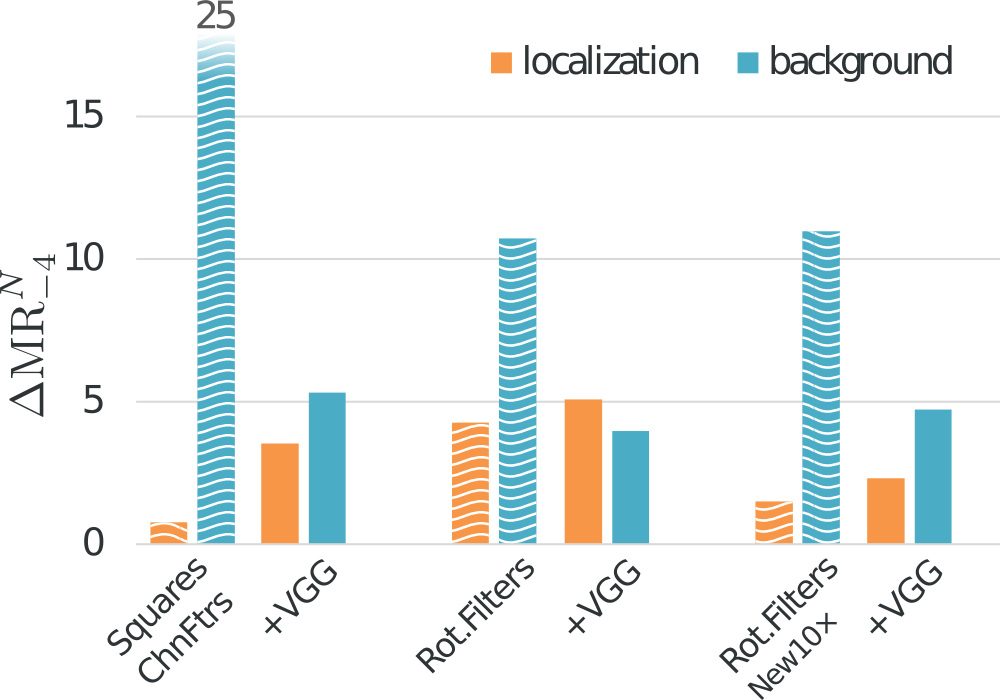}
\par\end{centering}

\vspace{-0.5em}
\caption{\label{fig:oracles-convnets}Oracle case analysis of proposals + convnets
(after second NMS). Miss-rate gain, $\Delta\mbox{MR}_{-4}^{O}$. The
convnet significantly improves background errors, while slightly increasing
localisation ones.}
\end{figure}
Table \ref{tab:rcnn-proposals} shows that as the quality of the detection
proposals improves, AlexNet fails to provide a consistent gain, eventually
worsening the results of our ICF detectors (similar observation in
\cite{Hosang2015Cvpr}). Similarly VGG provides large gains for weaker
proposals, but as the proposals improve, the gain from the convnet
re-scoring eventually stalls.

After closer inspection of the resulting curves (see supplementary
material), we notice that both AlexNet and VGG push background instances
to lower scores, and at the same time generate a large number of high
scoring false positives. The ICF detectors are able to provide high
recall proposals, where false positives around the objects have low
scores (see \cite[supp. material, fig. 9]{Hosang2015Cvpr}), however
convnets have difficulties giving low scores to these windows surrounding
the true positives. In other words, despite their fine-tuning, the
convnet score maps are ``blurrier'' than the proposal ones. We hypothesise
this is an intrinsic limitation of the AlexNet and VGG architectures,
due to their internal feature pooling. Obtaining ``peakier'' responses
from a convnet most likely will require using rather different architectures,
possibly more similar to the ones used for semantic labelling or boundaries
estimation tasks which require pixel-accurate output.

\begin{figure}
\begin{centering}
\includegraphics[width=1\columnwidth]{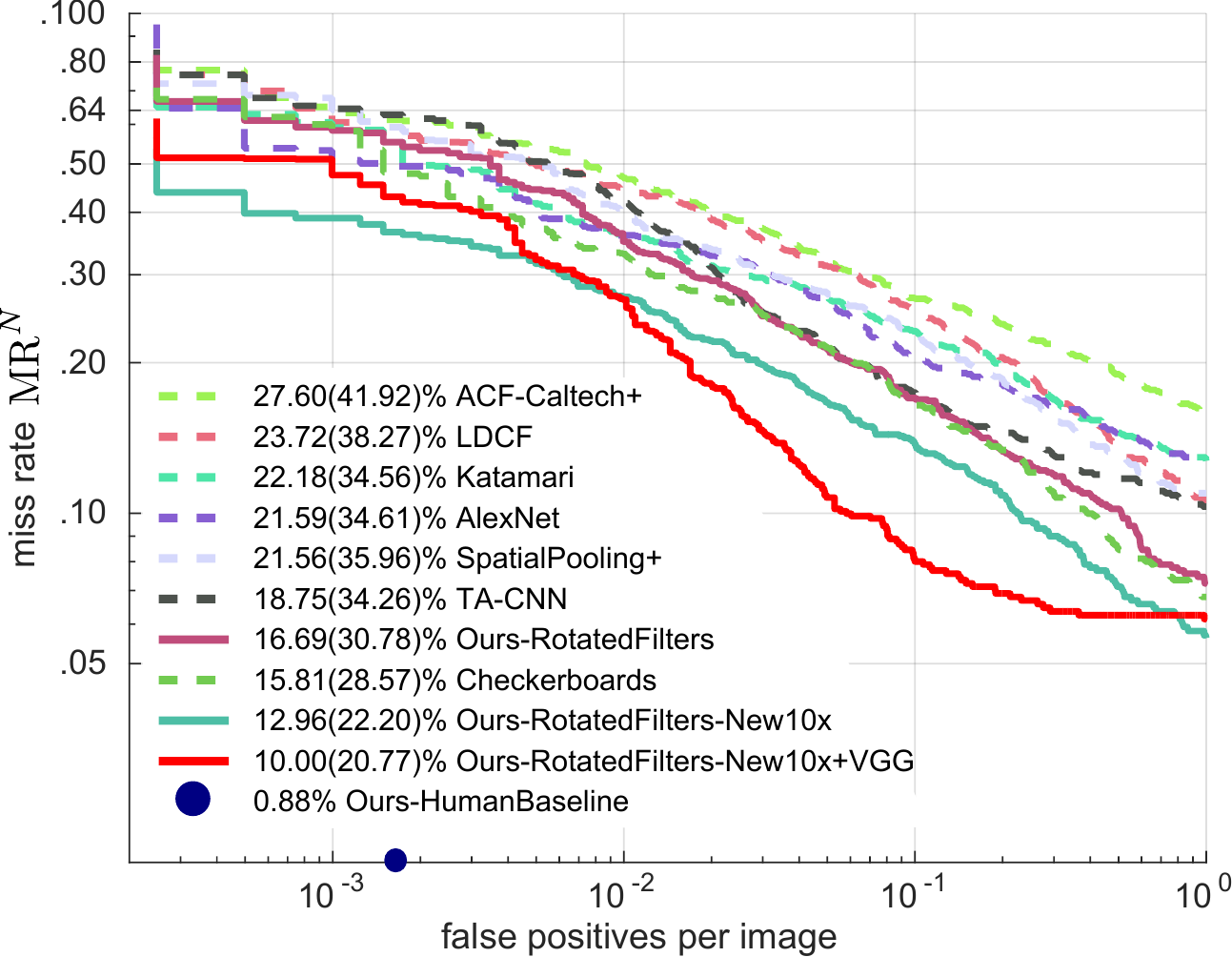}\vspace{-0em}

\par\end{centering}

\centering{}\caption{\label{fig:caltech-results}\label{fig:Caltech-results-new-annotations}Detection
quality on Caltech test set (reasonable subset), evaluated on the
new annotations (\emph{$\mbox{MR}_{-2}^{N}\left(\mbox{MR}_{-4}^{N}\right)$}).
Further results in the supplementary material.}
\vspace{-1em}
\end{figure}
\begin{table}
\begin{centering}
\setlength{\tabcolsep}{4pt} 
\begin{tabular}{c|c|c}
Detector aspect & $\mbox{MR}_{-2}^{O}$ ($\mbox{MR}_{-4}^{O}$) & $\mbox{MR}_{-2}^{N}$ ($\mbox{MR}_{-4}^{N}$)\tabularnewline
\hline 
\hline 
\texttt{Rotated\-Filters} & 19.20 (34.28) & 17.22 (31.65)\tabularnewline
+ Alignment \S\ref{sec:Annotations-impact} & 16.97 (28.01) & 14.54 (25.06)\tabularnewline
+ New annotations \S\ref{sec:Annotations-impact} & 16.77 (29.76) & 12.96 (22.20)\tabularnewline
+ VGG \S\ref{sec:ConvNets} & 16.61 (34.79) & 11.74 (28.37)\tabularnewline
+ bbox reg \& NMS & \emph{14.16 }(\emph{28.39}) & \emph{10.00 }(\emph{20.77})\tabularnewline
\hline 
\texttt{Checkerboards} & 18.47 (33.20) & 15.81 (28.57)\tabularnewline
\end{tabular}
\par\end{centering}

\caption{\label{tab:improvements-step-by-step}Step by step improvements from
previous best method \texttt{Checkerboards} to \texttt{Rotated\-Filters-New10x+VGG}.}
\vspace{-0.5em}
\end{table}
Fortunately, we can compensate for the lack of spatial resolution
in the convnet scoring by using bounding box regression. Adding bounding
regression over VGG, and applying a second round of non-maximum suppression
(first NMS on the proposals, second on the regressed boxes), has the
effect of ``contracting the score maps''. Neighbour proposals that
before generated multiple strong false positives, now collapse into
a single high scoring detection. We use the usual $\mbox{IoU}\geq0.5$
merging criterion for the second NMS.

The last column of table \ref{tab:rcnn-proposals} shows that bounding
box regression + NMS is effective at providing an additional gain
over the input proposals, even for our best detector \texttt{Rotated\-Filters\--New10$\mathtt{\times}$}.
On the original annotations \texttt{Rotated\-Filters\--New10$\mathtt{\times}$+VGG
}reaches $14.2\%\ \mbox{MR}_{-2}^{O}$ , which improves over \cite{Hosang2015Cvpr,Yang2015Cvpr}.
Our best performing detector \texttt{Rotated\-Filters\--New10$\mathtt{\times}$}
runs on a $640\times480$ image for \textasciitilde{}3.5 seconds,
including the ICF sliding window detection and VGG rescoring. Training
times are counted 1\textasciitilde{}2 days for the \texttt{Rotated\-Filters}
detector, and 1\textasciitilde{}2 days for VGG fine-tunning.

Figure \ref{fig:oracles-convnets} repeats the oracle tests of section
\ref{sec:Oracle-tests} over our convnet results. One can see that
VGG significantly cuts down the background errors, while at the same
time slightly increases the localisation errors.

\paragraph{Conclusion}

Although convnets have strong results in image classification and
general object detection, they seem to have limitations when producing
well localised detection scores around small objects. Bounding box
regression (and NMS) is a key ingredient to side-step this limitation
with current architectures. Even after using a strong convnet, background-versus-foreground
remains the main source of errors; suggesting that there is still
room for improvement on the raw classification power of the neural
network.%

\section{\label{sec:Conclusion}Summary}

In this paper, we make great efforts on analysing the failures
for a top-performing detector on Caltech dataset. Via our human baseline
we have quantified a lower bound on how much improvement there is
to be expected. There is a $10\times$ gap still to be closed. To
better measure the next steps in detection progress, we have provided
new sanitised Caltech train and test set annotations. 

Our failure analysis of a top performing method has shown that most
of its mistakes are well characterised. The error characteristics
lead to specific suggestions on how to engineer better detectors (mentioned
in section \ref{sec:Errors-analysis}; e.g. data augmentation for
person side views, or extending the detector receptive field in the
vertical axis). 

We have partially addressed some of the issues by measuring the impact
of better annotations on localisation accuracy, and by investigating
the use of convnets to improve the background to foreground discrimination.
Our results indicate that significantly better alignment can be achieved
with properly trained ICF detectors, and that, for pedestrian detection,
convnet struggle with localisation issues, that can be partially addressed
via bounding box regression. Both on original and new annotations,
the described detection approach reaches top performance, see progress
in table \ref{tab:improvements-step-by-step}.

We hope the insights and data provided in this work will guide the
path to close the gap between machines and humans in the pedestrian
detection task.

\bibliographystyle{plain}
\bibliography{How_far_are_we_from_solving_pedestrian_detection}

\clearpage{}

\appendix

\part*{Supplementary material}

\section{\label{sec:Content}Content}

This supplementary material provides a more detailed view of some
of the aspects presented in the main paper. 
\begin{itemize}
\item Section \ref{sec:Rotated-filters-detector} gives details of the \texttt{Rotated\-Filters}
detector we used for our experiments (section 2.2 in main paper).
\item Section \ref{sec:Subsets-results-supp} provides the detailed curves
behind the summary bar plots for different test set subsets (see figure
3 and section 3.1 in main paper).
\item Section \ref{sec:Errors-analysis-supp} shows examples for each error
type from the analysed detector, discusses the scale, blur and contrast
evaluations, and revisits the oracle cases experiments in more detail
(section 3.2 in main paper). 
\item Section \ref{sec:Improved-annotations-supp} shows examples of how
the new training annotations improve over the original ones (section
3.3 in main paper).
\item Section \ref{sec:Old-vs-new-ranking} discuss the impact of new annotations
on the evaluation of existing methods (MR ranking and recall-versus-IoU
curves) (section 4.1 in main paper).
\item Section \ref{sec:Impact-of-training-annotations-supp} shows the effects
of automatically aligning $10\times$ data with $1\times$data (section
4.1 in main paper).
\item Figure \ref{fig:final-curves} summarises our final detection results
both in original and new annotations.
\end{itemize}
\clearpage{}

\section{\label{sec:Rotated-filters-detector}Rotated filters detector}

\begin{figure}
\begin{centering}
\subfloat[\texttt{\label{fig:ACF-SquaresChnFtrs-filters}SquaresChntrs }{\small{}\cite{Benenson2014Eccvw}}
filters]{\centering{}\qquad{}\includegraphics[width=0.2\textwidth]{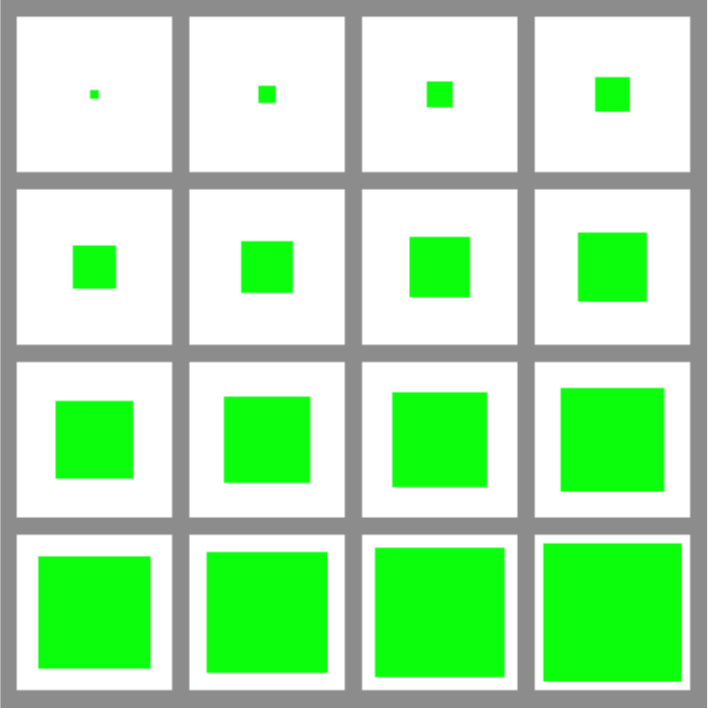}\qquad{}}
\par\end{centering}

\begin{centering}
\subfloat[\texttt{\label{fig:LDCF-filters}}Some of the \texttt{LDCF} \cite{Nam2014arXiv}
filters. Each column shows filters for one channel.]{\centering{}\qquad{}\includegraphics[width=0.2\textwidth]{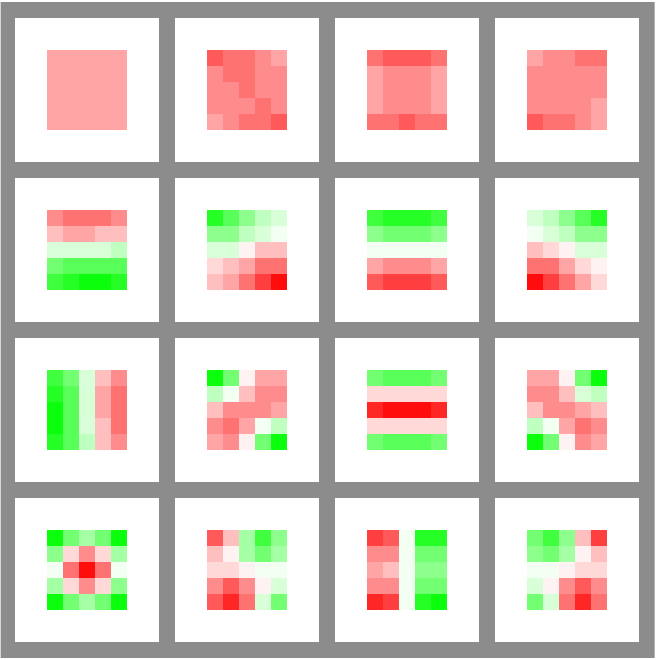}\qquad{}}
\par\end{centering}

\begin{centering}
\subfloat[\label{fig:checkerboard-filters}Some examples of the 61 Checkerboards
filters (from \cite{Zhang2015Cvpr})]{\begin{centering}
\qquad{}\includegraphics[width=0.8\columnwidth]{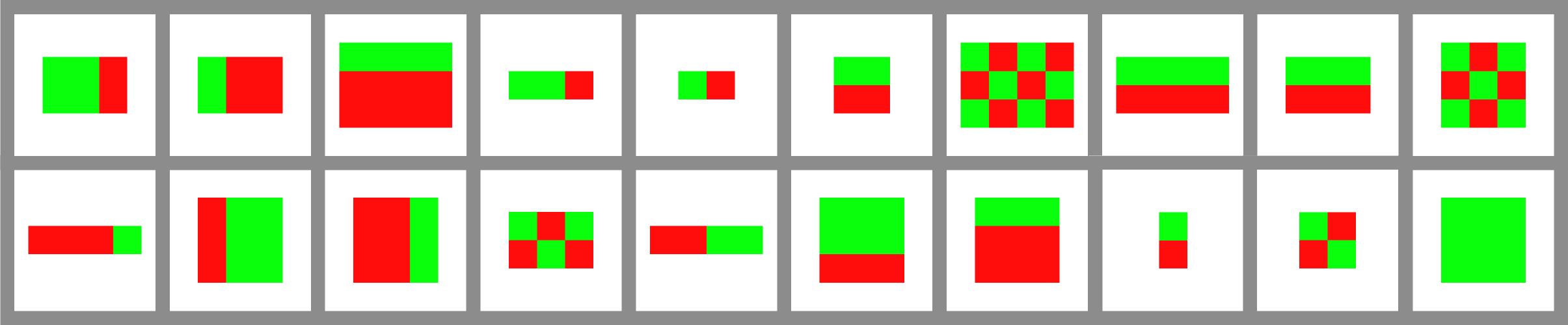}\qquad{}
\par\end{centering}

}
\par\end{centering}

\begin{centering}
\subfloat[\label{fig:rotated-filters-visualization}Illustration of Rotated
filters applied on each feature channel]{\begin{centering}
\qquad{}\includegraphics[width=0.8\columnwidth]{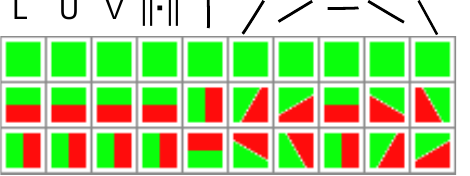}\qquad{}
\par\end{centering}

}
\par\end{centering}

\caption{\label{fig:detector-filters}Comparison of filters between some filtered
channels detector variants.}
\end{figure}

For our experiments we re-implement the filtered channel feature \texttt{Checkerboards}
detector \cite{Zhang2015Cvpr} using the \texttt{LDCF} \cite{Nam2014arXiv}
codebase. The training procedure turns out to be slow due to the
large number of filters (61 filters per channel). To accelerate the
training and test procedures, we design a small set of 9 filters per
channel that still provides good performance. We call our new filtered
channel feature detector; \texttt{Ro\-ta\-ted\-Fil\-ters} (see
figure \ref{fig:rotated-filters-visualization}).

The rotated filters are inspired by the filterbank of \texttt{LDCF}
(obtained by applying PCA to each feature channel). The first three
filters of \texttt{LDCF} of each features channel are the constant
filter and two step functions in orthogonal directions, with the particularities
that the oriented gradient channels also have rotated filters (see
figure \ref{fig:LDCF-filters}). Our rotated filters are stylised
versions of \texttt{LDCF}. The resulting \texttt{Ro\-ta\-ted\-Fil\-ters}
filterbank is somewhat intuitive, while filters from \texttt{Che\-cker\-bo\-ards},
are less systematic and less clear in their function (see figure \ref{fig:checkerboard-filters}).

To integrate richer local information, we repeat each filter per channel
over multiple scales, in the same spirit as \texttt{Squa\-res\-Chn\-Ftrs}
{\small{}\cite{Benenson2014Eccvw}} (figure \ref{fig:ACF-SquaresChnFtrs-filters}). 

On the Caltech validation set, \texttt{Ro\-ta\-ted\-Fil\-ters}
obtains 31.6\% $\mbox{MR}_{-2}^{O}$ using one scale (4x4); and 28.9\%
$\mbox{MR}_{-2}^{O}$ using three scales (4x4, 8x8 and 16x16). Therefore,
we select this 3-scale structure in our experiments. On the test set,
the performance of \texttt{Ro\-ta\-ted\-Fil\-ters} is 19.2\%
$\mbox{MR}_{-2}^{O}$, i.e.~a less than 1\% loss with respect to
\texttt{Che\-cker\-bo\-ards}, yet it is \textasciitilde{}6x faster
at feature computation. 

In this paper, we use \texttt{Ro\-ta\-ted\-Fil\-ters} for all
experiments involving training a new model.

\section{\label{sec:Subsets-results-supp}Results per test subset}

\begin{figure*}[h]
\begin{centering}
\hfill{}\subfloat[\label{fig:subset-reasonable-0.5}Reasonable setting (IoU >= 0.5)]{\centering{}\includegraphics[bb=23bp 5bp 417bp 321bp,width=0.32\textwidth]{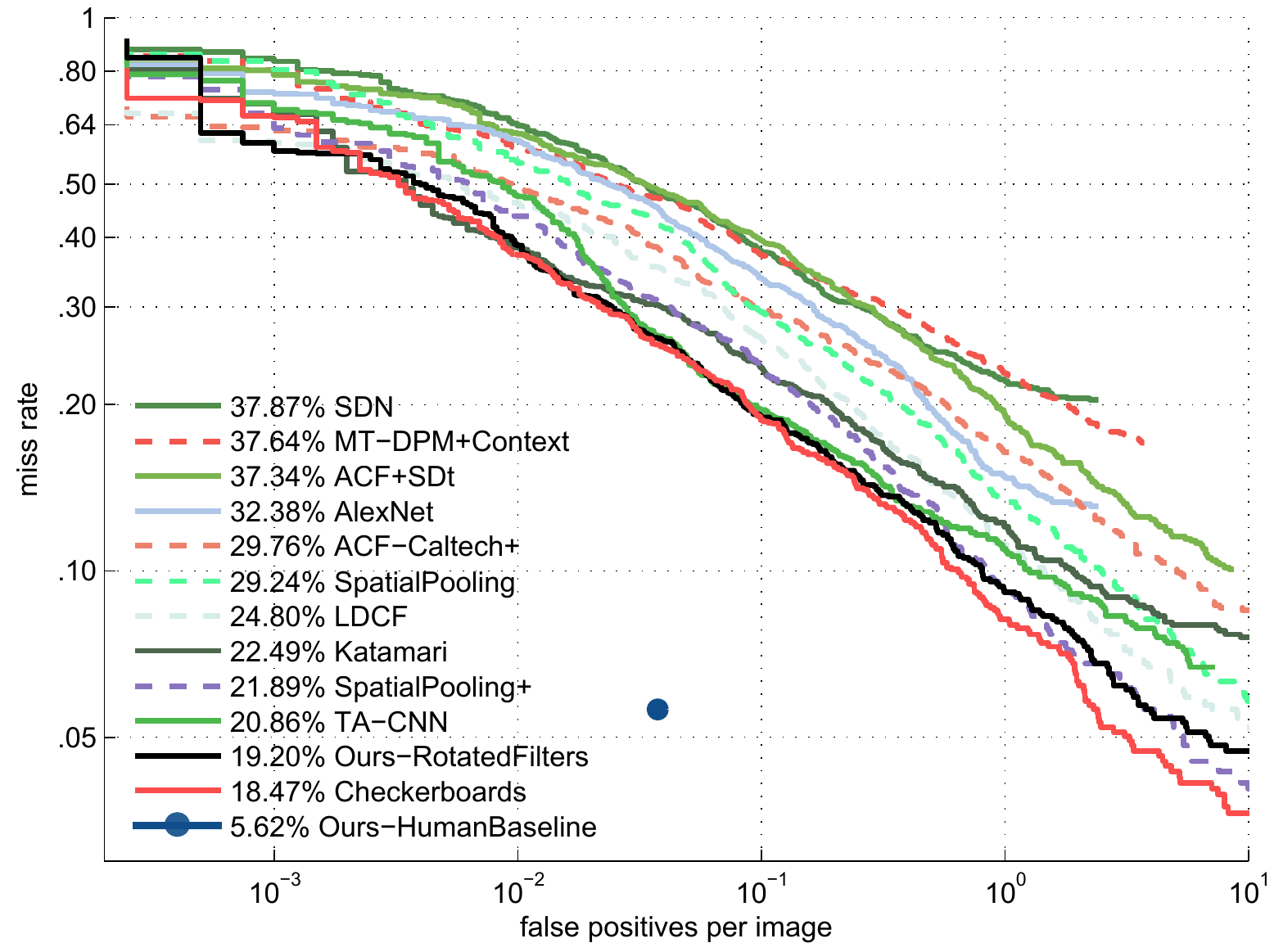}}\hfill{}\subfloat[\label{fig:subset-reasonable-0.8}Reasonable setting (IoU >= 0.8)]{\centering{}\includegraphics[bb=23bp 5bp 417bp 321bp,width=0.32\textwidth]{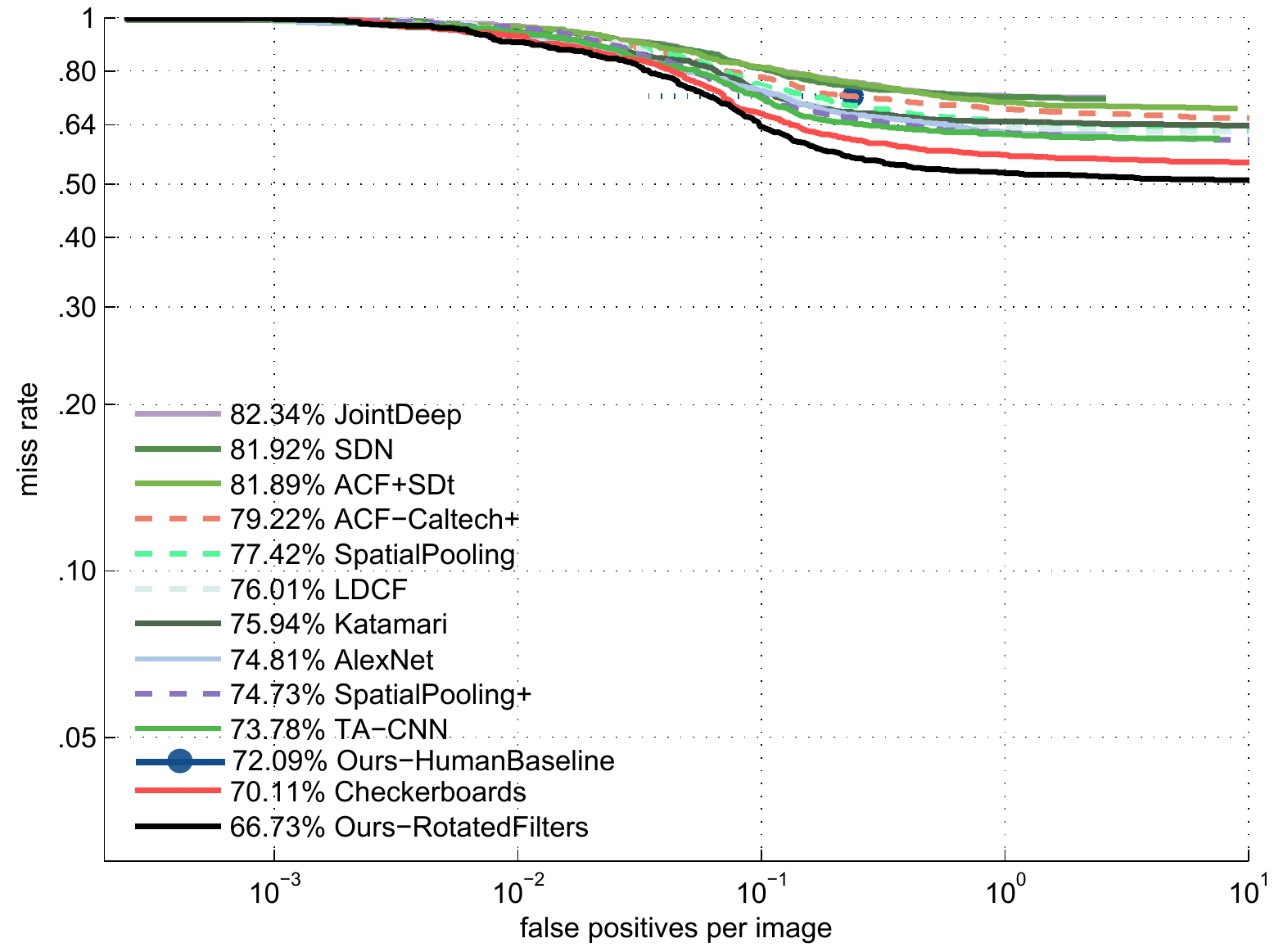}}\hfill{}
\par\end{centering}

\begin{centering}
\hfill{}\subfloat[\label{fig:subset-larger-than-80}Pedestrians larger than 80px in
height]{\begin{centering}
\includegraphics[bb=0bp 0bp 394bp 316bp,width=0.32\textwidth]{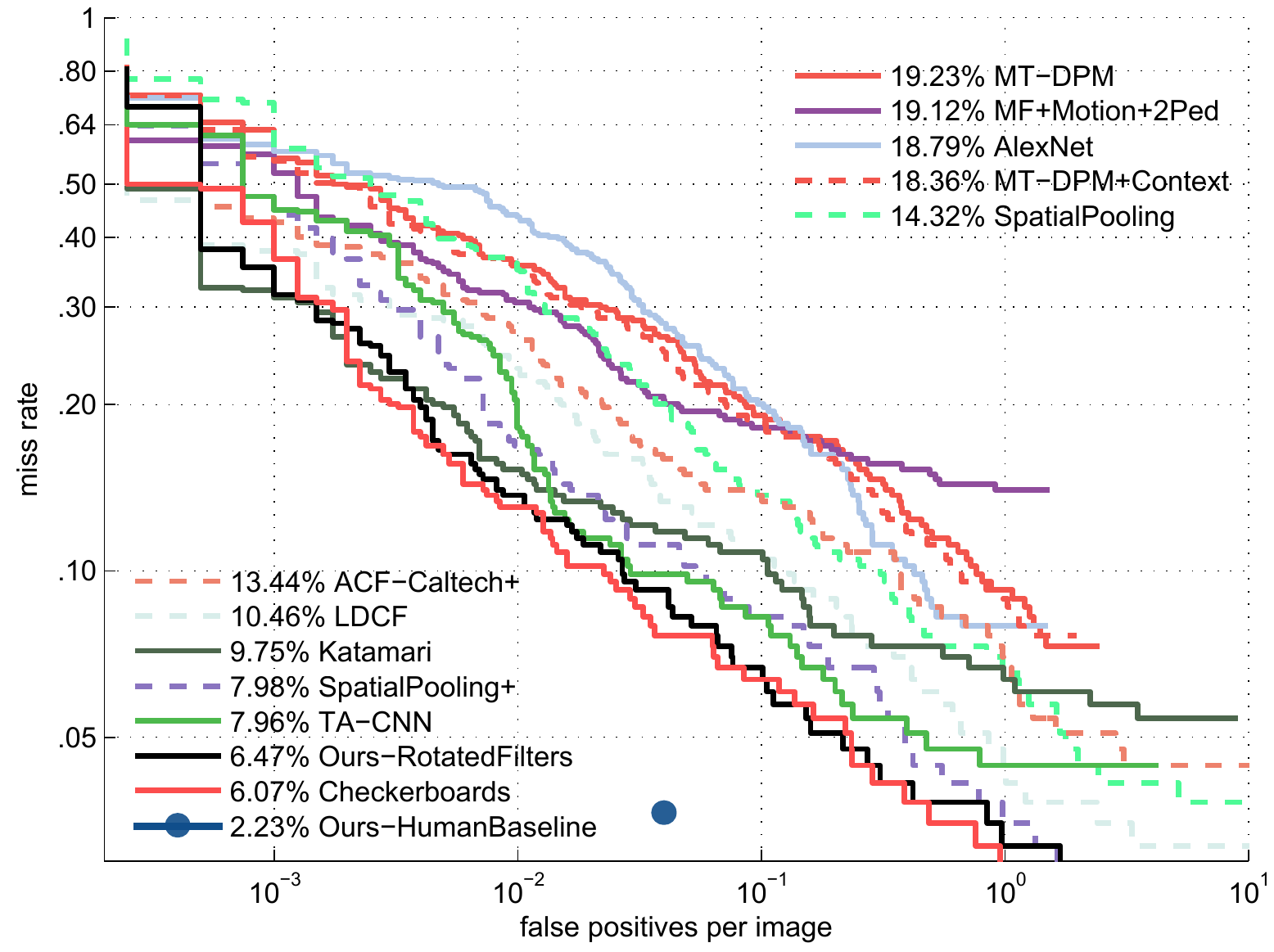}
\par\end{centering}

}\hfill{}\subfloat[\label{fig:subset-between-50-and-80} Pedestrian height between 50px
and 80px]{\begin{centering}
\includegraphics[bb=0bp 0bp 394bp 316bp,width=0.32\textwidth]{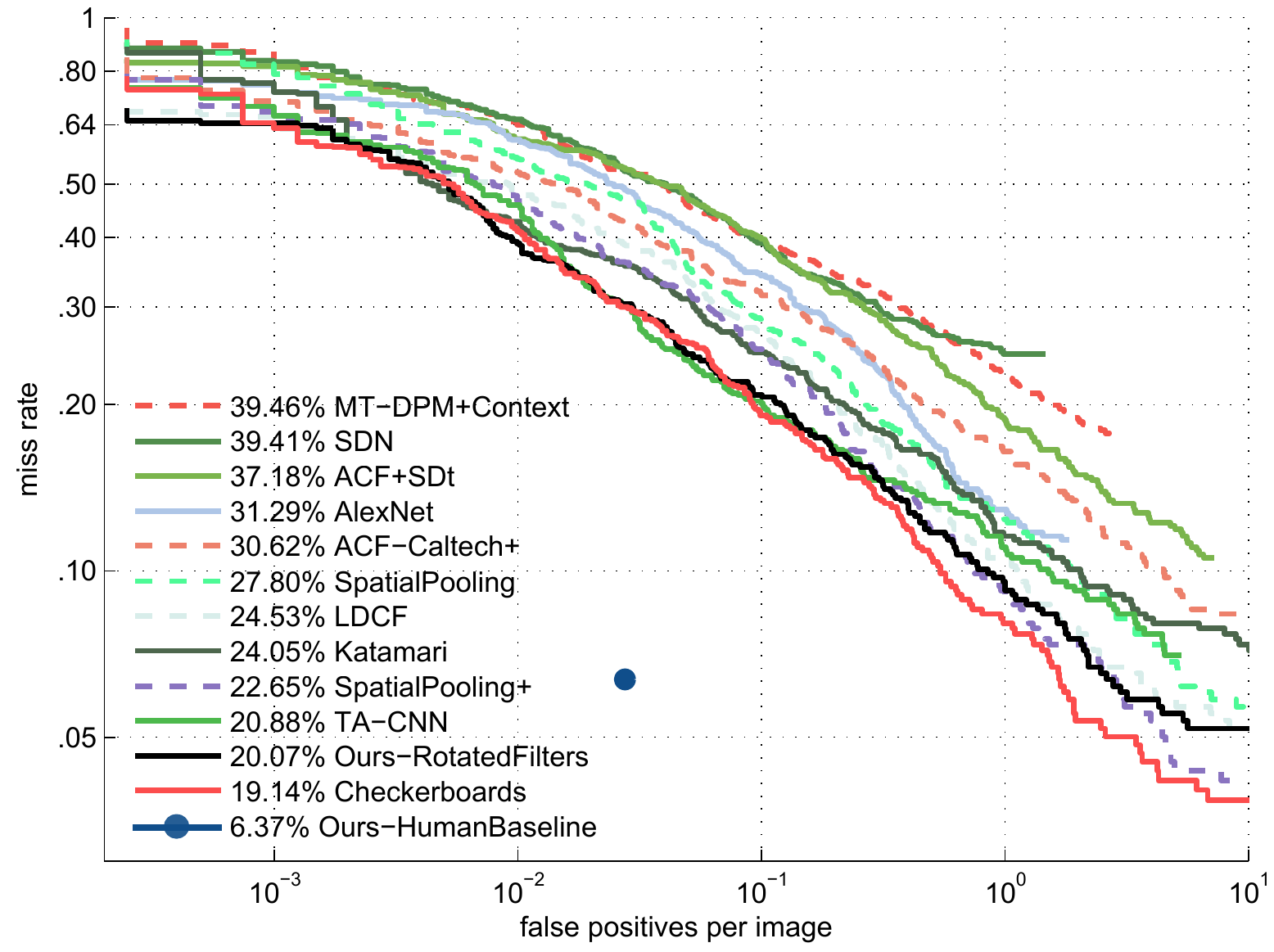}
\par\end{centering}

}\hfill{}
\par\end{centering}

\begin{centering}
\hfill{}\subfloat[\label{fig:subset-between-30-and-50} Pedestrian height between 30px
and 50px]{\begin{centering}
\includegraphics[bb=0bp 0bp 394bp 316bp,width=0.32\textwidth]{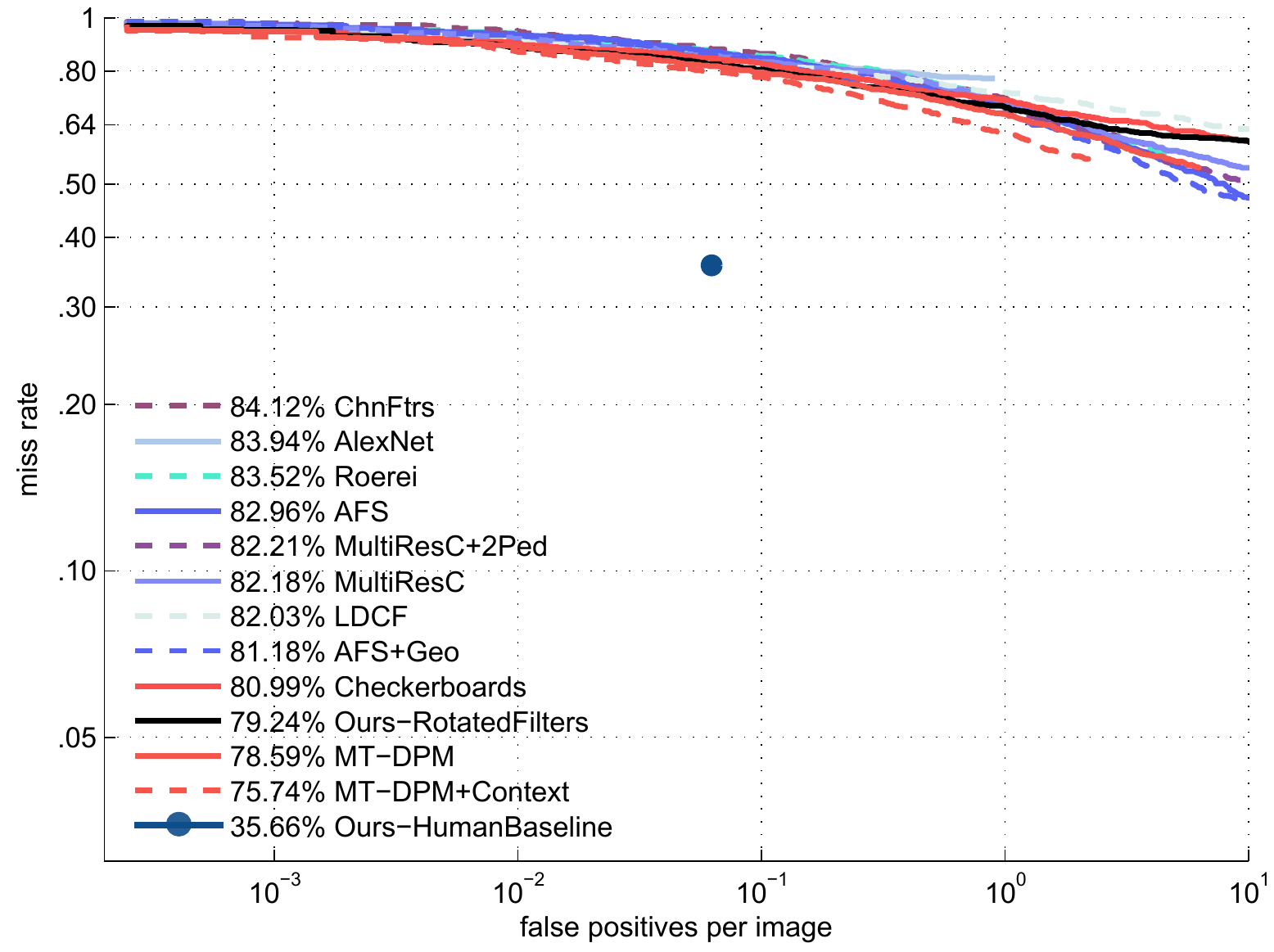}
\par\end{centering}

}\hfill{}\subfloat[\label{fig:subset-non-occluded}Non-occluded pedestrians]{\begin{centering}
\includegraphics[bb=0bp 0bp 394bp 316bp,width=0.32\textwidth]{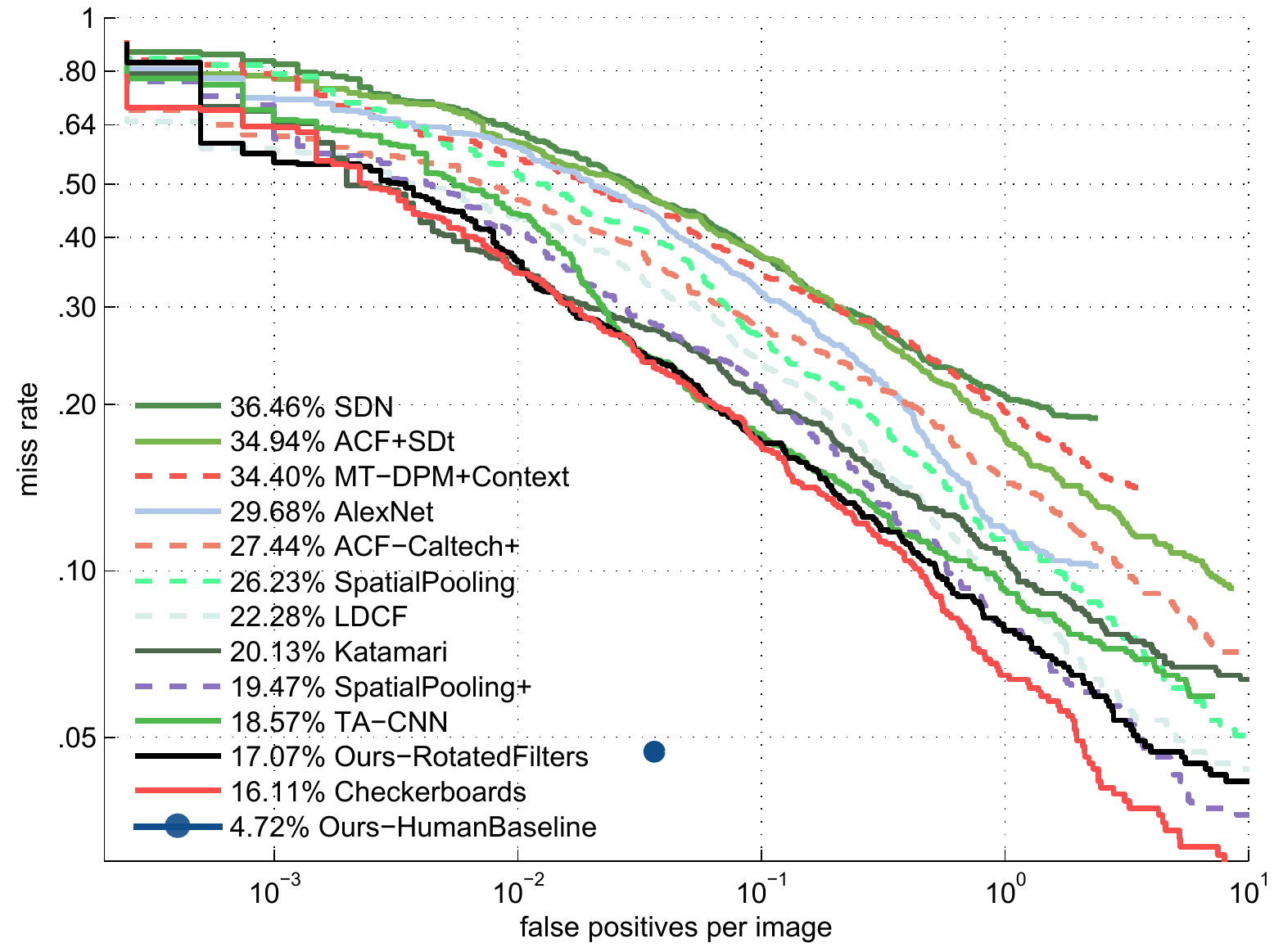}
\par\end{centering}

}\hfill{}
\par\end{centering}

\begin{centering}
\hfill{}\subfloat[\label{fig:subset-occluded-less-than-35}Pedestrians occluded by up
to 35\%]{\begin{centering}
\includegraphics[bb=0bp 0bp 394bp 316bp,width=0.32\textwidth]{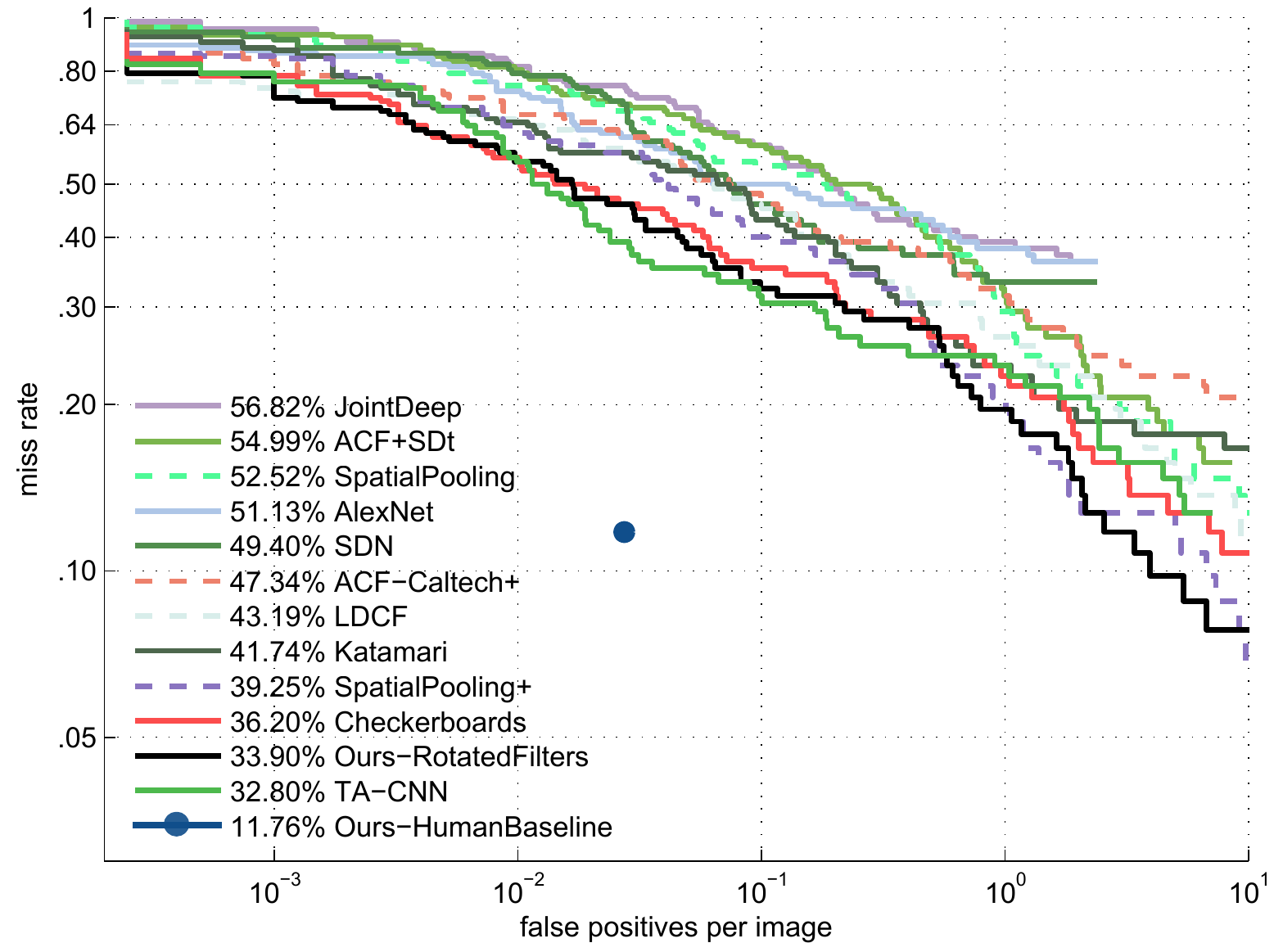}
\par\end{centering}

}\hfill{}\subfloat[\label{fig:subset-occluded-35-to-80}Pedestrians occluded by more
than 35\% and less than 80\%.]{\begin{centering}
\includegraphics[bb=0bp 0bp 394bp 316bp,width=0.32\textwidth]{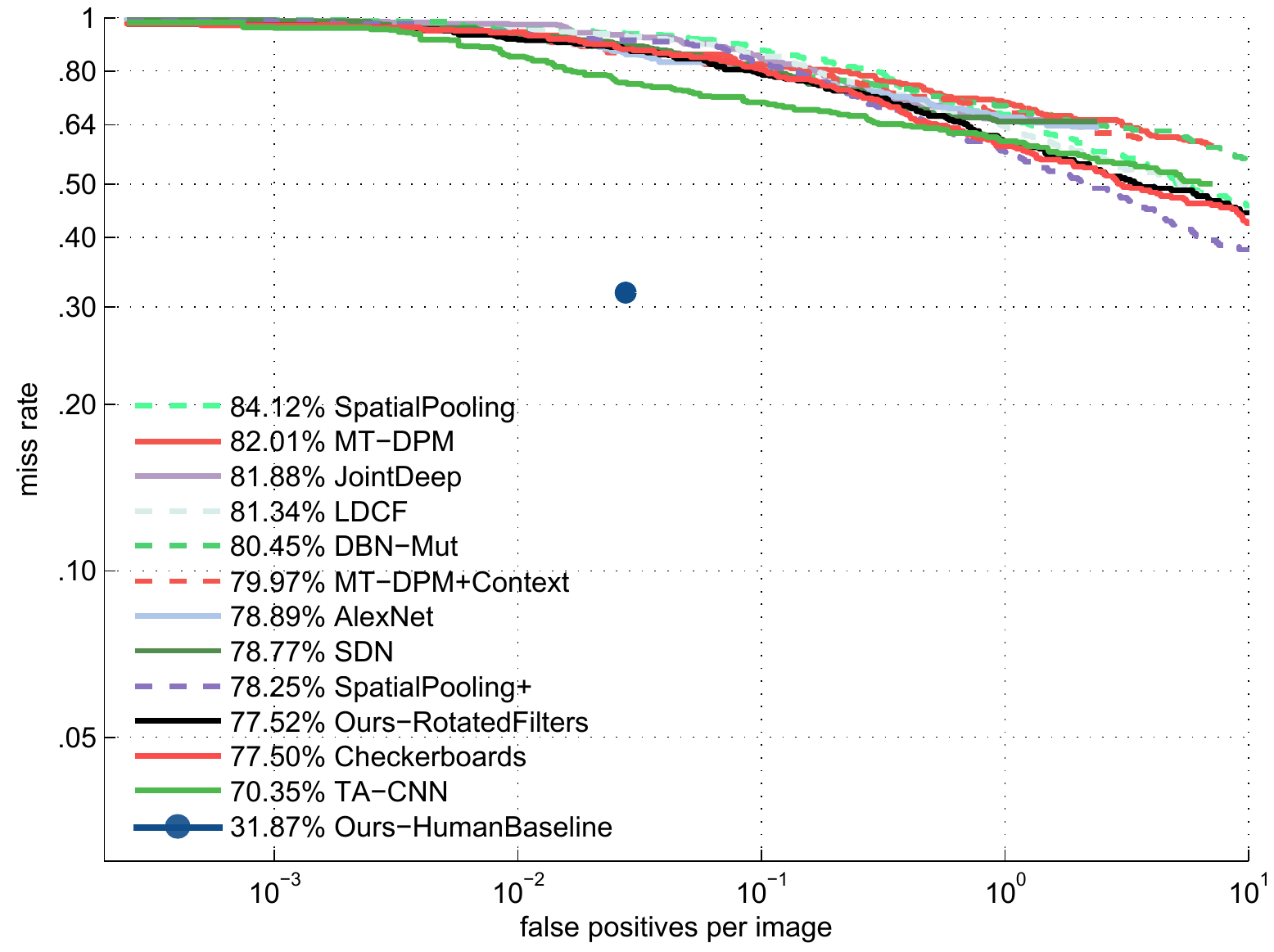}
\par\end{centering}

}\hfill{}
\par\end{centering}

\caption{\label{fig:subsets-curves-supp}Detection quality of top-performing
methods on experimental settings depicted in ``subsets bar plot''
figure in the main paper.}
\end{figure*}

Figure \ref{fig:subsets-curves-supp} contains the detailed curves
behind figure 3 in the main paper (``subsets bar plot''). We can
see that \texttt{Che\-cker\-boards} and \texttt{Ro\-tated\-Fil\-ters}
show good performance across all subsets. The few cases where they
are not top ranked (e.g. figures \ref{fig:subset-between-30-and-50}
and \ref{fig:subset-occluded-35-to-80}) all methods exhibit low detection
quality, and thus all have similarly poor scores. 

Figure \ref{fig:subsets-curves-supp} shows that \texttt{Checker\-boards
}is not optimised for the most common case on the Caltech dataset,
but instead shows good performance across a variety of situations;
and is thus an interesting method to analyse.

 \newpage{}

\clearpage{}\clearpage{}

\section{\label{sec:Errors-analysis-supp}\texttt{Checkerboards} errors analysis}

\paragraph{Error examples}

Figure \ref{fig:errors-examples-fp-localization}, \ref{fig:errors-examples-fp-background},
\ref{fig:errors-examples-fp-annotation} and \ref{fig:eeasupp1},
show four examples for each error type considered in the analysis
of the main paper (for both false positives and false negatives).

\paragraph{Blur and contrast measures}

To enable our analysis regarding blur and contrast, we define two
automated measures. We measure blur using the method from \cite{Roffet07SPIE},
while contrast is measured via the difference between the top and
bottom quantiles of the grey scale intensity of the pedestrian patch.\\
Figures \ref{fig:Examples-for-blur} and \ref{fig:Examples-for-contrast}
show pedestrians ranked by our blur and contrast measures. One can
observe that our quantitative measures correlate well with the qualitative
notions of blur and contrast.

\paragraph{Scale, blur, or contrast?}

For false negatives, a major source of error is small scale, but we
find small pedestrians are often of low contrast or blurred. In order
to investigate the three factors separately, we observe the correlation
between size/contrast/blur and score, as shown in figure \ref{fig:Correlation-between-size/contrast/blur-score}.
We can see that the overlap between false positive and true positive
is equally distributed across different levels of contrast and blur;
while for scale, the overlap is quite dense at small scale. To this
end, we conclude that small scale is the main factor negatively impacting
detection quality; and that blur and contrast are uninformative measures
for the detection task.

\subsection{Oracle cases}

\begin{figure}
\noindent \begin{centering}
\subfloat[Low-scoring objects]{\begin{centering}
\includegraphics[width=1\columnwidth]{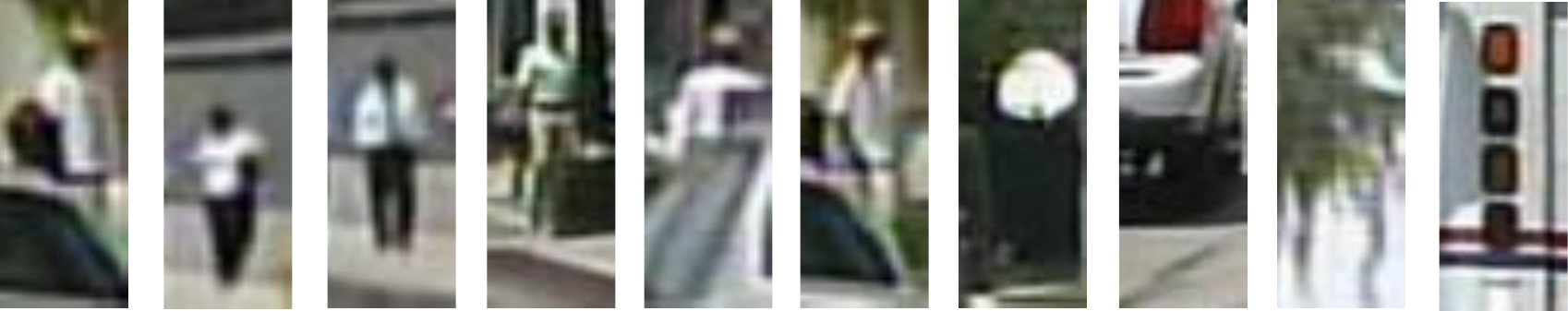}
\par\end{centering}

}
\par\end{centering}

\subfloat[High-scoring objects]{\begin{centering}
\includegraphics[width=1\columnwidth]{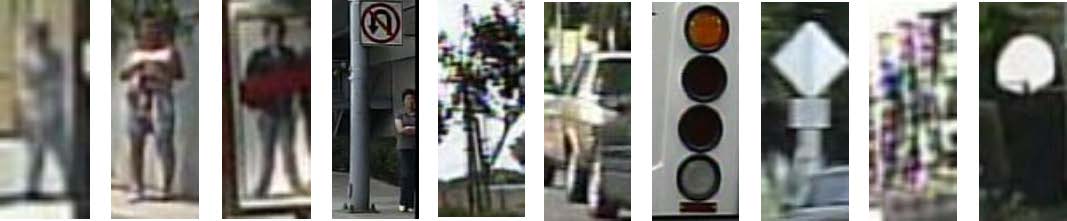}
\par\end{centering}

}

\caption{\label{fig:same-score-objects} Failure cases of Checkerboards detector
\cite{Zhang2015Cvpr}. Each group shows image patches of similar scores:
some background objects are of high scores; while some persons are
of low scores. We aim to understand when the detector fails through
analysis.}
\end{figure}

\begin{figure}
\subfloat[\label{fig:Size-vs-score}Size versus score]{\begin{centering}
\includegraphics[width=1\columnwidth]{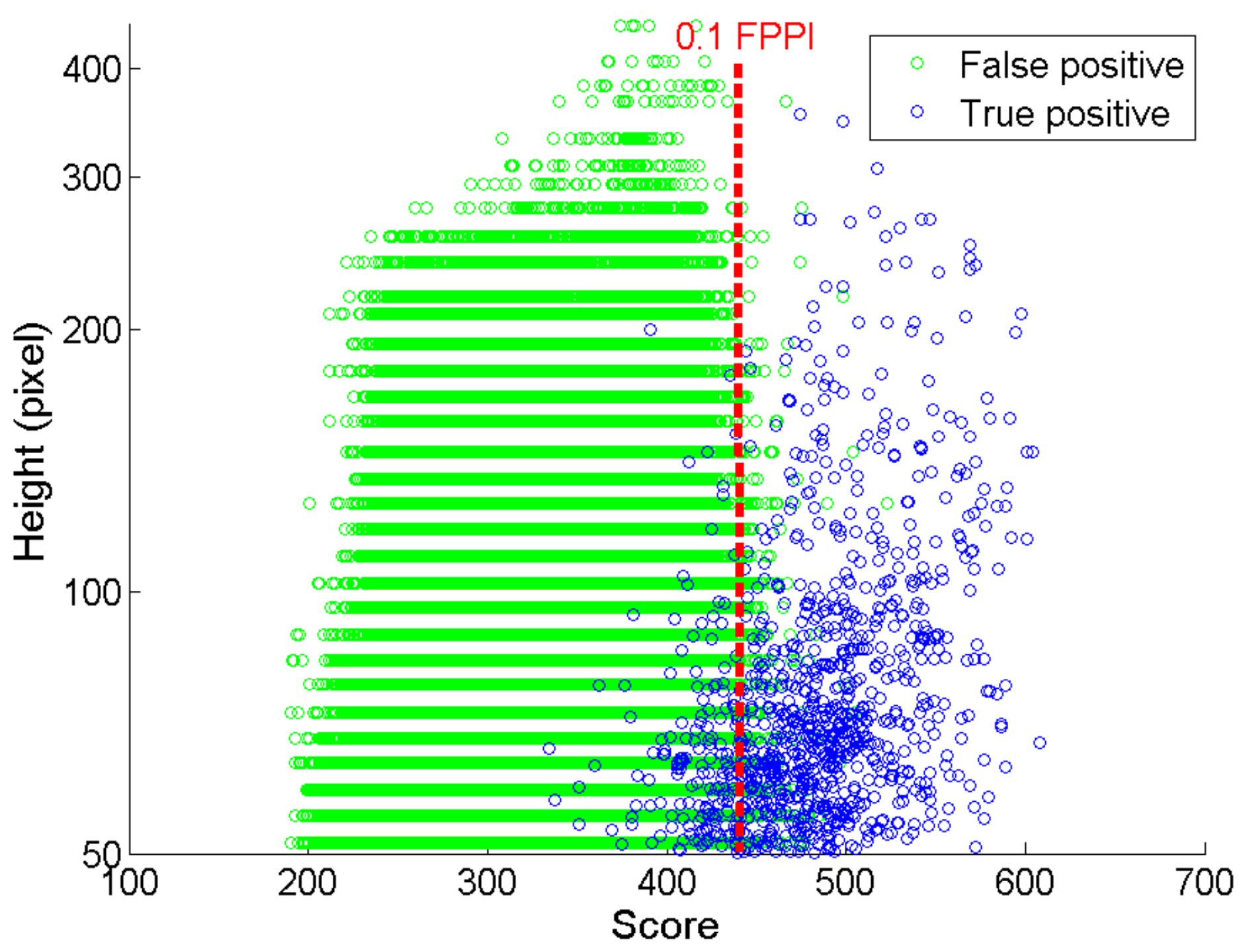}
\par\end{centering}

}

\subfloat[\label{fig:Contrast-vs-score-1}Contrast versus score]{\centering{}\includegraphics[width=1\columnwidth]{contrast_vs_score5}}

\subfloat[\label{fig:Blur-vs-score}Blur versus score]{\begin{centering}
\includegraphics[width=1\columnwidth]{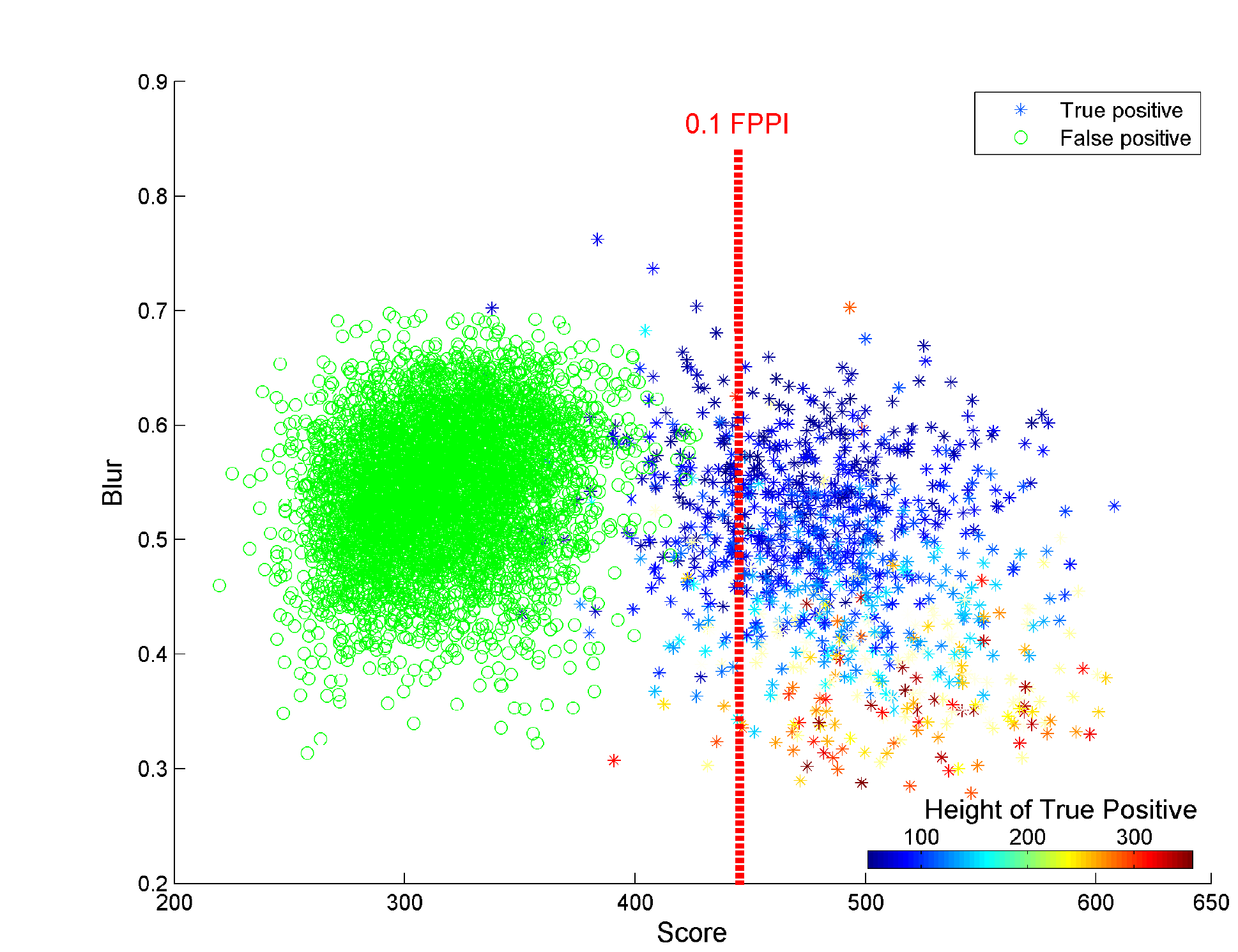}
\par\end{centering}

}

\caption{\label{fig:Correlation-between-size/contrast/blur-score}Correlation
between size/contrast/blur and score.}
\end{figure}

\begin{figure*}
\begin{centering}
\includegraphics[width=1\linewidth]{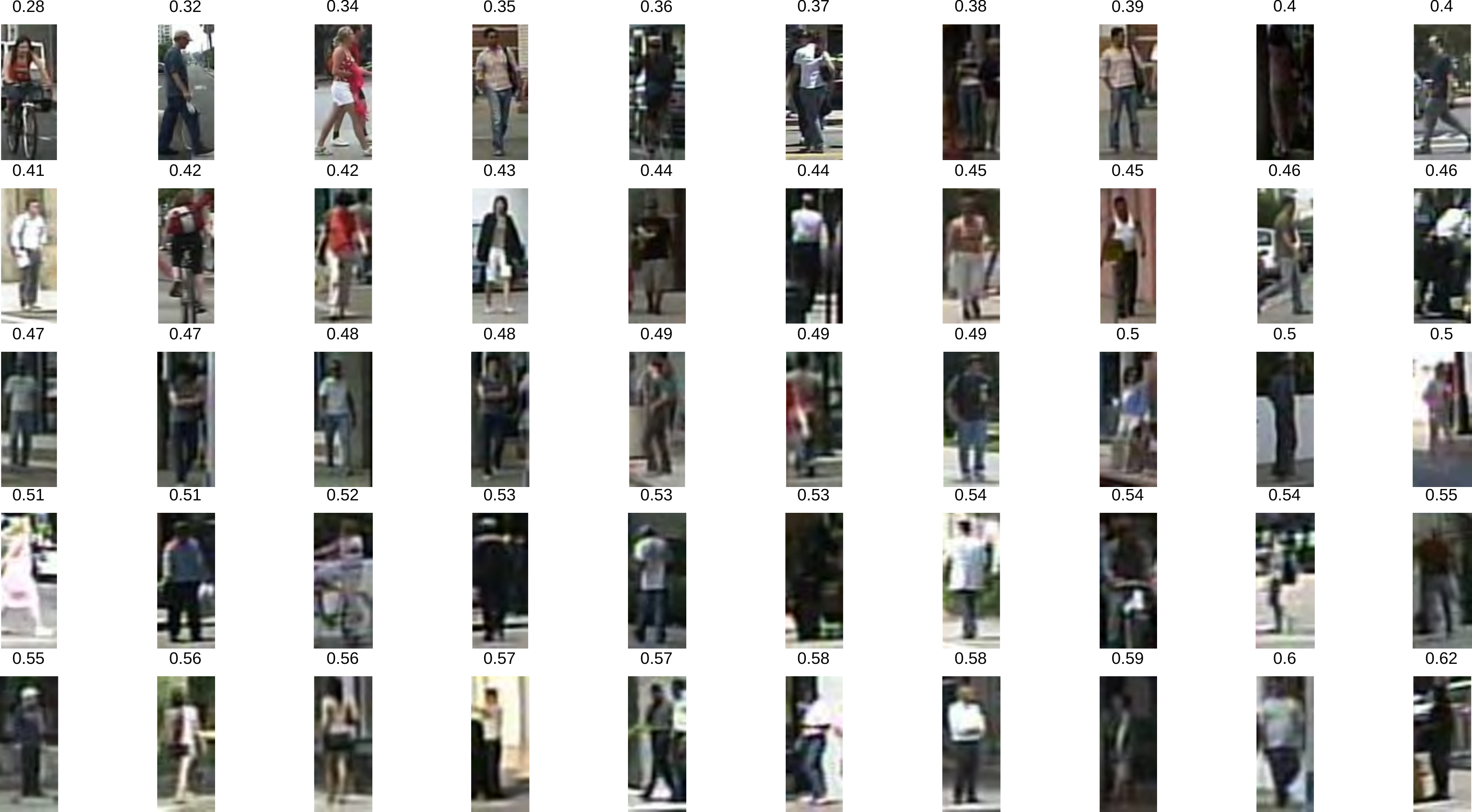}
\par\end{centering}

\caption{\label{fig:Examples-for-blur}Examples for images with different levels
of blur.}
\end{figure*}

\begin{figure*}
\begin{centering}
\includegraphics[width=1\linewidth]{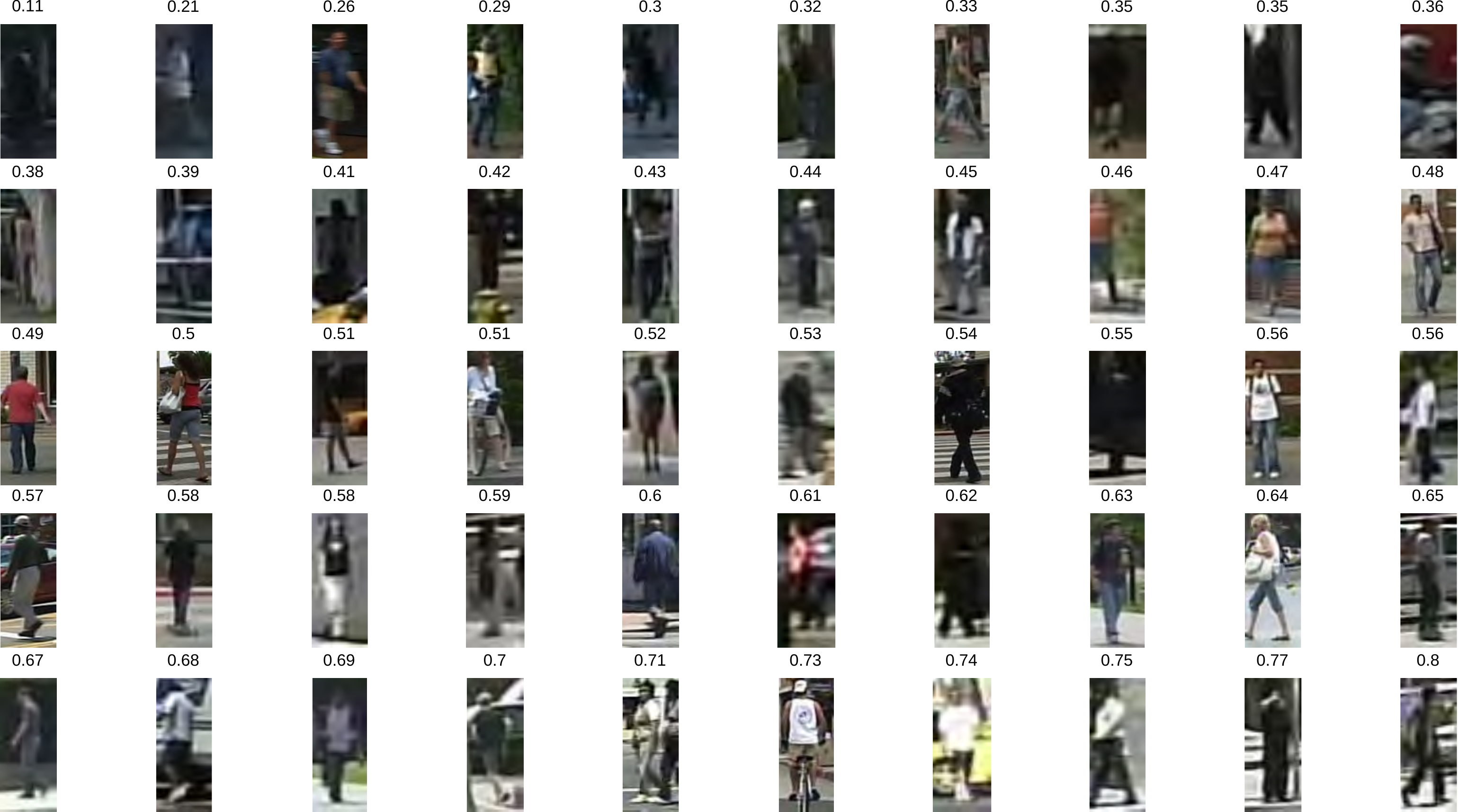}
\par\end{centering}

\caption{\label{fig:Examples-for-contrast}Examples for images with different
levels of contrast.}
\end{figure*}

\begin{figure*}
\begin{centering}
\subfloat[\label{fig:example-error-type-A}Double detection]{\begin{centering}
\includegraphics[width=1\linewidth]{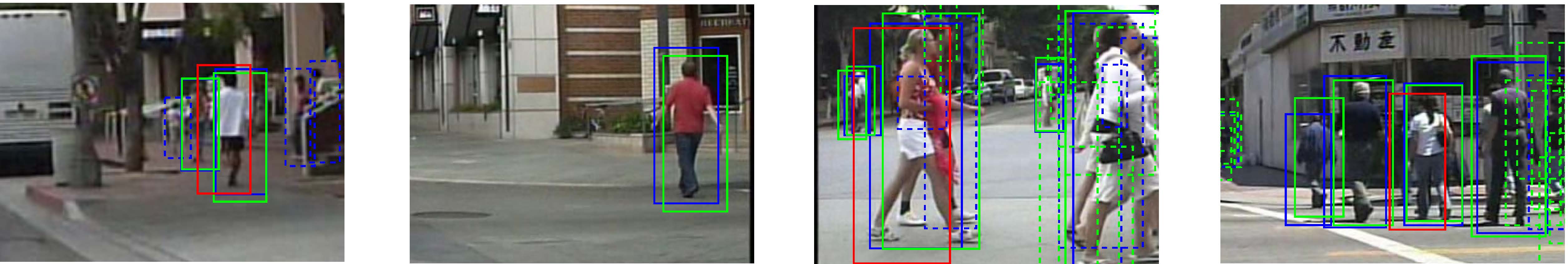}
\par\end{centering}

} 
\par\end{centering}

\begin{centering}
\subfloat[\label{fig:example-error-type-B}Body parts]{\begin{centering}
\includegraphics[width=1\linewidth]{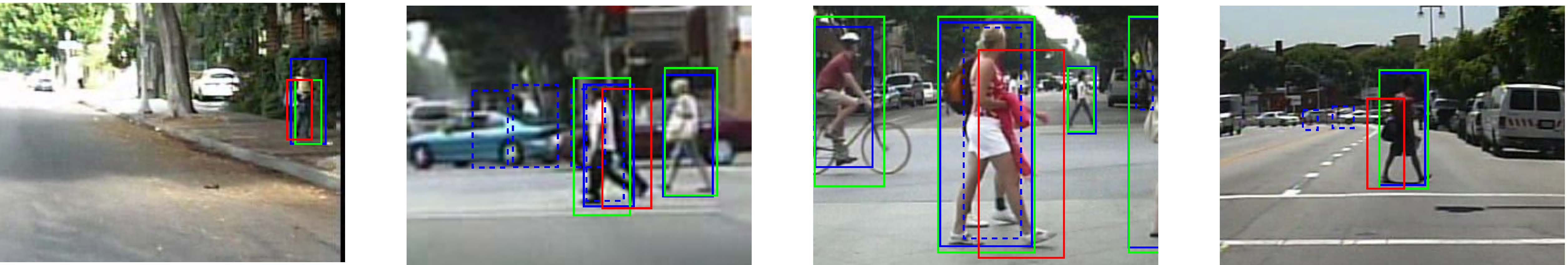}
\par\end{centering}

}
\par\end{centering}

\begin{centering}
\subfloat[\label{fig:example-error-type-C}Too large bounding boxes]{\begin{centering}
\includegraphics[width=1\linewidth]{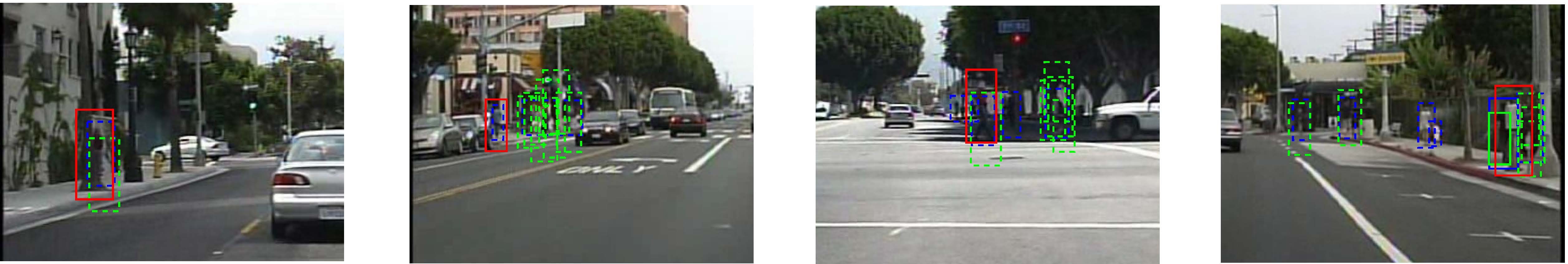}
\par\end{centering}

}
\par\end{centering}

\caption{\label{fig:errors-examples-fp-localization}Example localisation errors,
a subset of false positives. False positives in red, original annotations
in blue, ignore annotations in dashed blue, true positives in green,
and ignored detections in dashed green (because they overlap with
ignore annotations).}
\end{figure*}

\begin{figure*}
\begin{centering}
\subfloat[\label{fig:example-error-type-D}Vertical structures]{\begin{centering}
\includegraphics[width=1\linewidth]{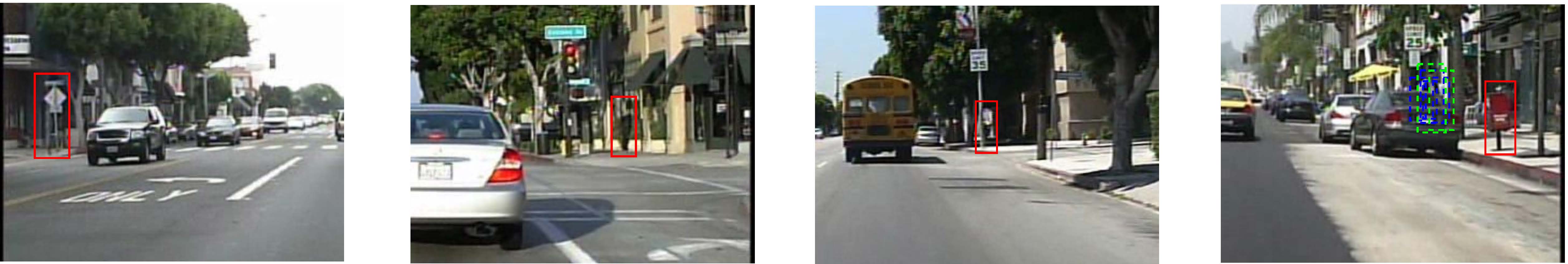}
\par\end{centering}

}
\par\end{centering}

\begin{centering}
\subfloat[\label{fig:example-error-type-E}Traffic lights]{\begin{centering}
\includegraphics[width=1\linewidth]{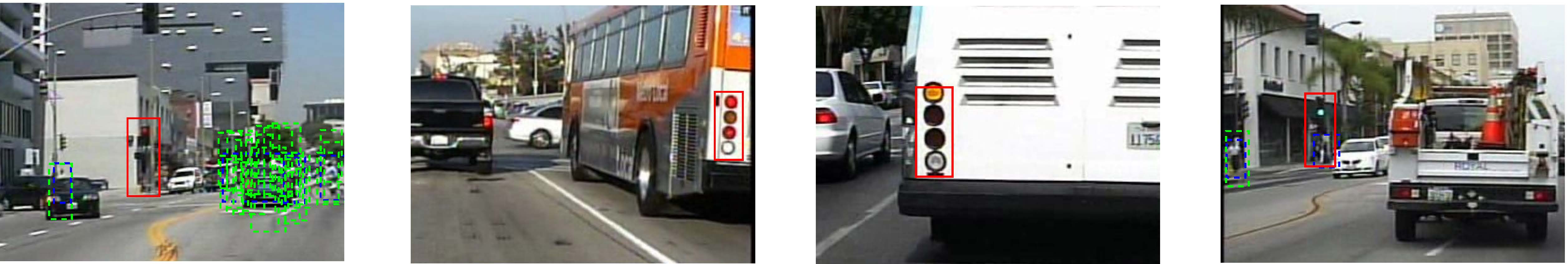}
\par\end{centering}

}
\par\end{centering}

\begin{centering}
\subfloat[\label{fig:example-error-type-F}Car parts]{\begin{centering}
\includegraphics[width=1\linewidth]{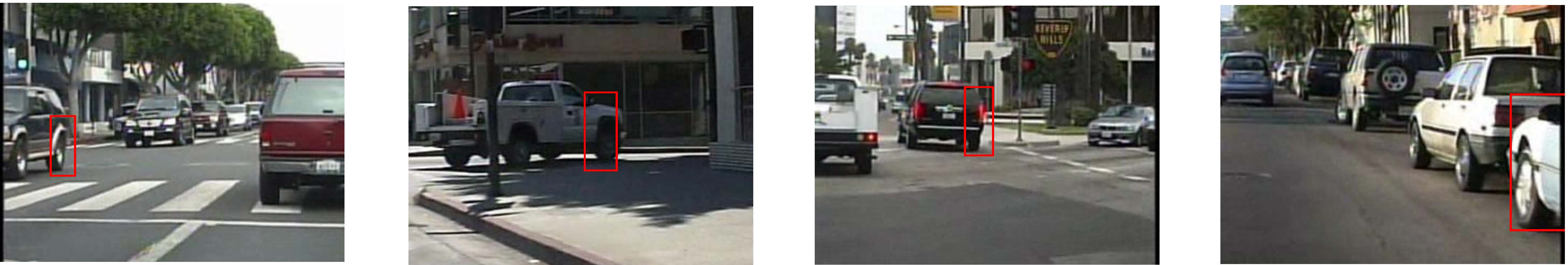}
\par\end{centering}

}
\par\end{centering}

\begin{centering}
\subfloat[\label{fig:example-error-type-G}Tree leaves]{\begin{centering}
\includegraphics[width=1\linewidth]{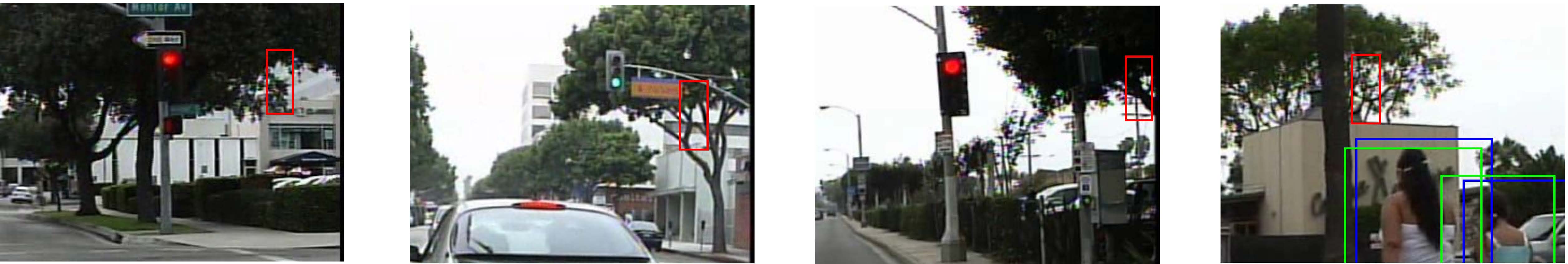}
\par\end{centering}

}
\par\end{centering}

\begin{centering}
\subfloat[\label{fig:example-error-type-H}Other background]{\begin{centering}
\includegraphics[width=1\linewidth]{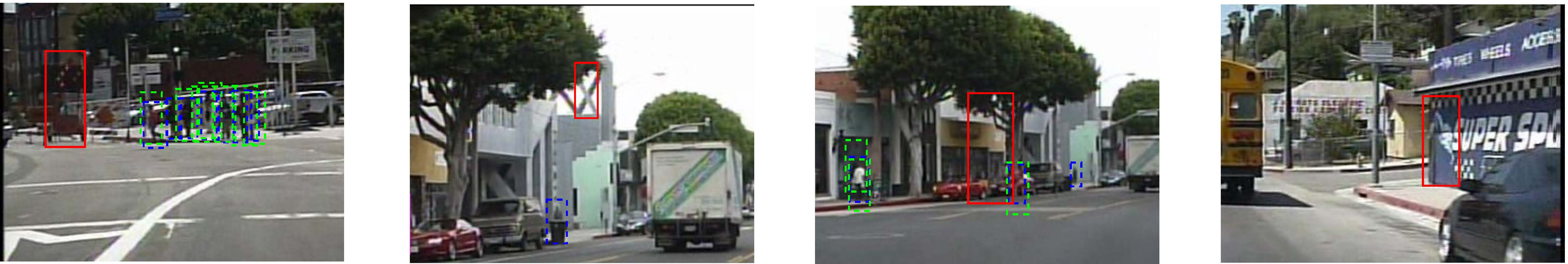}
\par\end{centering}

}
\par\end{centering}

\caption{\label{fig:errors-examples-fp-background}Example background errors,
a subset of false positives. False positives in red, original annotations
in blue, ignore annotations in dashed blue, true positives in green,
and ignored detections in dashed green (because they overlap with
ignore annotations).}
\end{figure*}

\begin{figure*}
\begin{centering}
\subfloat[\label{fig:example-error-type-I}Fake humans]{\begin{centering}
\includegraphics[width=1\linewidth]{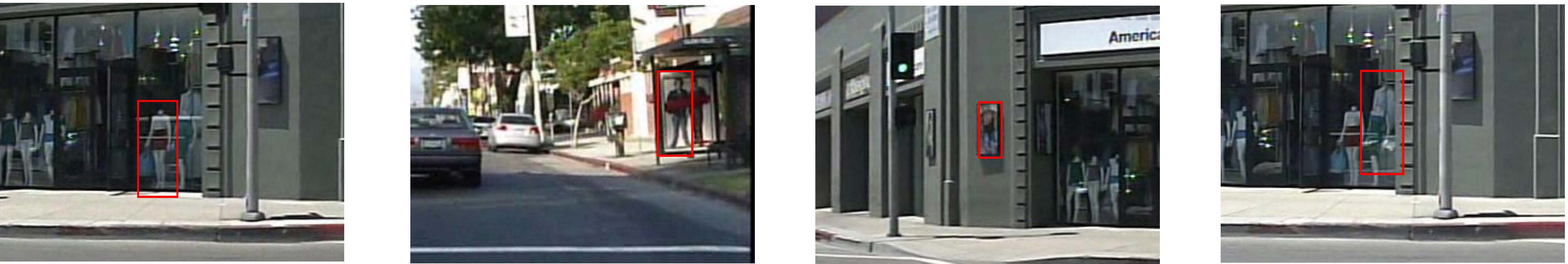}
\par\end{centering}

}
\par\end{centering}

\begin{centering}
\subfloat[\label{fig:example-error-type-J}Missing annotations]{\begin{centering}
\includegraphics[width=1\linewidth]{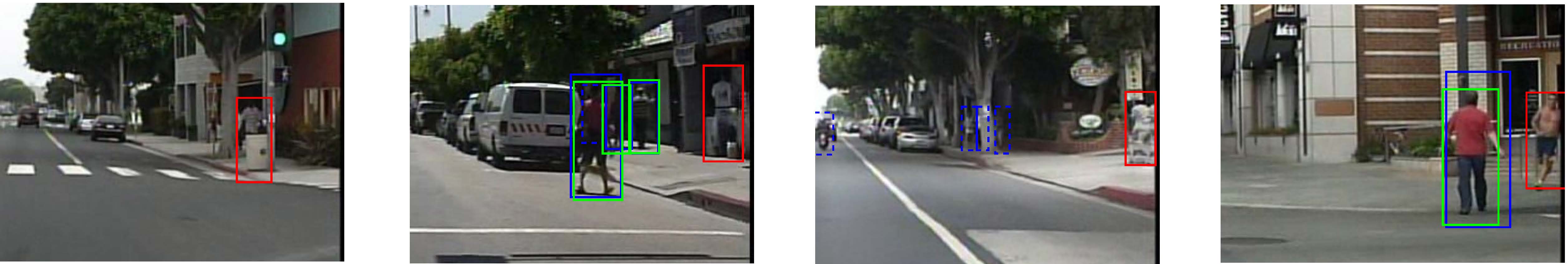}
\par\end{centering}

}
\par\end{centering}

\begin{centering}
\subfloat[\label{fig:example-error-type-K}Confusing]{\begin{centering}
\includegraphics[width=1\linewidth]{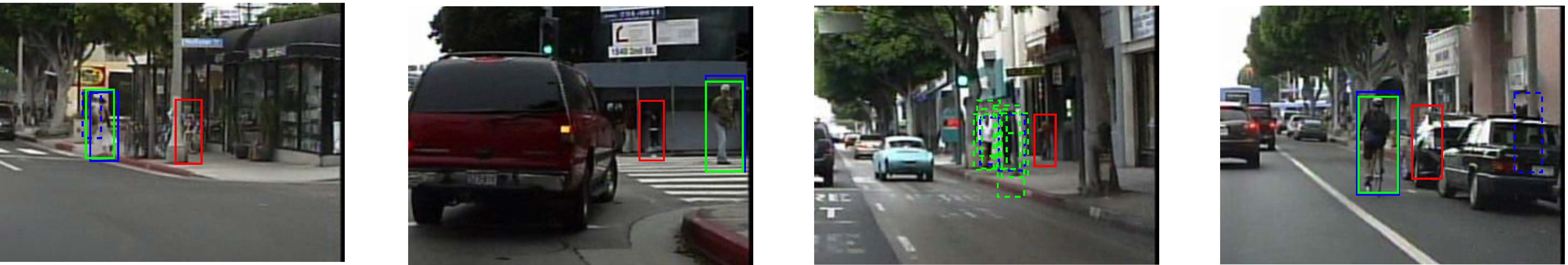}
\par\end{centering}

}
\par\end{centering}

\caption{\label{fig:errors-examples-fp-annotation}Example annotation errors,
a subset of false positives. False positives in red, original annotations
in blue, ignore annotations in dashed blue, true positives in green,
and ignored detections in dashed green (because they overlap with
ignore annotations).}
\end{figure*}

\begin{figure*}
\begin{centering}
\vspace{-0.8em}
\subfloat[\label{fig:eetA1}Small scale]{\begin{centering}
\includegraphics[width=1\textwidth]{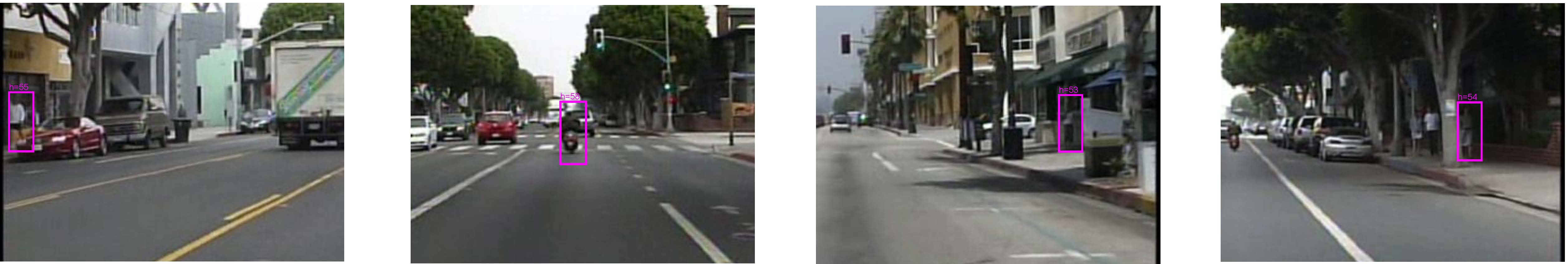}
\par\end{centering}

}
\par\end{centering}

\begin{centering}
\vspace{-0.8em}
\subfloat[\label{fig:eetB1}Side view]{\begin{centering}
\includegraphics[width=1\textwidth]{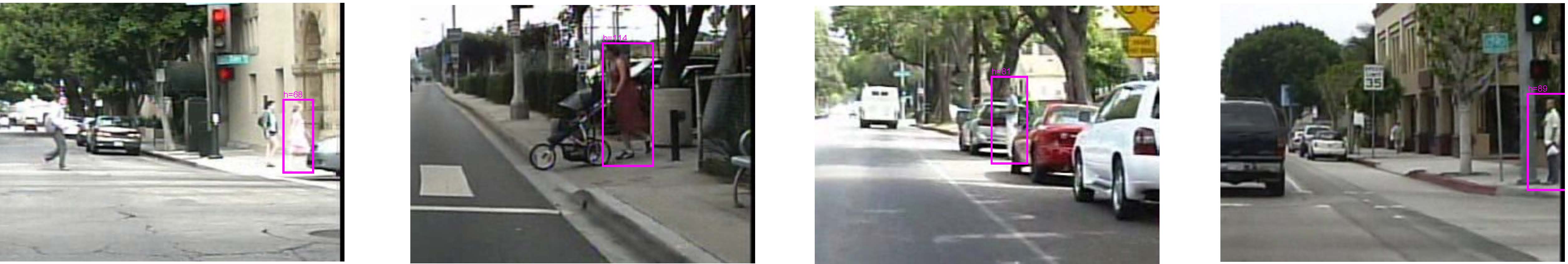}
\par\end{centering}

}
\par\end{centering}

\begin{centering}
\vspace{-0.8em}
\subfloat[\label{fig:eetC1}Cyclists]{\begin{centering}
\includegraphics[width=1\textwidth]{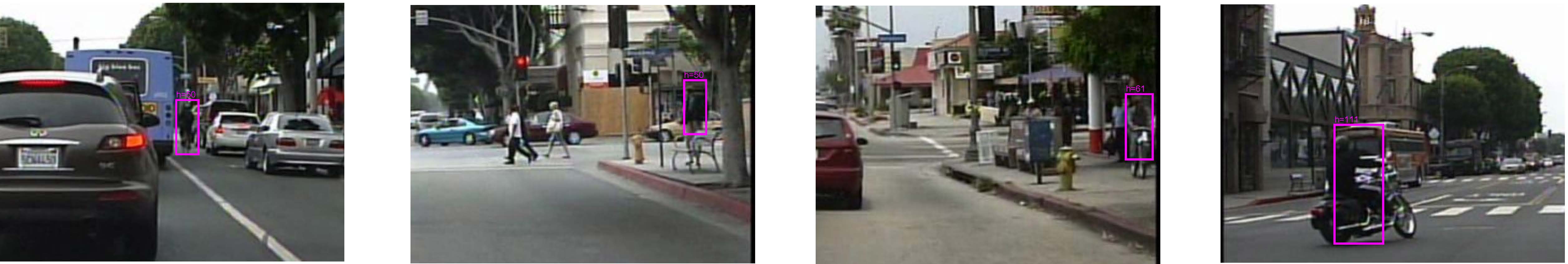}
\par\end{centering}

}
\par\end{centering}

\begin{centering}
\vspace{-0.8em}
\subfloat[\label{fig:eetD1}Occlusion]{\begin{centering}
\includegraphics[width=1\textwidth]{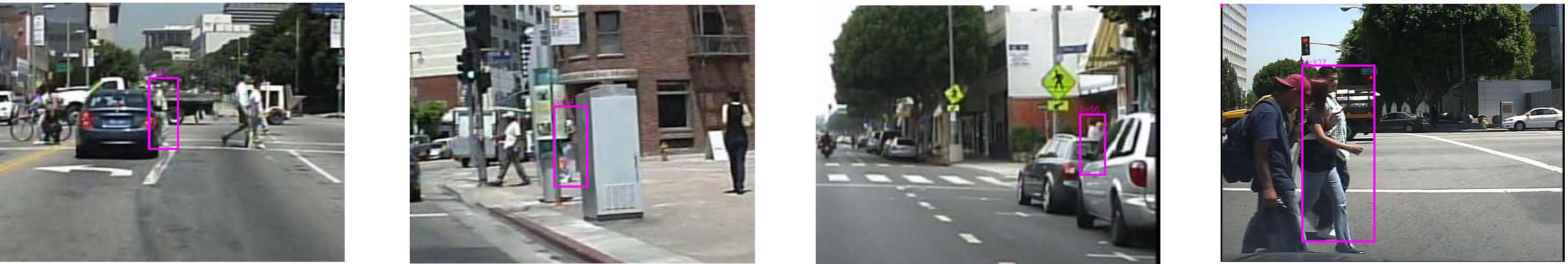}
\par\end{centering}

}
\par\end{centering}

\begin{centering}
\vspace{-0.8em}
\subfloat[\label{fig:eetE1}Annotation errors]{\begin{centering}
\vspace{-1em}
\includegraphics[width=1\textwidth]{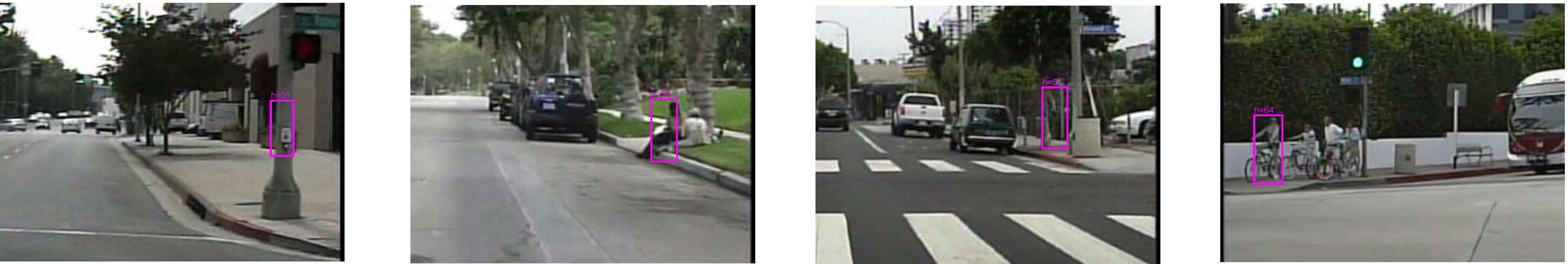}
\par\end{centering}

}
\par\end{centering}

\begin{centering}
\vspace{-0.8em}
\subfloat[\label{fig:eetF1}Others]{\begin{centering}
\includegraphics[width=1\textwidth]{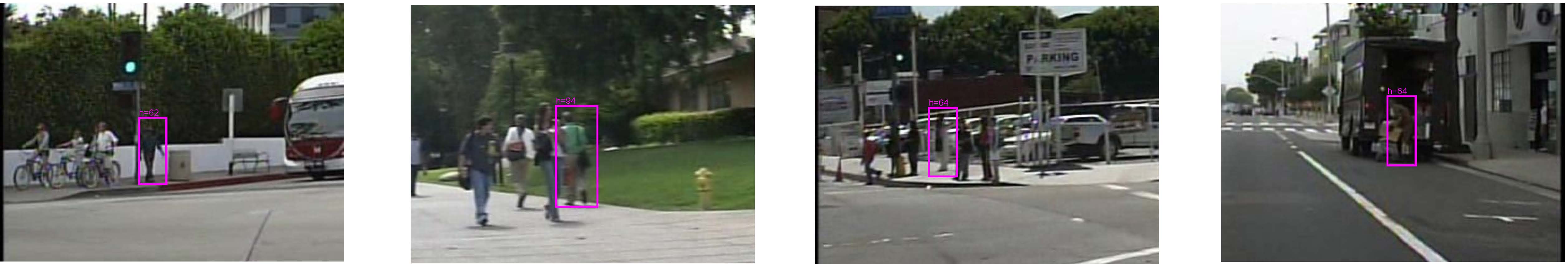}
\par\end{centering}

}
\par\end{centering}

\caption{\label{fig:eeasupp1}Example errors for different error types of false
negatives. False positives in red, original annotations in blue, ignore
annotations in dashed blue, true positives in green, and ignored detections
in dashed green (because they overlap with ignore annotations).}
\end{figure*}

In figure \ref{fig:oracle-cases}, we show the standard evaluation
and oracle evaluation curves for state-of-the-art methods. For the
localisation oracle, false positives that overlap with the ground
truth are not considered; for the background-versus-foreground oracle,
false positives that do not overlap with the ground truth are not
considered. Based on the curves, we have the following findings: 
\begin{itemize}
\item All methods are significantly improved in each oracle evaluation.
\item The ranking of all methods stays relatively stable in each oracle
case.
\item In terms of MR$_{-4}^{\mbox{O}}$, the improvement is comparable for
localisation or background-versus-foreground oracle tests; the detection
performance can be boosted by fixing either problem.
\end{itemize}
We also show some examples of objects with similar scores in figure
\ref{fig:same-score-objects}. In both low-scoring and high-scoring
groups, we can see both pedestrians and background objects, which
shows that the detector fails to rank foreground and background adequately.

\subsection{\label{sec:Log-scale}Log scale visual distortion}

In the paper we show results for so called oracle experiments that
emulate the case in which we do not make one type of error: we remove
either mistakes that touch annotated pedestrians (localisation oracle)
or mistakes that are located on background (background oracle). 

It is important to note that these are the only two types of false
positives. If we remove both types the only mistakes that remain stem
from missing recall and the result would be a horizontal line with
very low miss rate. 

Because of the double log scale in the performance plots on Caltech
the curves look like both oracles improve performance slightly but
the bulk of mistakes arise from a different type of mistakes, which
is not the case.

In figure \ref{fig:loglog-is-strange} we illustrate how much double
log scales distort areas. We often think of the average miss rate
as the area under the curve, so we colour code the false positives
in the plots by their type: the plot shows the ratio between localisation
(blue) and background (green) mistakes at every point on the miss
rate, but also for the entire curve. Both curves, \ref{fig:loglog-is-strange-localize}
and \ref{fig:loglog-is-strange-background} show the same data with
the only difference that one shows localisations on the left and the
other one on the right. Due to the double log scale, the error type
that is plotted on the left seems to dominate the metric.

\clearpage{}\clearpage{}

\clearpage{}\clearpage{}

\begin{figure}
\begin{centering}
\subfloat[\label{fig:caltech-test-set-reasonable}Standard evaluation (reasonable
subset)]{\begin{centering}
\includegraphics[width=0.88\columnwidth]{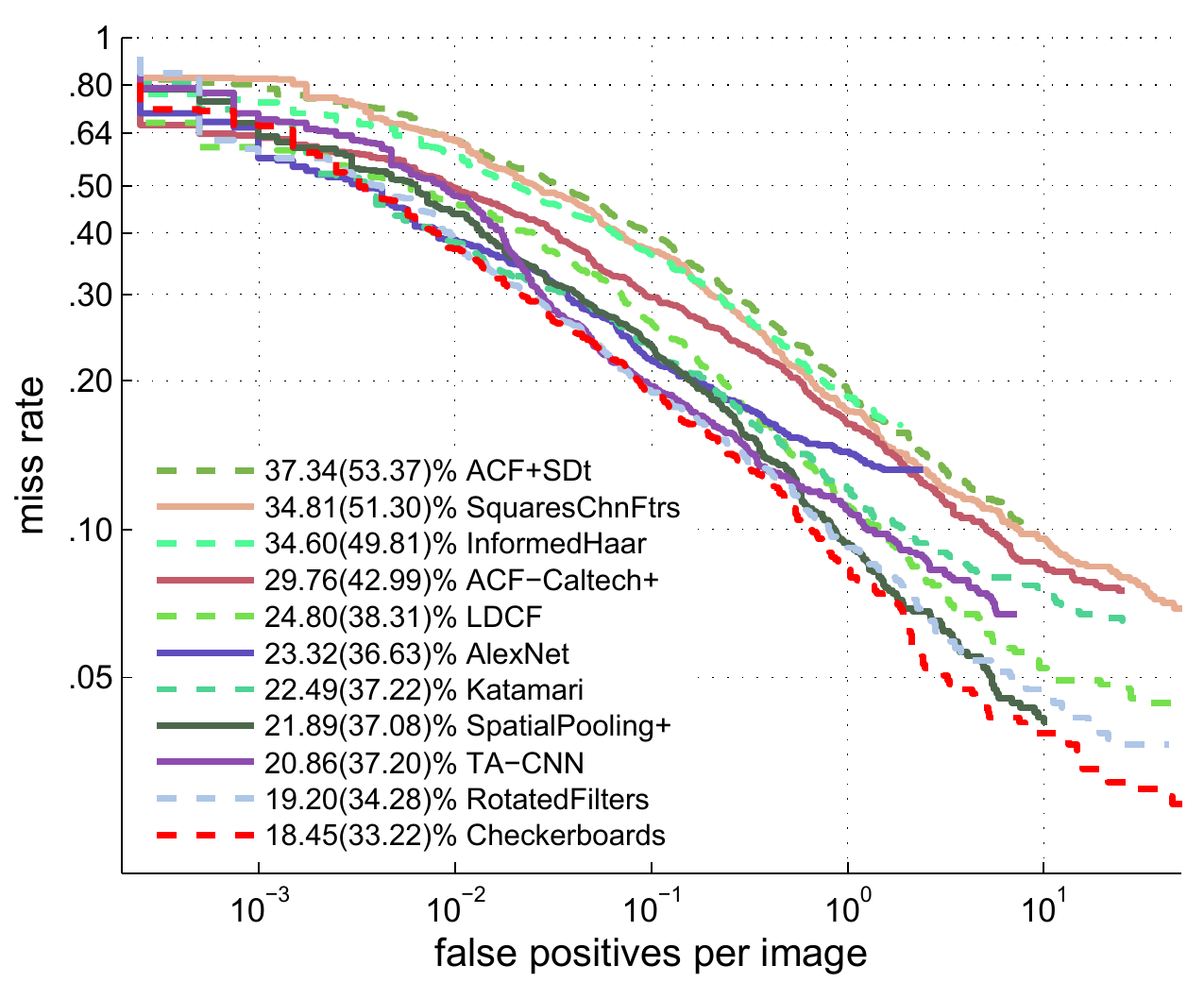}
\par\end{centering}

}
\par\end{centering}

\begin{centering}
\subfloat[\label{fig:localization-oracle}Localisation oracle]{\begin{centering}
\includegraphics[width=0.88\columnwidth]{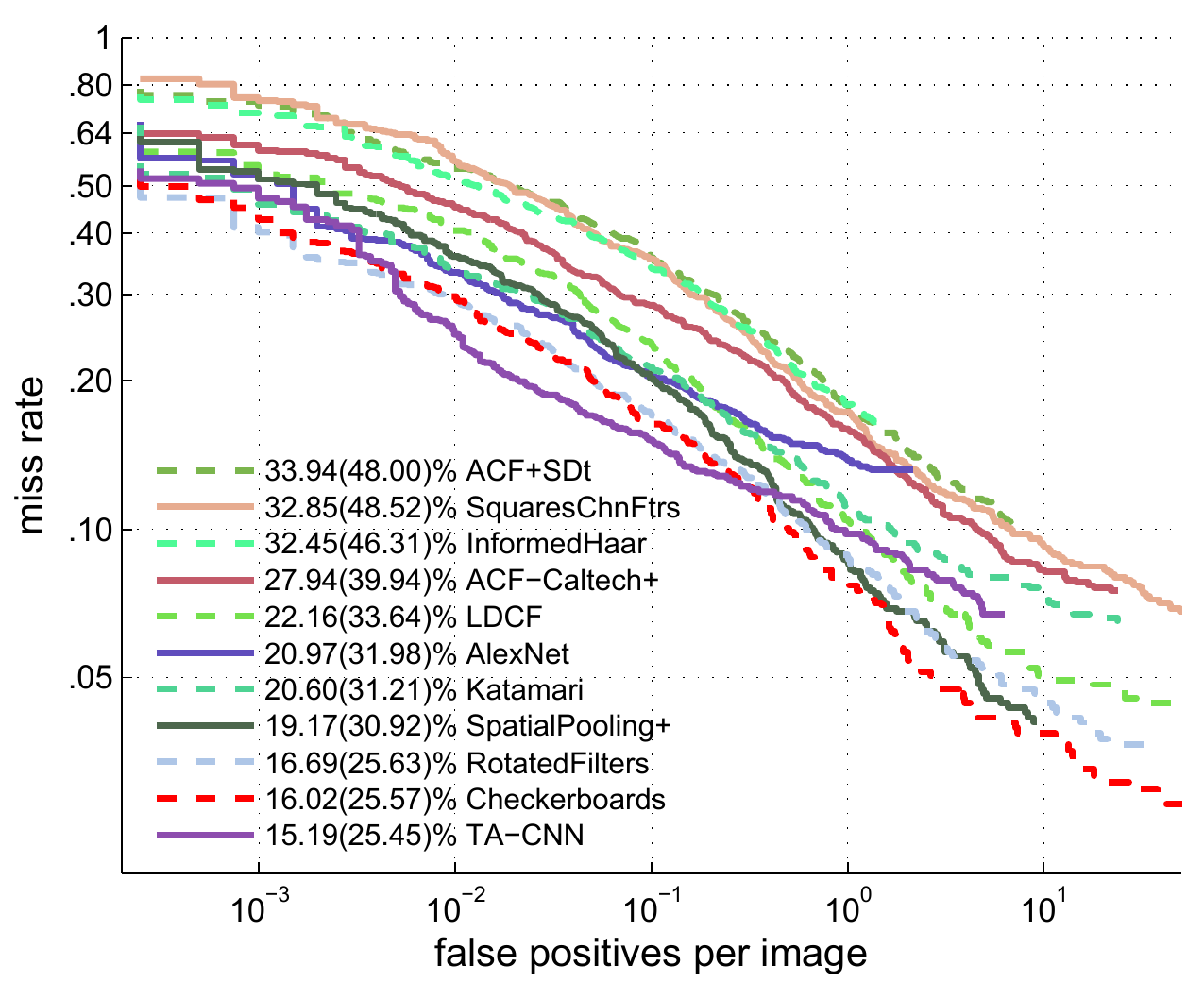}
\par\end{centering}

}
\par\end{centering}

\begin{centering}
\subfloat[\label{fig:background-oracle}Background-vs-foreground oracle]{\begin{centering}
\includegraphics[width=0.88\columnwidth]{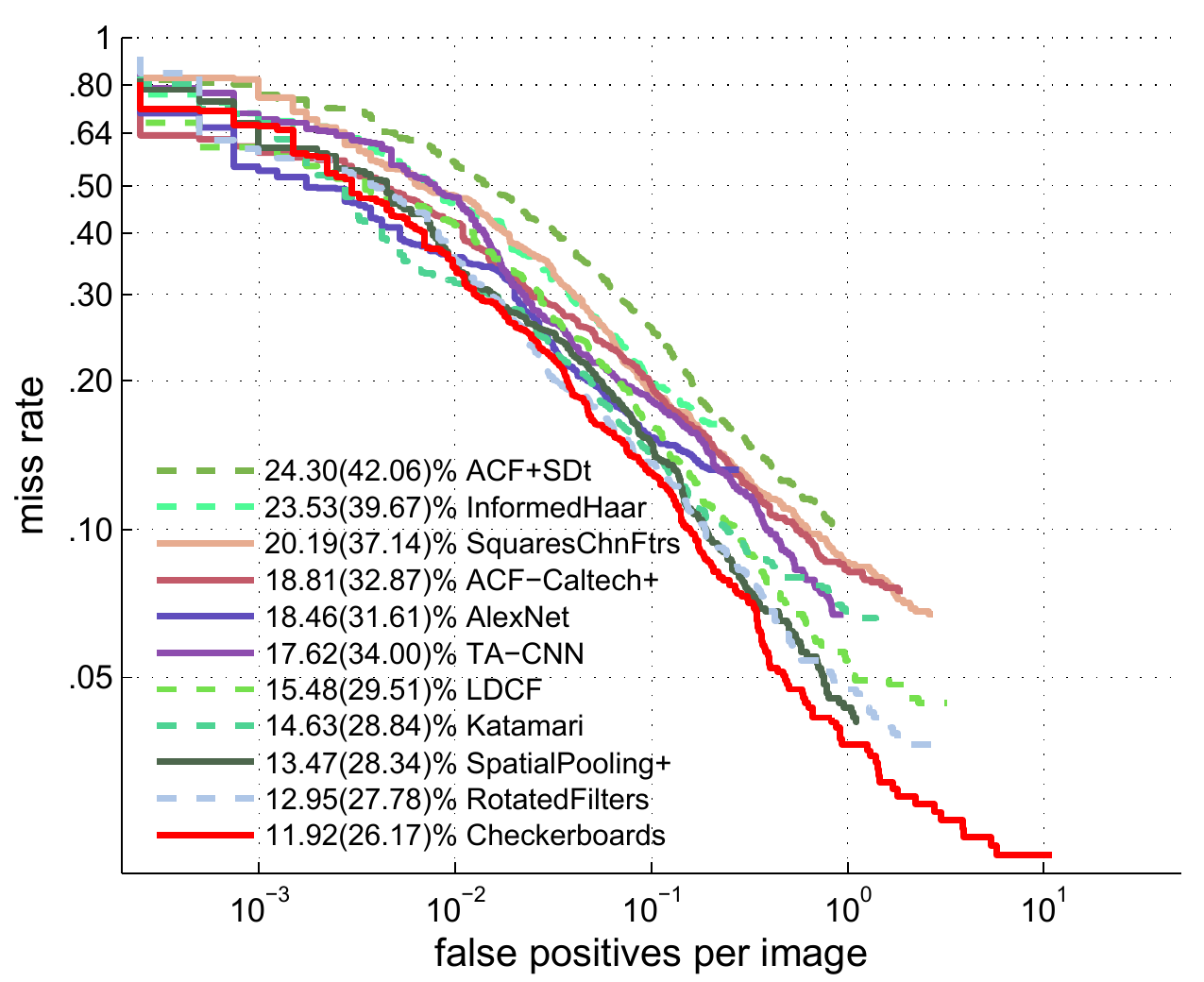}
\par\end{centering}

}
\par\end{centering}

\caption{\label{fig:oracle-cases-1}Caltech test set error with standard and
oracle case evaluations. Both localisation and background-versus-foreground
show important room for improvement. Both MR$_{-2}^{\mbox{O}}$ and
MR$_{-4}^{\mbox{O}}$ are shown for each method at each evaluation.}
\end{figure}

\begin{figure}[h]
\begin{centering}
\subfloat[\label{fig:oracle-gain-checkerboards-1}Original and two oracle curves
for \texttt{Checkerboards} detector.]{\begin{centering}
\includegraphics[width=0.8\columnwidth]{oracle_curves}
\par\end{centering}

\centering{}}
\par\end{centering}

\begin{centering}
\subfloat[\label{fig:loglog-is-strange-localize}Localisation FPs on the left.]{\begin{centering}
\includegraphics[width=0.7\columnwidth]{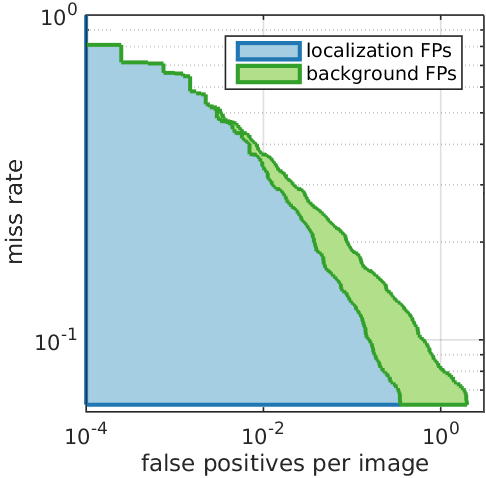}
\par\end{centering}

}
\par\end{centering}

\begin{centering}
\subfloat[\label{fig:loglog-is-strange-background}Background FPs on the left.]{\begin{centering}
\includegraphics[width=0.7\columnwidth]{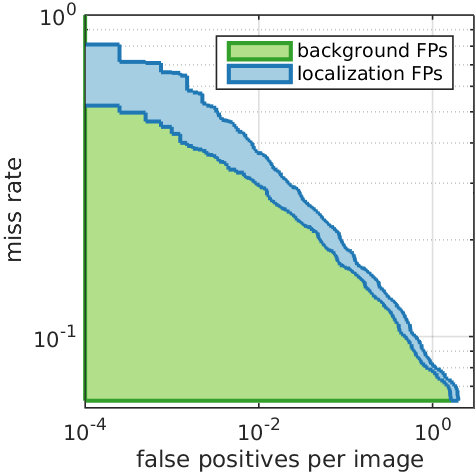}
\par\end{centering}

}
\par\end{centering}

\caption{\label{fig:loglog-is-strange}\texttt{Checkerboards} performance on
standard Caltech annotations, when considering oracle cases . Localisation
mistakes are blue, background mistakes green.}
\end{figure}

\clearpage{}\clearpage{}

\section{\label{sec:Improved-annotations-supp}Improved annotations}

\begin{figure*}
\begin{centering}
\includegraphics[width=0.8\linewidth]{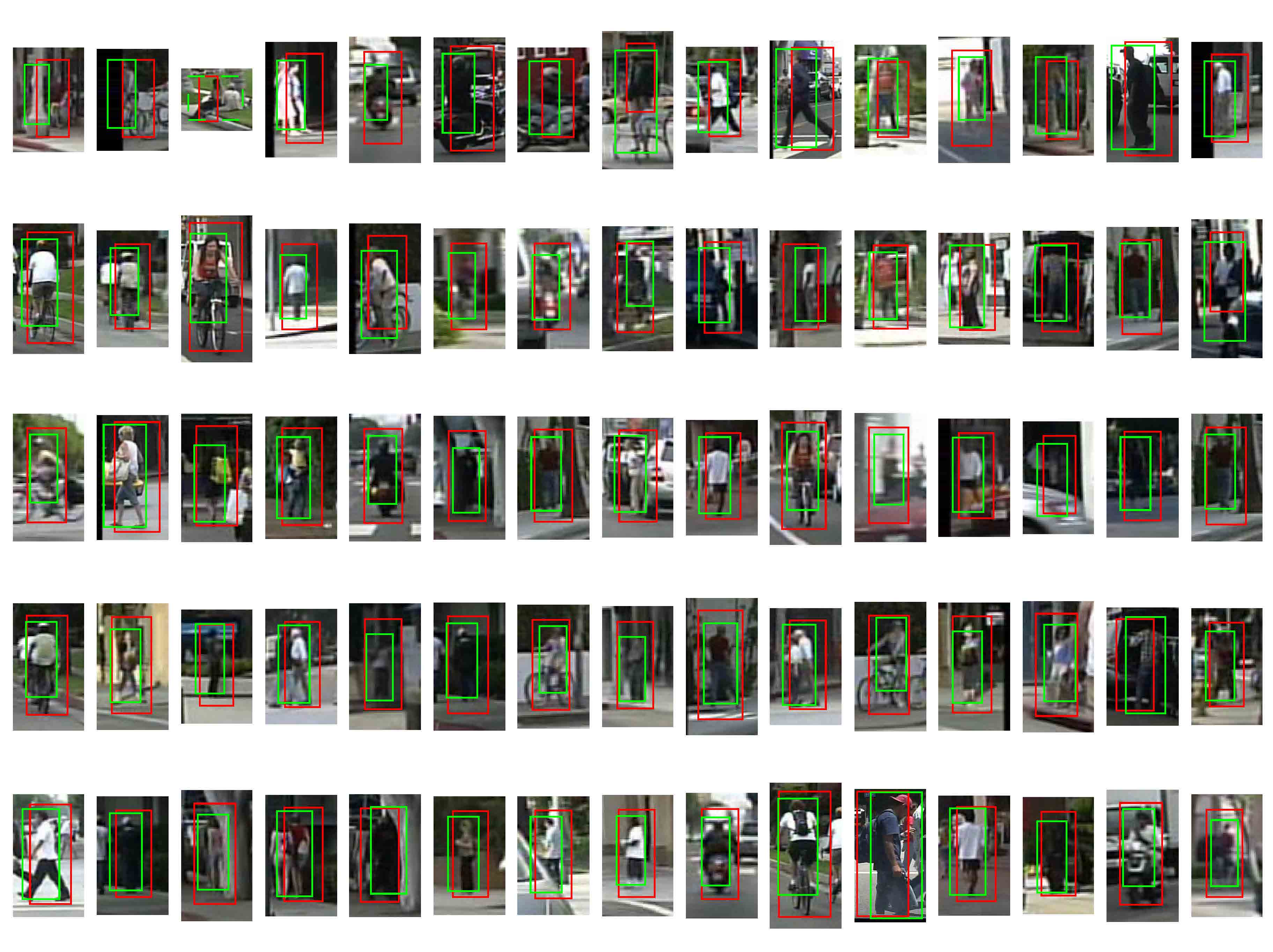}
\par\end{centering}

\begin{centering}
\includegraphics[width=0.8\linewidth]{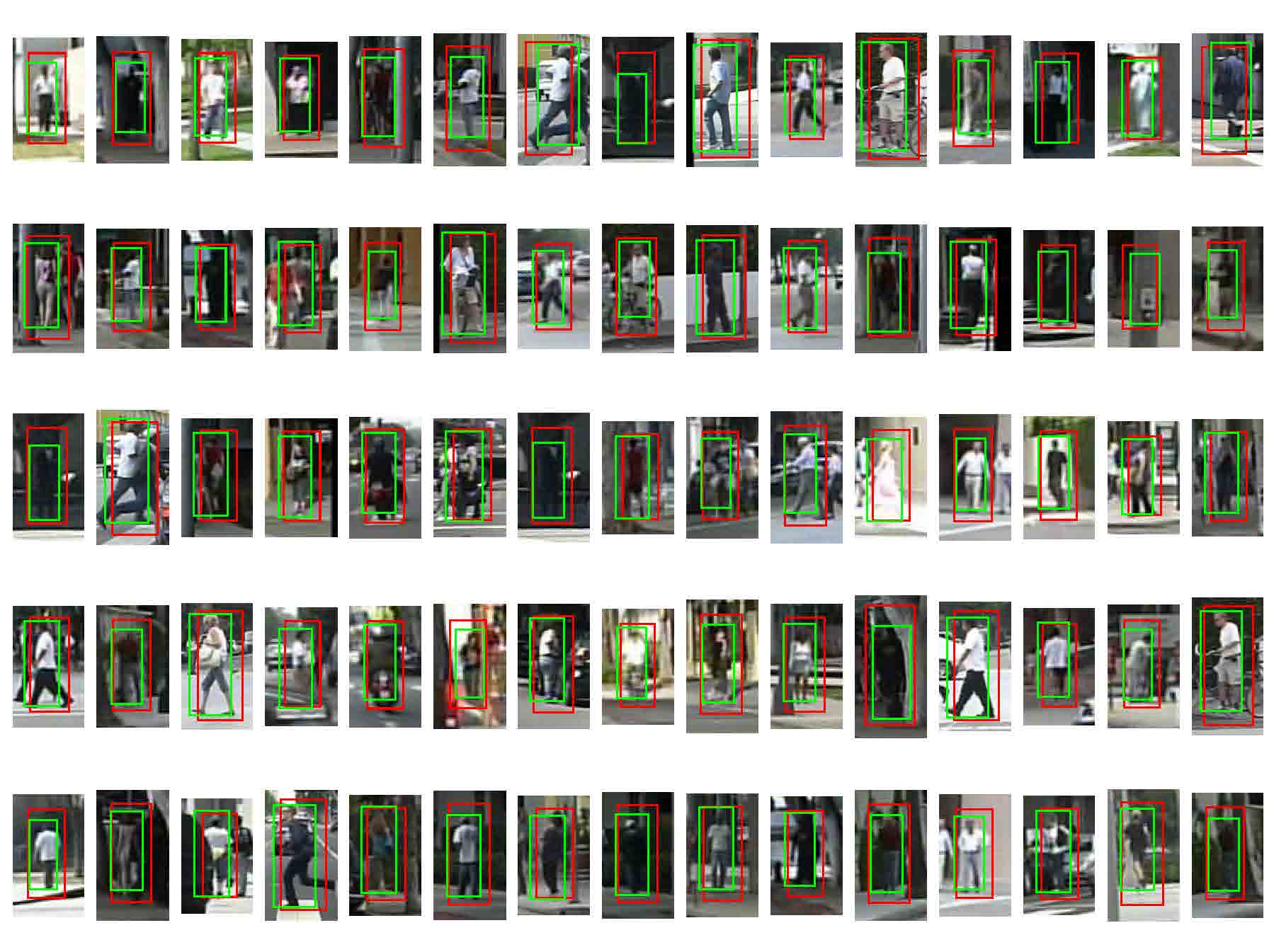}
\par\end{centering}

\caption{\label{fig:improved-annotations-all}Examples of differences between
original (red) and new annotations (green). Ignore regions are drawn
with dashed lines. These are the top 150 annotations, when sorted
from smallest to largest IoU between original and new annotations.}
\end{figure*}

In figure \ref{fig:improved-annotations-all} we show original (red)
and new annotations (green) on example frames from the test set. From
the comparison, we can see that the new annotations are better aligned
to the pedestrians. This results from the fact that head and feet
are closer to the centre of the new bounding boxes. 

\clearpage{}

\clearpage{}

\section{\label{sec:Old-vs-new-ranking}Evaluation on original and new annotations}

\paragraph{Ranking}

Figure \ref{fig:Ranking-of-methods} presents the ranking of all published
Caltech methods up to CVPR 2015 when evaluated on $\mbox{MR}_{-2}^{\text{{O}}}$
(original annotations), or on $\mbox{MR}_{-2}^{\text{{N}}}$ (proposed
new annotations). Although there are a few changes in ranking (e.g.
\texttt{JointDeep} versus \texttt{SDN}), the overall trend is preserved.
This is a good sign that the improved annotations are not a radical
departure from previous ones. As discussed in the paper (and in other
sections of the supplementary material), improved annotations matter
most for future methods (going further down in MR), and for the low
FPPI region of the curves (high confidence mistakes).

\paragraph{\texttt{Rotated\-Filters}}

Figures \ref{fig:final-results-oldanno} and \ref{fig:final-curves-newanno}
show the results of our methods \texttt{Ro\-ta\-ted\-Fil\-ters},
\texttt{Ro\-ta\-ted\-Fil\-ters\--New\-10x}, and \texttt{Ro\-ta\-ted\-Fil\-ters\--New\-10x\-+VGG};
on the original and new annotations respectively. Using improved annotations
during training (\texttt{-New10x) }does improve results both on original
and new annotations.

\paragraph{MR versus IoU}

\begin{figure}[t]
\begin{centering}
\hfill{}\subfloat[\label{fig:mr-vs-iou-original-annotations-1}Original annotations,
$\mbox{MR}_{-2}^{\text{O}}$]{\begin{centering}
\hspace*{-1em}\includegraphics[width=1.05\columnwidth]{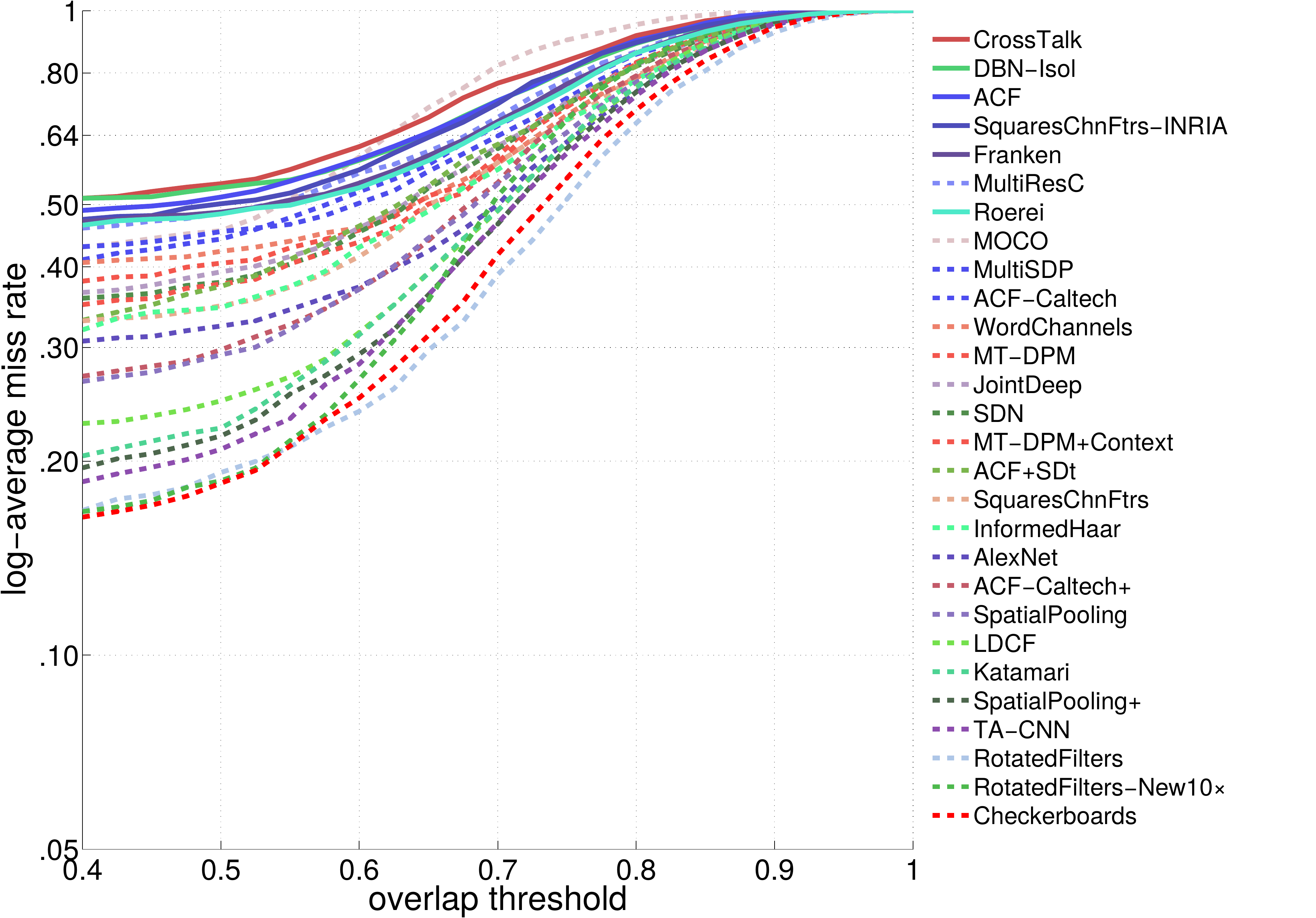}
\par\end{centering}

}\hfill{}\subfloat[\label{fig:mr-vs-iou-new-annotations-1}New annotations, $\mbox{MR}_{-2}^{\text{N}}$]{\begin{centering}
\hspace*{-1em}\includegraphics[width=0.9\columnwidth]{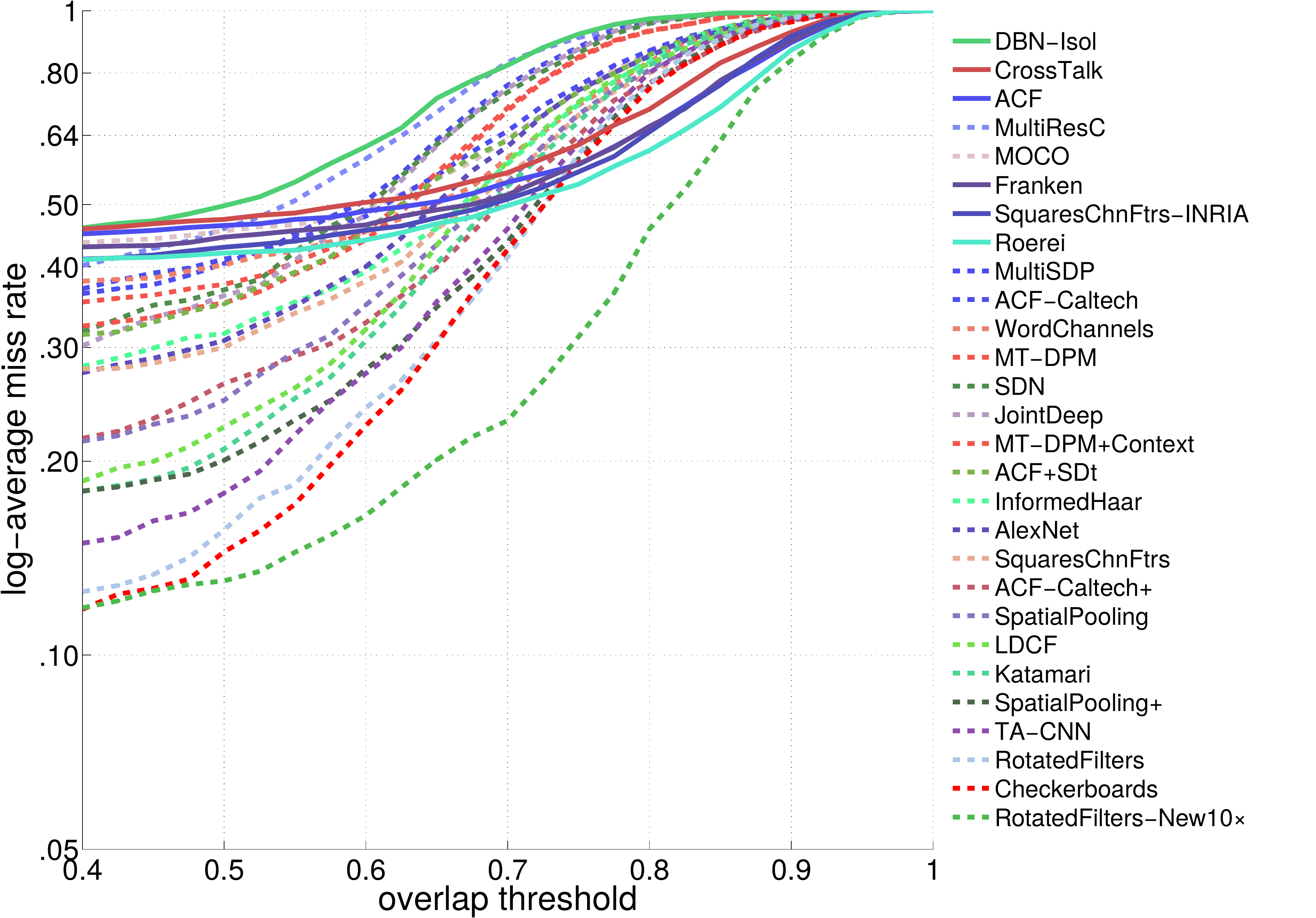}
\par\end{centering}

}\hfill{}
\par\end{centering}

\caption{\label{fig:mr-vs-iou-1}Plot of log-average miss rate versus overlap
threshold (IoU) for the top-performing methods on the ``reasonable''
experimental setting. Methods trained on INRIA are represented with
solid curves. On the new annotations, these behave better than methods
trained on Caltech-USA original when we apply a stricter overlap criterion.}
\end{figure}

\begin{figure*}
\centering{}\hspace*{\fill}\subfloat[\label{fig:mr_o2-ranking}$\mbox{MR}_{-2}^{\mbox{O}}$ Ranking of
methods]{\centering{}\includegraphics[width=0.48\textwidth]{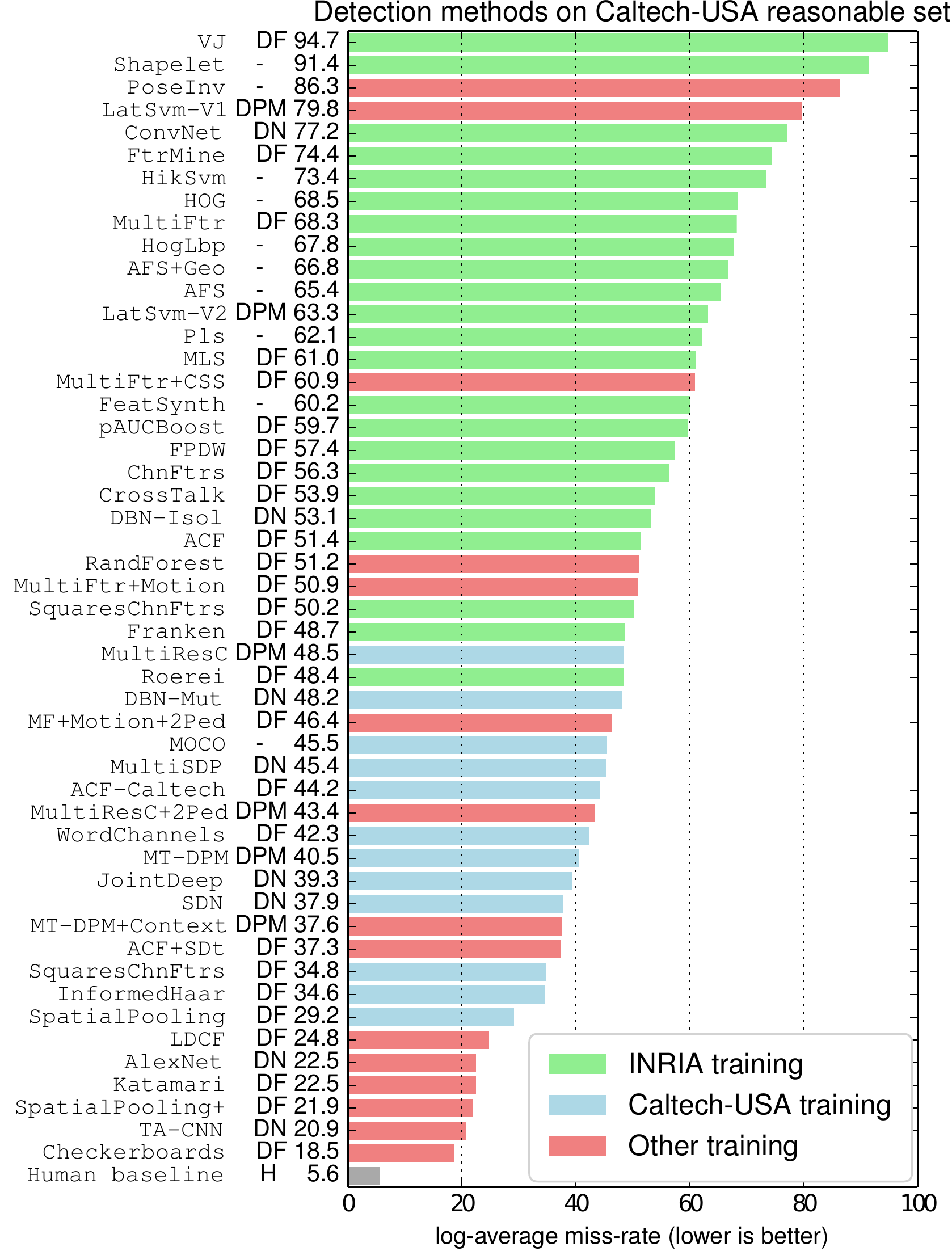}}\hspace*{\fill}\subfloat[\label{fig:mr_n2-ranking}$\mbox{MR}_{-2}^{\mbox{N}}$ Ranking of
methods]{\centering{}\includegraphics[width=0.48\textwidth]{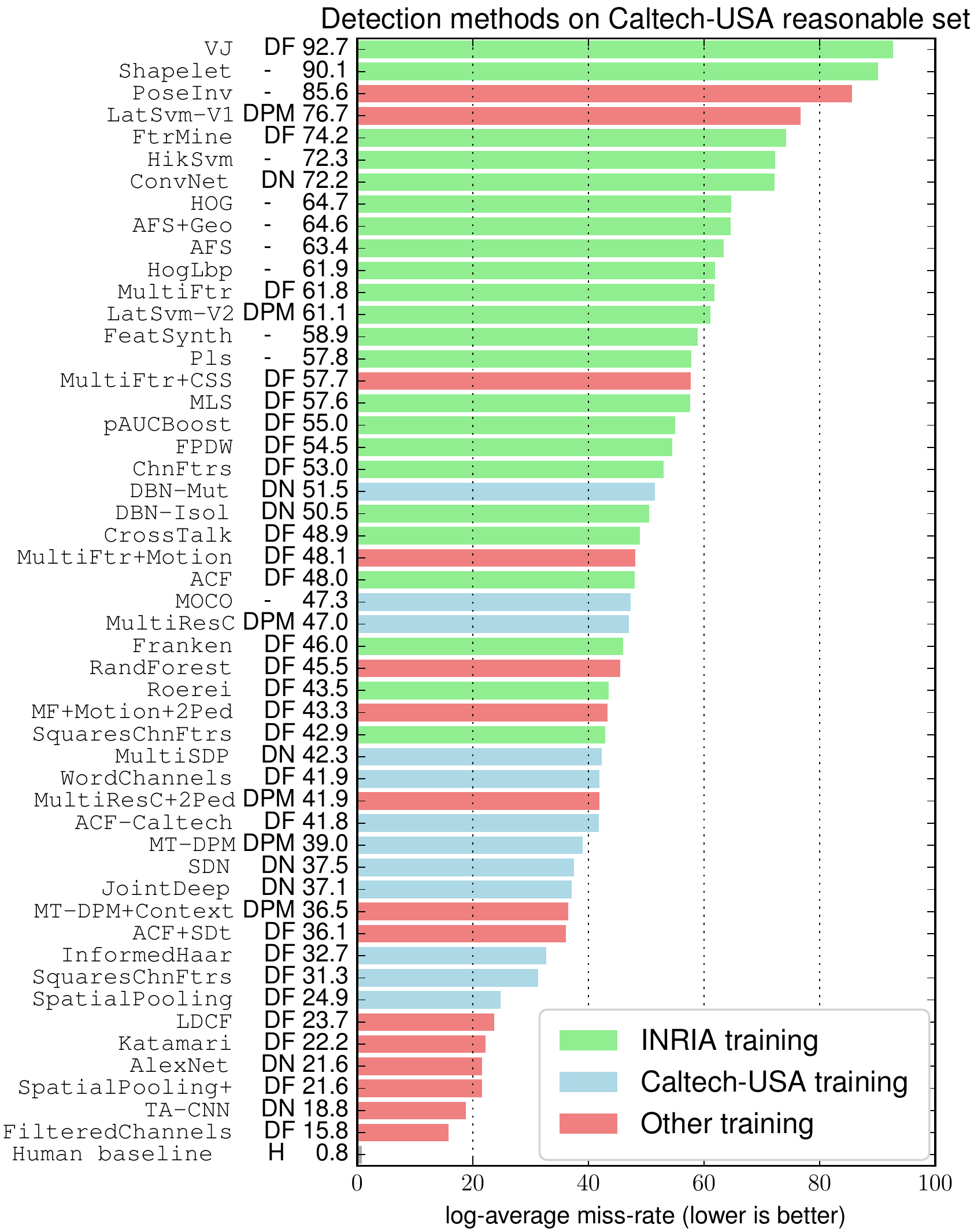}}\hspace*{\fill}\caption{\label{fig:Ranking-of-methods}Ranking of Caltech methods (CVPR 2015
snapshot) with original and new annotations. DF: decision forest,
DPM: deformable parts model, DN: deep network.}
\end{figure*}

\begin{figure*}
\begin{centering}
\hspace*{\fill}\subfloat[\label{fig:final-results-oldanno}Evaluation on original annotations.
Legend indicates MR$_{-2}^{\text{{O}}}$(MR$_{-4}^{\text{{O}}}$).]{\begin{centering}
\includegraphics[width=1\columnwidth]{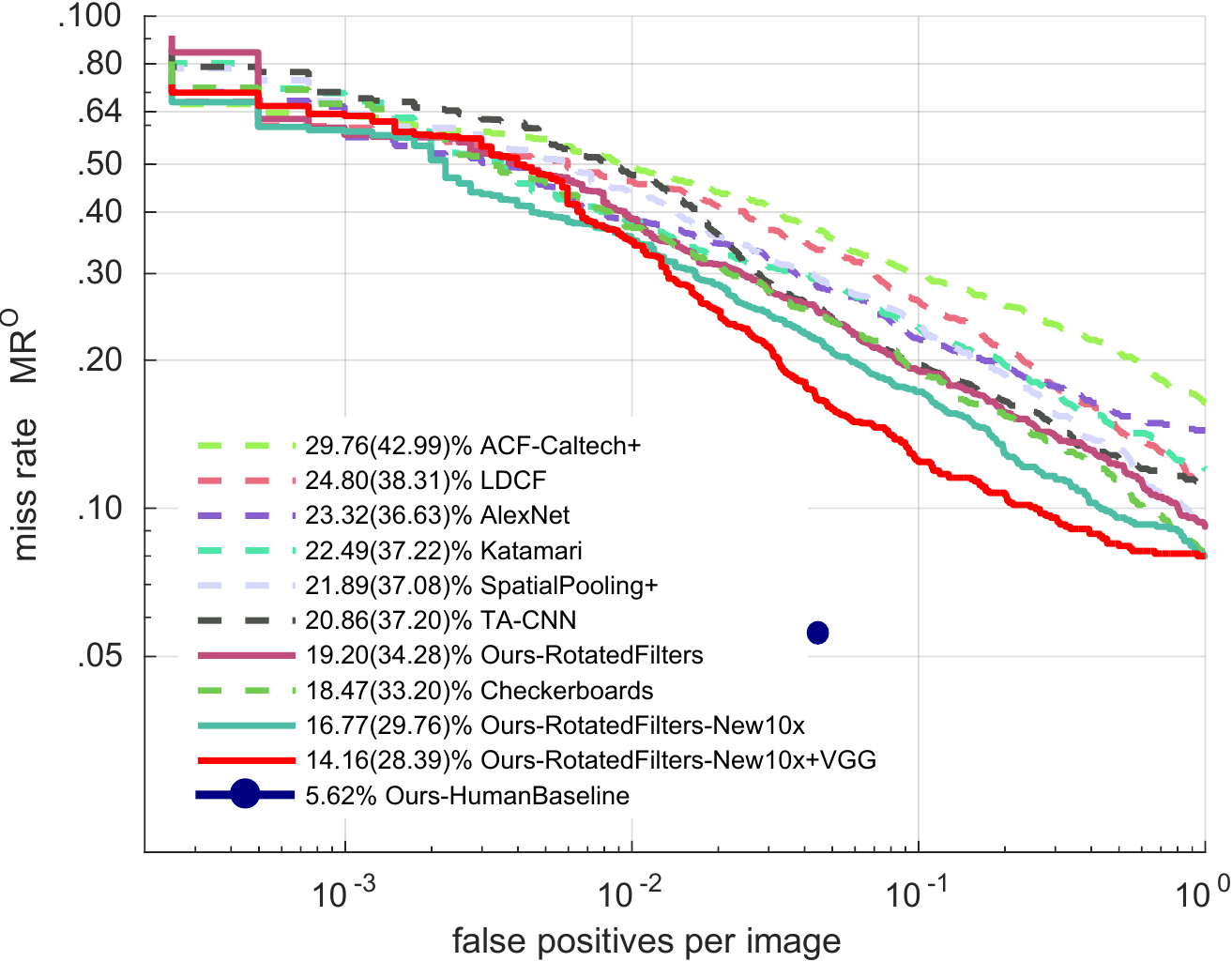}
\par\end{centering}

}\hspace*{\fill}\subfloat[\label{fig:final-curves-newanno}Evaluation on new annotations. Legend
indicates MR$_{-2}^{\text{{N}}}$(MR$_{-4}^{\text{{N}}}$).]{\begin{centering}
\includegraphics[width=1\columnwidth]{curves-final-newanno-ssclean}
\par\end{centering}

}\hspace*{\fill}
\par\end{centering}

\caption{\label{fig:final-curves}Performance of top detectors evaluated on
original and new annotations. }
\end{figure*}

Section 3.3 (and table 3) of the main paper discuss an empirical measure
of how the new annotations are better aligned. Here we provide some
more details.\\
Figure \ref{fig:mr-vs-iou-1} plots $\mbox{MR}_{-2}^{\text{{O}}}$
and $\mbox{MR}_{-2}^{\text{{N}}}$ of top performing methods versus
the overlap criterion for accepting detections as true positives (IoU
threshold). The standard evaluation uses IoU threshold $0.5$. On
these plots methods trained on INRIA have continuous lines, methods
trained on Caltech dashed ones (see also figure \ref{fig:Ranking-of-methods}).\\
In figure \ref{fig:mr-vs-iou-original-annotations-1} (original annotations)
the ranking of the methods remains stable as the overlap threshold
becomes stricter (consistent with the observations in \cite{Dollar2012Pami}).
Interestingly we observe a different trend in figure \ref{fig:mr-vs-iou-new-annotations-1}
(new annotations).\\
When evaluating $\mbox{MR}_{-2}^{\text{{N}}}$(new annotations) we
see that methods training on INRIA, albeit having a poor performance
at $\text{{IoU}}=0.5$, perform comparatively well at higher IoU,
eventually overpassing all methods trained on raw Caltech data. We
attribute this to the fact that INRIA training data is of better quality
(better aligned training samples), and thus the detectors have learnt
to localise better. This difference in trend between original and
new annotations confirms that our improved annotations are better
with respect to localization. Table 3 in the main paper provides a
summarised version of figure \ref{fig:mr-vs-iou-1}.

\section{\label{sec:Impact-of-training-annotations-supp}Impact of aligning
Caltech$10\times$}

\begin{figure*}[h]
\begin{centering}
\includegraphics[width=0.8\linewidth]{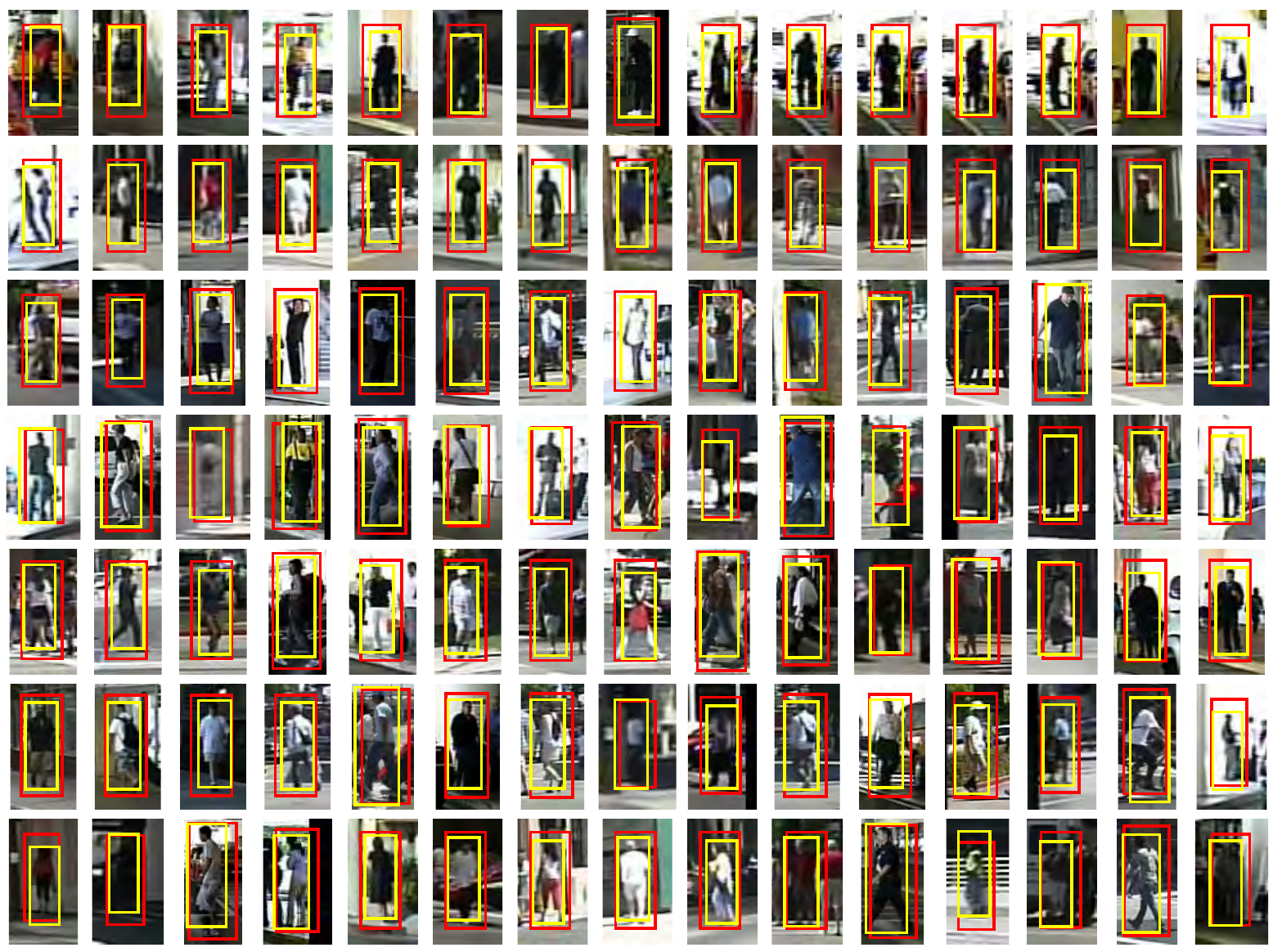}
\par\end{centering}

\begin{centering}
\includegraphics[width=0.8\linewidth]{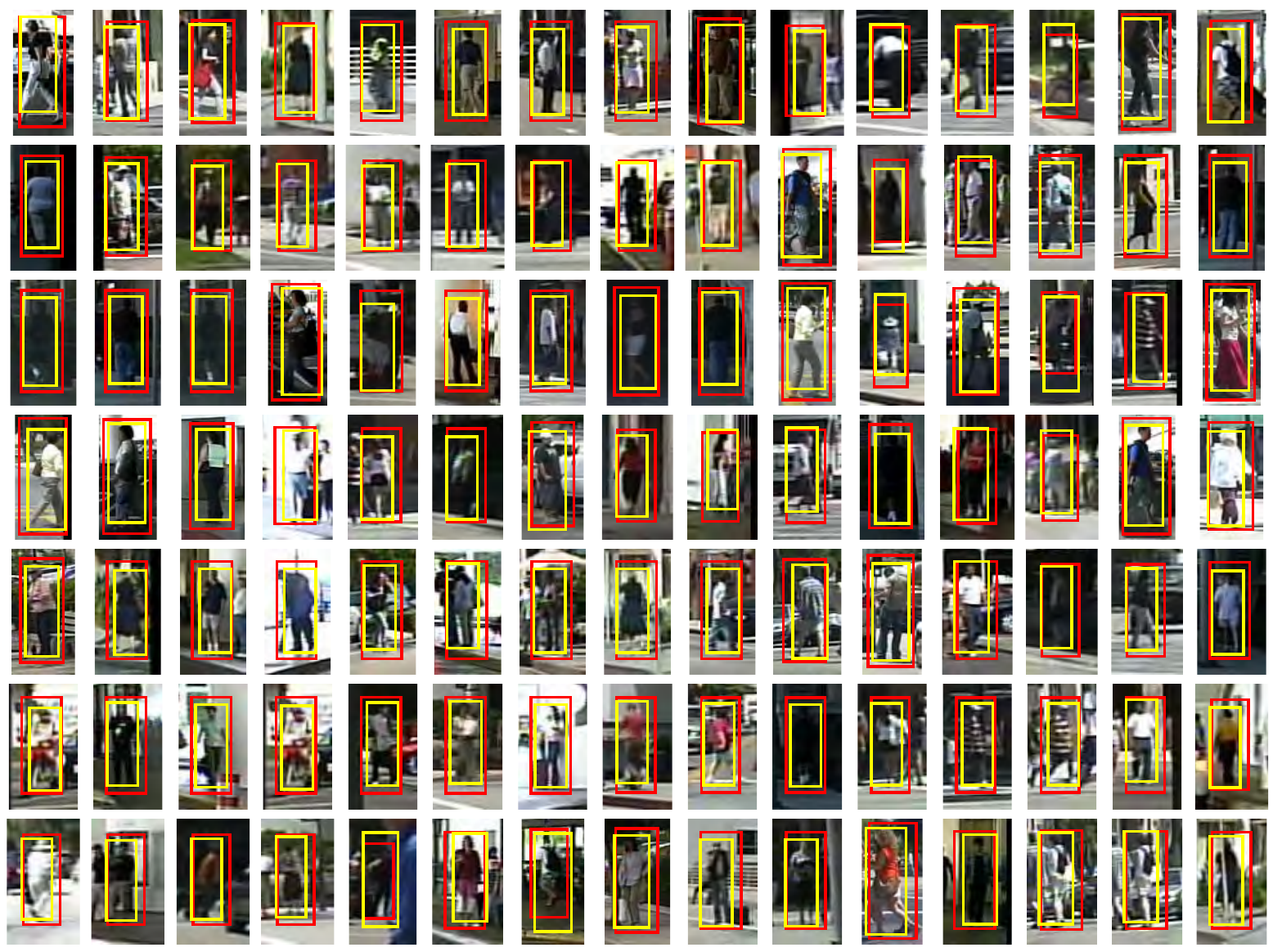}
\par\end{centering}

\centering{}\caption{\label{fig:annotation-alignment}Examples of original annotations
before (red bounding boxes) and after automatic alignment (yellow
bounding boxes) using the \texttt{Rotated\-Filters} detector.}
\end{figure*}

We can see from \ref{fig:mr-vs-iou-new-annotations-1} that using
our semi-automatically aligned Caltech $10\times$ training data provides
a significant boost in localization quality. From \texttt{Rotated\-Filters}
to \texttt{Rotated\-Filters\--New10x} the $\mbox{MR}_{-2}^{\text{{N}}}$
improves across the full $\mbox{IoU}$ range. Figure \ref{fig:annotation-alignment}
shows qualitative results for the alignment procedure done over the
$10\times$ training data.

\clearpage{}
\end{document}